\documentclass[preprint,journal]{vgtc}            

\onlineid{0}

\preprinttext{}

\vgtccategory{Research}

\title{Neural Projection Mapping Using Reflectance Fields}

\author{%
  Yotam Erel,
  Daisuke Iwai, and 
  Amit H.\ Bermano
}

\authorfooter{
  \item
  	Yotam Erel is with Tel-Aviv University.
  \item
  	Daisuke Iwai is with Osaka University.

  \item Amit H. Bermano is with Tel-Aviv University.
}

\abstract{%
  We introduce a high resolution spatially adaptive light source, or a projector, into a neural reflectance field that allows to both calibrate the projector and photo realistic light editing. The projected texture is fully differentiable with respect to all scene parameters, and can be optimized to yield a desired appearance suitable for applications in augmented reality and projection mapping. Our neural field consists of three neural networks, estimating geometry, material, and transmittance. Using an analytical BRDF model and carefully selected projection patterns, our acquisition process is simple and intuitive, featuring a fixed uncalibrated projected and a handheld camera with a co-located light source. As we demonstrate, the virtual projector incorporated into the pipeline improves scene understanding and enables various projection mapping applications, alleviating the need for time consuming calibration steps performed in a traditional setting per view or projector location. In addition to enabling novel viewpoint synthesis, we demonstrate state-of-the-art performance projector compensation for novel viewpoints, improvement over the baselines in material and scene reconstruction, and three simply implemented scenarios where projection image optimization is performed, including the use of a 2D generative model to consistently dictate scene appearance from multiple viewpoints. We believe that neural projection mapping opens up the door to novel and exciting downstream tasks, through the joint optimization of the scene and projection images.
}


\teaser{
  \centering
  \includegraphics[width=\linewidth]{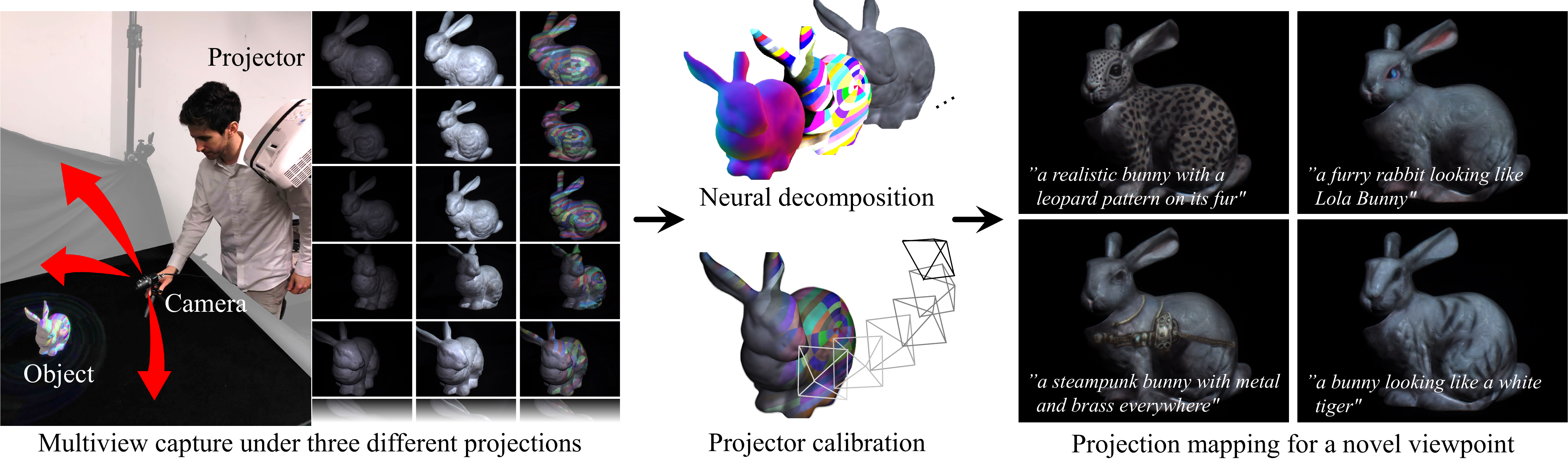}
  \caption{%
  	A real scene augmented by our method using a projector. Left: acquisition of a scene using a handheld camera (moving in the indicated directions) with a co-located light, and a fixed projector shining 3 types of patterns. Middle: We reconstruct and decompose the scene as a neural reflectance field allowing us to synthesize novel viewpoints with novel high resolution lighting. The depicted decomposed channels are (left to right) normal directions, projected irradiance, and albedo. Projector's intrinsics and extrinsics are jointly optimized. Right: We perform a realistic and content aware relighting using projection mapping and reproject the result onto the real scene. The augmentations are driven by a natural language user text prompt (shown within each sub-image) and a diffusion-based generative model, and can be optimized for multiple viewpoints, accounting for surface geometry and material.
  }
  \label{fig:teaser}
}

\graphicspath{{figs/}{figures/}{pictures/}{images/}{./}} 

\usepackage{tabu}                      
\usepackage{booktabs}                  
\usepackage{tabularx}
\usepackage{lipsum}                    
\usepackage{mwe}                       
\usepackage{wrapfig}
\usepackage{array}
\usepackage{graphicx}
\usepackage{multirow}
\usepackage{stackengine}
\usepackage{array}
\usepackage{amsmath}
\usepackage{amssymb}
\usepackage{colortbl}
\usepackage{color}
\usepackage[normalem]{ulem}
\begin{document}
\newcolumntype{Y}{>{\centering\arraybackslash}X}

\firstsection{Introduction}

\maketitle
\label{sec:intro}

Augmenting the appearance of objects using projection mapping is a well established technique which has major usage in numerous fields including medicine~\cite{00000658-201806000-00024}, teleconferencing~\cite{10.1145/2818048.2819965,8172039,10.1145/280814.280861}, museum guides~\cite{SCHMIDT20191,1377099}, makeup~\cite{https://doi.org/10.1111/cgf.13128,8007312}, object searches~\cite{10.1145/1015706.1015738,10.1145/1959826.1959828,8007248,10.1145/1180495.1180519}, product design~\cite{8797923,https://doi.org/10.1111/j.1467-8659.2011.02066.x,6949562,CASCINI2020103308}, urban planning~\cite{10.1145/302979.303114}, artwork creation~\cite{10.1145/2366145.2366176,10.1145/1166253.1166290,970539}, theme parks~\cite{6193074} and more. Even though projectors serve as a convenient and affordable tool for SAR applications, the process of preparing a convincing augmentation entails geometric calibration of the optical elements including cameras and projectors, and color compensation techniques that help project dynamically on irregular shaped objects, or objects that have various reflective properties. This process is traditionally performed using classical structured light pattern projection and radiometric calibration techniques~\cite{https://doi.org/10.1111/j.1467-8659.2008.01175.x,https://doi.org/10.1111/cgf.13387}, and are constrained to work for a limited or fixed set of viewpoints and projector locations. Furthermore, the editing capabilities that these processes enable from an artistic point of view require specific technical expertise and is a very long and laborious process of fine tuning the output. Still, projectors are much more accessible than high-end industrial light stages used in commercial applications \cite{light_stage}.

In parallel, recent advancements in the field of neural implicit representations exhibit remarkable results for novel multi view synthesis, scene decomposition, and scene generation \cite{nerf, nerfactor, dreamfusion, clip-nerf, nvidia_extract_models}. 
While this approach seems promising for projection mapping applications, no attempts that we are aware of have been made to integrate such capabilities. Hence our first insight is to introduce high-resolution direct lighting into a neural field framework. 
Neural reflectance fields \cite{nerfaa, nerv, nerfactor} model a scene as an absorption-reflectance field which receives radiation and reflects it, and is highly suitable for projection mapping where the direct light component dominates the light field. However, we demonstrate that simply using such models to describe a scene and placing a projector in them during inference is not enough for high quality projection mapping.

We propose an end-to-end technique to incorporate a self-calibrating projector into a neural reflectance field that helps with scene understanding, augments synthetic and real scenes photo-realistically, and naturally generalizes to different viewpoints with no extra hardware. The main challenge for such a system is acquisition effort. Our method only requires a handful of freely taken photos of the scene while being illuminated by the projector used in the setup, and a light co-located with the camera. We use a simple yet expressive BRDF (bidirectional reflectance distribution function) material model that lends itself to projector like-illumination. In addition, we guide the training process through a three-step training policy, aided by different projection images. 

During acquisition, the projector cycles between three illumination presets, aiding in geometry extraction, material estimation, and projector calibration. For the latter, we propose a lollipop pattern that exhibits better guidance for projector localization. Next, a neural reflectance field is trained. Inspired by handful of neural relighting works \cite{nerfaa, nerv, nerfactor, refnerf}, our architecture consists of three networks, estimating geometry, material, and transmittance (i.e. obstructions along the line-of-sight between any given two points). The material and light transport models we use translate the aforementioned predictions into a color value, that in turn is compared against the acquired images. Our calibration compensates for the change in perceived color due to light and material interaction, which allows editing the scene or observing it from novel viewpoints and projections without changing the physical setup. Previous techniques do not allow for novel view projection mapping augmentation without additional equipment or specialized hardware. Finally, since our virtual projector has all its parameters differentiable relative to scene appearance, we are able to optimize its projected image toward some desired appearance. 

Using our framework, we demonstrate superior results for novel viewpoint estimation under projector illumination compared to base-line neural field approaches, and show the merit of project-based illumination for material estimation. We further demonstrate state-of-the-art appearance compensation results for novel views. Lastly, we present the first tool for text-to-projection mapping generation. We do so by using a text-to-image diffusion-based generative model \cite{cdc} conditioned to match the geometry seen for a desired point-of-view, to generate desired appearances from different angles. We then jointly optimize a single projection images to match the generated results. The core contributions of this paper are:
\begin{itemize}
    \item The introduction of a high resolution direct light source into a neural reflectance field, and its self-calibration.
    \item Improved scene decomposition using the new light source, enabling better projection mapping quality.  
    \item A technique to photo realistically augment a scene for multiple views without the need for extra hardware or new calibration.
\end{itemize}

\section {Related Work}
\label{sec:related}

\subsection{Neural reflective fields}
Our study can be most accurately framed as an inverse rendering problem, where given some images the goal is to reconstruct the 3D scene including its geometry, materials and lighting conditions. In their seminal work, NeRF \cite{nerf} represents the scene using an absorbtion-emmision model that is parameterized by a neural network for every 3D continuous location. The scene however must remain static, and lighting conditions are assumed to be fixed. Bi et al. \cite{nerfaa} expanded on this idea by letting a neural network predict parameters of an analytical BRDF, and by using an absorbtion-reflection model instead. Provided with a co-located light, they have shown it is possible to extract normals, albedo and roughness essentially defining a SVBRDF (spatially varying BRDF). This in turn allows moving the light during inference and create plausible appearances. However the light is of single color which allows very limited editing and it is not clear how to handle more global illumination effects such as shadows. Furthermore the sole usage of a co-located light means only a crude approximation of material properties is possible, since the angle between view and light are fixed which reduces the ability to estimate BRDF parameters faithfully. NeRV \cite{nerv} extended this idea by estimating visibility using an ad-hoc visibility network, and accounting for global illumination by probing it instead of tracing secondary rays. However, NeRV requires a known lighting as input and the lighting model is perhaps simplistic as it consists of sparse point light sources and low resolution environment map. These can indeed be used to edit the lighting of the scene, but in a very smoothed and global way. It can’t serve to directly augment object appearance. Zhang et al. \cite{nerfactor} removed the requirement of a known light source, by performing a multi-step optimization to first extract the surface, and jointly estimating a HDR light probe image together with visibility, normals and a BRDF function embedding learned and mapped to real-world BRDFs using a pre-trained network. This allows to inject novel light sources in inference, however the light source is still an environment map and of relatively low-resolution, far from the details demanded of augmented reality applications. In PS-NeRF~\cite{psnerf}, the authors use a similar representation to reconstruct the scene materials and geometry to a very accurate degree, but they require a set of multi-view and multi-light images taken in a light stage, which limits the practicality of such method. Munkberg et al. \cite{nvidia_extract_models}, depart somewhat from the reflective field representation and perform high quality decomposition of the scene leveraging a differentiable rasterizer, and optimizing geometry explicitly as a 3D mesh along with textures and an environment map lighting, their method does not however support a direct illumination source augmenting the scene during training.

As opposed to these, our method can handle one unknown high resolution direct light source, allowing to both calibrate its optical properties and to optimize over its projected image. The only hardware requirements are a RGB camera and a projector. This in turn allows for better surface understanding and facilitates new applications in the field of projection mapping.

\subsection{Geometric correction and color compensation in projection mapping}\label{sec:relwork_procam}

In projection mapping, the projected textures are geometrically and photometrically distorted on ordinary non-planar and textured surfaces.
The undesired geometric distortion can be corrected by obtaining the pixel correspondences between the projector and a camera placed at the observer's viewpoint.
The most well-known technique for the correspondence acquisition is structured light (e.g., graycode) pattern projection.
This allows to undistort the projected result, though it is constrained to a single, calibrated viewpoint.
Researchers overcome this limitation by developing self-calibration techniques which calibrate the projector's extrinsics (translation and rotation in 3D space with respect to the surface) and reconstruct the surface shape from the obtained pixel correspondences~\cite{5981781,8115403}.
However, the above geometric correction techniques suffer from photometric distortion on textured surfaces. Color compensation techniques have been investigated to neutralize the surface textures~\cite{naive_compensate, tps}. They estimate the reflectance properties of the surface by projecting color patterns and compute projection colors by solving inverse reflectance models such that the desired colors are displayed on the surface (e.g., making a pixel brighter on a dark pigment). The classical techniques rely on the pixel correspondences obtained by the structured light pattern projection, and thus, are constrained to a single viewpoint. In theory, we should achieve color compensation for novel viewpoints by combining the geometric self-calibration and color calibration techniques. However, in reality, because a physical surface does not have ideal Lambertian properties, the color compensation fails for viewpoints different from the calibrated ones.

\begin{figure*}[h]
    \centering
    \includegraphics[width=0.99\textwidth]{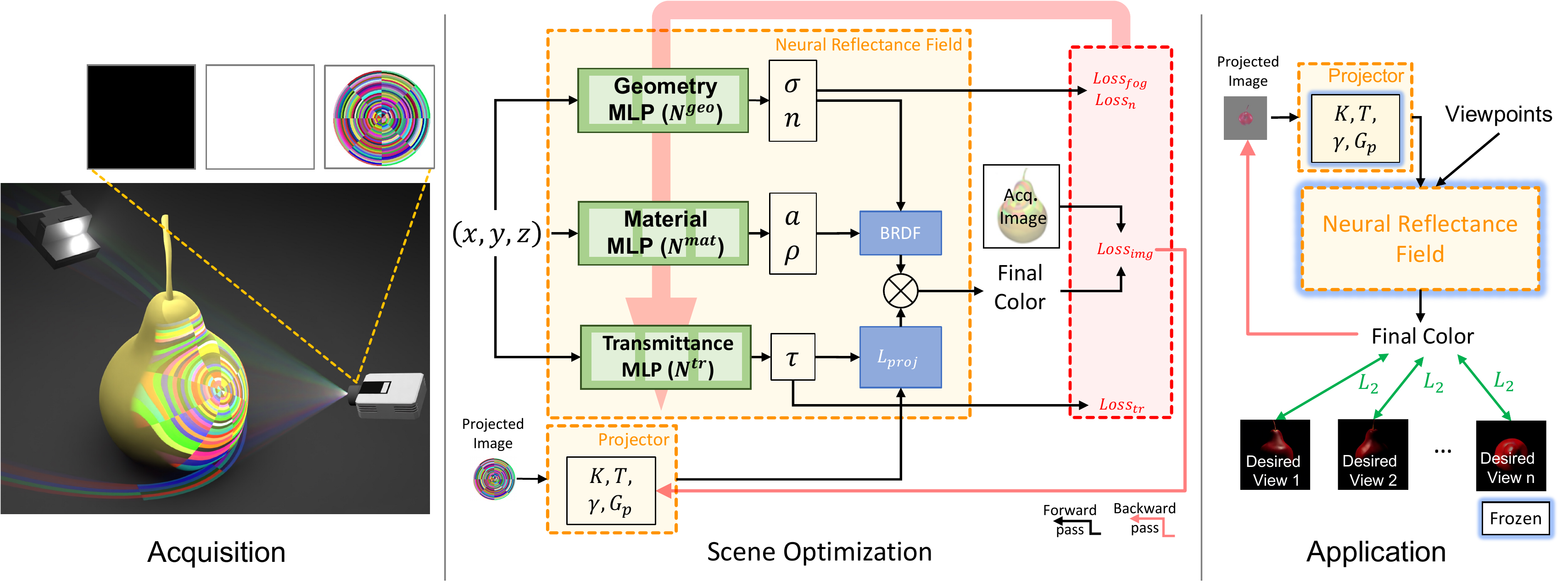}
    \caption{Our neural projection mapping framework. \textbf{Left}: First, the scene is acquired. A hand held camera captures images around the scene, illuminated by a fixed uncalibrated projector, and a light co-located with the camera. The projector cycles between all white, black, and a prescribed lollipop illumination to better calibrate geometry and material. \textbf{Middle}: Then, a neural reflectance field is trained. Using black projector illuminated frames, geometry and materials are reconstructed, and using the white and lollipop illuminated frames, projector parameters are calibrated. Our Field consists of three neural networks, one to predict geometry, one material, and one that assists in evaluating transmittance. They directly predict BRDF parameters, and is translated to the final color taking into account the irradiance due, according to the input projected image. Projector extrinsics and intrinsics are co-optimized as learned constants. \textbf{Right}: Once a scene is learned, it can be leveraged in a gradient based optimization to approximate scene appearance from novel views, allowing the direct optimization of the projection image, taking into account multiple points of view.}
    \label{fig:overview}
\end{figure*}

In another line of work, recent projection mapping techniques perform a joint estimation of color compensation and geometric calibration with deep neural networks. Huang et al. \cite{compensnet_pp,deprocams} achieved high performance by training an image-to-image network which is tasked with finding parameters of a thin plate spline warp field to account for the geometry of the projected surface and the projector-camera physical setup, and jointly color compensates with a U-Net like neural network. They are however constrained to a single camera and single projector location, and project to (roughly) planar surfaces, since the warping field is limited in its application for difficult geometries. Park et al. \cite{diff_projector} posed the calibration and compensation as an inverse rendering problem in a similar manner to our solution, but again are constrained to a single projector-camera pair location and require a RGBD camera.
In many of these techniques, the decomposition of the scene into interpretable physical quantities is key to high performance.

\begin{table}[t]
    \caption{Comparison of existing and our projection image correction techniques. The calibrated viewpoint is the position where a camera is placed in calibration. The novel viewpoint is other than that. The simple surface geometry is roughly planar.}
    \label{table:proj_rel_work}
    \centering
    \small
    \begin{tabular}{l|c|c|c|c|c}
    \hline
    & \multicolumn{3}{c|}{Geometric correction} & \multicolumn{2}{c}{Color compensation}\\
    \cline{2-6}
    \multicolumn{1}{r|}{Viewpoint$\rightarrow$} & \multicolumn{2}{c|}{Calibrated} & Novel & Calibrated & Novel\\
    \cline{2-6}
    \multicolumn{1}{r|}{Surface geometry$\rightarrow$}& Simple & Complex & - & - & -\\
    \hline
    SL & \cellcolor[rgb]{0.5,1.0,0.5}yes & \cellcolor[rgb]{0.5,1.0,0.5}yes & \cellcolor[rgb]{1.0,0.5,0.5}no & \cellcolor[rgb]{1.0,0.5,0.5}no & \cellcolor[rgb]{1.0,0.5,0.5}no \\
    Geo. self-calib.~\cite{5981781,8115403} & \cellcolor[rgb]{0.5,1.0,0.5}yes & \cellcolor[rgb]{0.5,1.0,0.5}yes & \cellcolor[rgb]{0.5,1.0,0.5}yes & \cellcolor[rgb]{1.0,0.5,0.5}no & \cellcolor[rgb]{1.0,0.5,0.5}no \\
    Col. calib. + SL~\cite{naive_compensate, tps} & \cellcolor[rgb]{0.5,1.0,0.5}yes & \cellcolor[rgb]{0.5,1.0,0.5}yes & \cellcolor[rgb]{1.0,0.5,0.5}no & \cellcolor[rgb]{0.5,1.0,0.5}yes & \cellcolor[rgb]{1.0,0.5,0.5}no \\
    DNN~\cite{compensnet_pp,deprocams,diff_projector} & \cellcolor[rgb]{0.5,1.0,0.5}yes & \cellcolor[rgb]{1.0,0.5,0.5}no & \cellcolor[rgb]{1.0,0.5,0.5}no & \cellcolor[rgb]{0.5,1.0,0.5}yes & \cellcolor[rgb]{1.0,0.5,0.5}no \\
    Ours & \cellcolor[rgb]{0.5,1.0,0.5}yes & \cellcolor[rgb]{0.5,1.0,0.5}yes & \cellcolor[rgb]{0.5,1.0,0.5}yes & \cellcolor[rgb]{0.5,1.0,0.5}yes & \cellcolor[rgb]{0.5,1.0,0.5}yes \\
    \hline
    \multicolumn{6}{p{8cm}}{(SL) structured light. (Geo. self-calib.) geometric self-calibration. (Col. calib.) color calibration. (DNN) Joint optimization using deep neural networks.}
    \end{tabular}
\end{table}

We leverage their findings and introduce a new directed light component, namely a projector, into a reflectance field, that can be ``self-calibrated''. Despite having an unknown location and projection parameters, we exploit the fact projected textures are known and of high-resolution, and are projected from a new vantage point relative to camera positions. This enables new applications for the field of projection mapping, as previous studies optimize for a single viewpoint. We summarize the projection image correction techniques including ours in \cref{table:proj_rel_work} to better visualize the advantage of our method. In addition, we demonstrate that such a direct light source can yield superior scene and material understanding and serve many new applications that were not possible before such as multi-view text to projection.

\section{Method}
\label{sec:method}

Our approach to projection mapping using neural fields is depicted in \cref{fig:overview}. High quality projection mapping from novel views requires obtaining the geometry of the scene, its materials, and projector and camera properties. In our setting, a single uncalibrated projector shines onto the scene from a fixed point, and the scene is captured from a sparse set of viewpoints ($\sim100$ in our case) by a hand held camera while the projector shines different patterns given as input. For camera parameters, we use a structure from motion (SfM) algorithm \cite{colmap1, colmap2} to register and extract them (unless otherwise mentioned). See \cref{sec:data} for more details. The captured photos are used to train three neural networks, constituting the neural reflectance field. $N^{geo}$ predicts density and normals, $N^{mat}$ predicts BRDF parameters, and $N^{tr}$ allows the real-time approximation of occlusion between two points \cite{nerv}. The material and light transport models we use, including the incorporation of occlusions, are described in \cref{sec:preliminary} and \cref{sec:light_transport}. After a three-step training process (\cref{sec:optimize}, \cref{sec:optimization_schedule}), the networks are trained to evaluate geometry and material, and projector orientation and colors are calibrated. This allows for differentiably evaluating scene appearance with arbitrary projector illumination and from arbitrary view points, opening new and exciting opportunities for downstream tasks. See three example applications \cref{sec:applications}.

\subsection{Preliminary}
\label{sec:preliminary}
Our method uses a continuous absorption-reflection model, where every 3D location $x\in \mathbb{R}^3$ is described by a local density $\sigma(x)$ and attributes relating to its reflection properties $A(x)$. We leverage a differentiable volumetric rendering scheme \cite{nerf}, where an image from a certain viewpoint is produced by shooting rays into the scene that are sampled over and then integrating attributes collected over each sample along the rays. More concretely, a pixel $P$ in the final image of a certain attribute from a certain view point is produced by shooting a ray $R$ (defined by the cameras center of projection and $P$'s 3D location) into the scene with $s_i$ samples along it. The value for a specific scene attribute $A$ is computed using \cref{eq:vol_render} (e.g. color, normals, depth, etc.).

\begin{equation}
    P(A, R) = \sum_{s_i\in \mathbb{R}} W(s_i)\cdot A(s_i)
    \label{eq:vol_render}
\end{equation}
Where $W(s_i)$ are weights that are proportional to local density and transmittance computed using \cref{eq:weight_calc} and $s_0, s_1, ..., s_i$ are samples along a ray $R$. In the following, $\delta_i$ are the distances between the samples, and $\tau(s_i)$ is the transmittance, the product of densities along the ray.

\begin{align}
\alpha(s_i) &\triangleq 1-e^{-\delta_i \cdot clip(\sigma(s_i))} \nonumber \\
\tau(s_i) &\triangleq \prod_{j=0}^{j=i-1} 1 - \alpha(s_j) \nonumber \\
W(s_i) &=  \alpha(s_i) \cdot \tau(s_i)
\label{eq:weight_calc}
\end{align}
Another interpretation of $W(s_i)$ in \cref{eq:weight_calc} is the probability a ray will get attenuated in a certain sample $s_i$.

Additionally, by leveraging the reciprocity nature of light, we model a projector as an inverse pinhole camera. Each projector pixel behaves exactly the same as a camera pixel, only light radiates out from the image plane, instead of into it. This model allows us to reuse all of the mechanics describing pinhole cameras in our scene - the projector is parametrized by its extrinsics, intrinsics, projected image and color transformation parameters. In other words, if a projector with resolution $n \times m$, intrinsics $K\in \mathbb{R}^{3\times3}$ and extrinsics $T\in \mathbb{R}^{4\times3}$, projects pattern $I\in \mathbb{R}^{n\times m\times3}$, a point $P_w = (Px, Py, Pz, 1)$ in homogeneous world coordinates has incident irradiance $L'$ from pixel $P_P$ that is calculated using \cref{eq:light_transport}. 

\begin{align}
P_p &= KTP_w \nonumber \\ 
L'_{proj}(P_w) &= G_p \cdot (I_{P_p})^\gamma
\label{eq:light_transport}
\end{align}
Where $\gamma$ is a coefficient modeling the nonlinear behavior of light emitted by a projector, $G_p$ is an arbitrary multiplicative coefficient determining the gain and $I_{P_p}$ is the projected image $I$ in pixel $P_p$. Note that strictly speaking there should also be an inverse squared distance term, but in practice the projector was placed far enough from the scene such that this term is roughly constant for all objects in the scene and swallowed by the gain $G_p$.

\subsection{Light transport}
\label{sec:light_transport}
The purpose of this section is to expand on the calculation of $A(s_i)$ in \cref{eq:vol_render} for shading a sample (i.e., determining the color). Consider a point $P_w$ in world coordinates. In general, its color, as perceived by a viewer at location $P_v$ can be described by the light it receives $L(P_w)$ (irradiance), its local material properties $M(P_w)$, and the geometrical relationship between $P_w$ and $P_v$. Our light transport model assumes light arriving at any particular point is a superposition of direct illumination, and indirect illumination.

\subsubsection{Direct illumination}
For an empty scene with no density, if a projector shines some known pattern, then a point $P_w$ has incident irradiance that can be computed using \cref{eq:light_transport}. Note that if the point is outside the projector's frustum, the irradiance due to direct illumination is zero.

\begin{figure}[t]
  \centering
  \includegraphics[width=0.9\hsize]{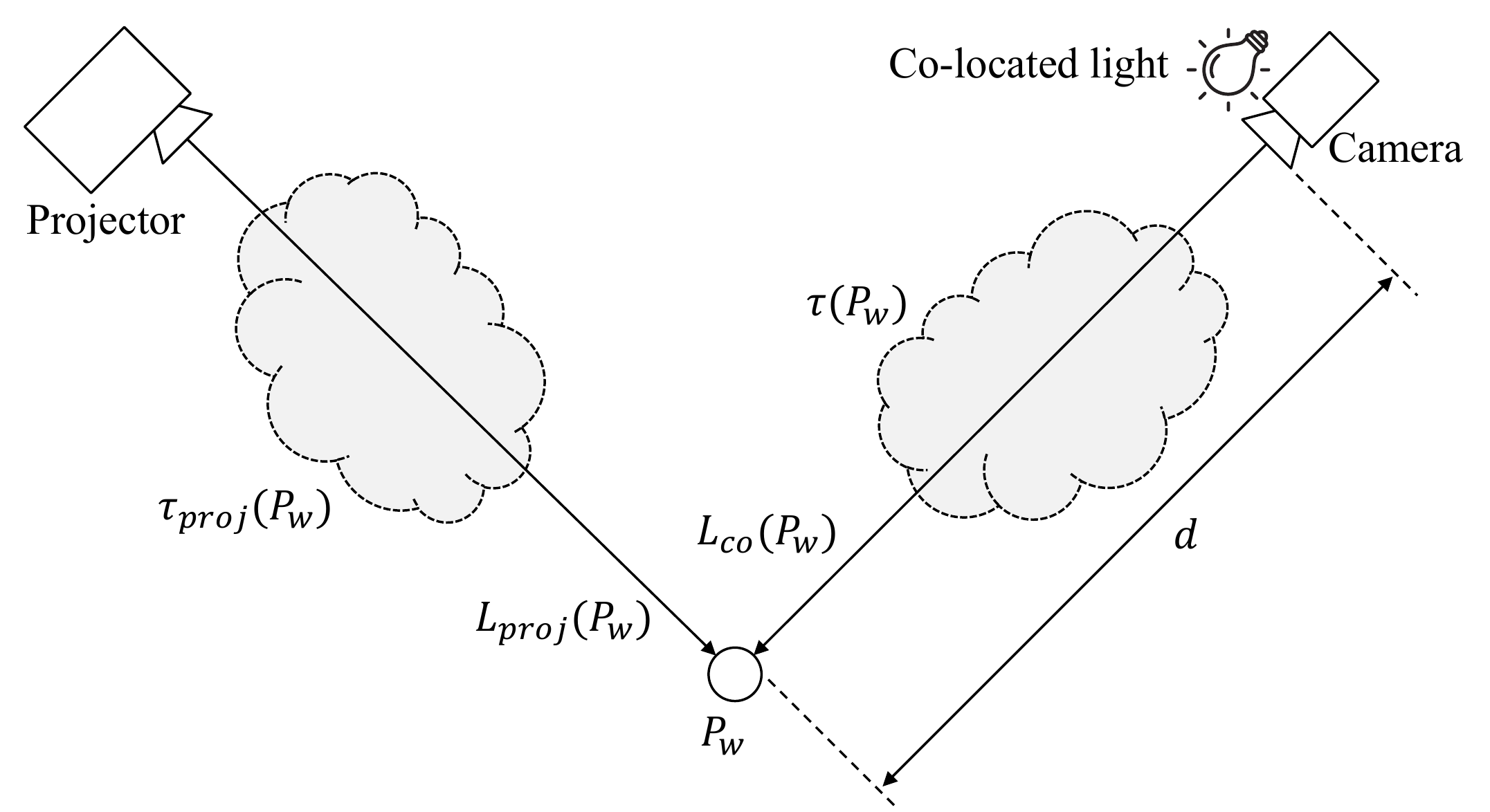}
  \caption{A scene point $P_w$ is illuminated by two types of direct illuminations, a projector and a point light source co-located with the capturing camera.}
  \label{fig:direct_illumination}
\end{figure}

Any real scene, however, does contain density and the actual light that reaches a certain location depends on the transmittance with regards to the projector. If the transmittance $\tau$ with respect to the projector for location $P_w$ is known, the actual irradiance received is attenuated by $L_{proj}({P_w}) = L'_{proj}({P_w}) \cdot \tau_{proj}({P_w})$ (See \cref{fig:direct_illumination}). 
To obtain the transmittance value $\tau_{proj}({P_w})$, we can use the definition in \cref{eq:weight_calc} for $\tau$ by casting a ray from $P_w$ towards the projector.

We will need to evaluate $L_{proj}$ for many samples $P_w$ along rays shot from known cameras and therefore need to compute $\tau_{proj}(P_w)$ for all such samples. However, na\"{i}vely doing so is very computationally intensive, so we leverage the technique introduced by NeRV \cite{nerv} to predict $\tau_{proj}$ at any location and from any view point using a neural network, which is trained to yield transmittance values as close as possible to the ones obtained from samples along primary rays cast into the scene. Furthermore, we found that estimating $\tau_{proj}$ using a network induces a smoothness prior to the visibility values, which reduces pixelation artifacts in the final result. Similar findings were described by Zhang et al. \cite{nerfactor}.

\paragraph{Co-located light.} In addition to a projector, we placed a white LED light co-located with the camera \cite{nerfaa}. Since they are co-located, the irradiance received by a point $P_w$ from this light source follows the inverse square law: $L_{co}(P_w) = G_{co} \cdot \tau(P_w) \cdot 1/d^2$ where $G_{co}$ is the gain associated with the light source, $d$ is the distance between the light source (i.e. camera) and $P_w$, and $\tau$ is the transmittance (\cref{fig:direct_illumination}). Note that $\tau$ for this particular light source is simply the one computed for the primary ray (\cref{eq:weight_calc}), and is reused to save computational time.

\subsubsection{Indirect illumination}
Up until this point we assumed light received at any given point is solely due to direct illumination. However, in real scenes light interacts with the objects in the scene and the surroundings as inter-reflections by bouncing until it gets completely absorbed by the environment. We make the simplifying assumption that all these global illumination effects are explained by the surface albedo used as part of the material model (described below). This assumption works sufficiently well when the direct light irradiating the scene produced by the projector is significantly stronger than any other light interaction. We also collect our data in a relatively dark room, allowing us to ignore other sources of light which further validates this assumption.

\subsubsection{Material}
We chose to define the final color of $P_w$ using an analytical BRDF \cite{microfacet}. This material model assumes dielectric materials, and greatly simplifies the search space for plausible solutions to explain the observed images. The BRDF function describes the emitted light towards any direction given an incoming light direction and local material properties, which in our case are: local alebdo $a \in \mathbb{R}^3$, local normal $n \in \mathbb{R}^3$ and local roughness $\rho \in \mathbb{R}^1$. Using this function, given in a closed form in the supplementary, we are able to quantify light emitted toward any viewer direction given an incoming light direction and the aforementioned local material properties, which are all predicted using neural networks.

\subsection{Data acquisition}
\label{sec:data}
As a reminder, the input for our method are raw captured images and their respected camera extrinsics (as well as camera intrinsics). For real scenes, we obtain data by capturing them from different view points, and projecting multiple patterns from the projector per view. For synthetic scenes, we add more challenge by capturing views from a ``video-like'' camera path, that simulates a video taken continuously. In this scenario, the projector projects a single pattern per view. This was done in an attempt to demonstrate an easy acquisition process, and to emphasize how biased view points yield artifacts in the reconstruction without a projector (\cref{fig:roughness}).

The choice of which patterns to project during acquisition is a significant degree of freedom. Clearly, a good set of patterns should allow best material and geometry reconstruction, as well as projector calibration. One significant constraint is that any strong augmentation to the appearance of the scene prevents SfM algorithms from registering those frames and extracting the camera extrinsics.

We found this to be a problem because when a scene has unknown viewpoints together with an unknown projector the optimization almost never converged. On the other hand, strong augmentations with high dynamic range help extract better geometry and materials, as well as calibrate the projector more easily and find gamma coefficient for correct color transformation. We eventually settled for projecting random ``lollipop'' like patterns that can be seen on the right.

\begin{wrapfigure}{r}{0.10\textwidth}
    \centering
    \includegraphics[width=0.10\textwidth]{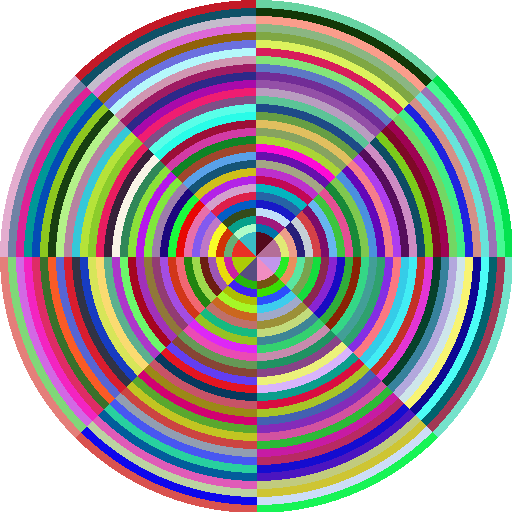}
\end{wrapfigure}

\begin{figure}[t]
  \centering
  \includegraphics[width=0.98\hsize]{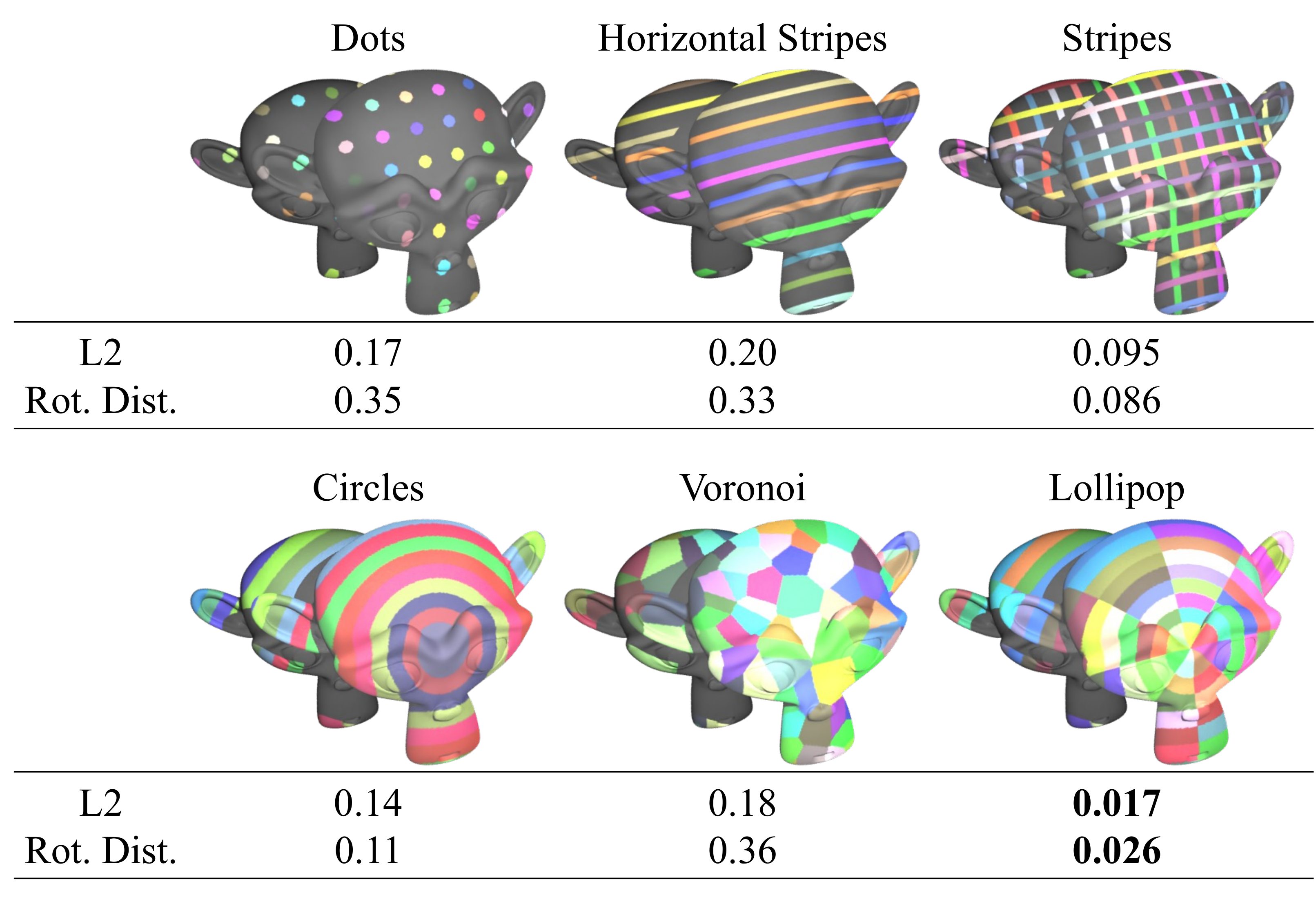}
  \caption{Projection pattern evaluation. We ran the \textbf{Projector} step of the optimization (networks are frozen) with different types of patterns being projected onto a synthetic scene. The L2 distance is measured between the ground truth projector location and the result after optimization, and the Rotation Distance is measured as $\arccos(\frac{tr(R)-1}{2})$ where $R = PQ^T$ is the rotation matrix obtained by multiplication of the ground truth projector rotation and the optimization result. The lollipop patterns consistently yielded robust calibration.}
  \label{fig:patterns}
\end{figure}

These patterns yielded the best results in terms of geometric calibration (see \cref{fig:patterns}), largely because the center of the patterns encouraged the projector to drop into a single global minimum, and because they are densely packed (preventing getting stuck on other local minima). In addition they break the circular symmetry associated with simple concentric circles. For real scenes, we project 3 different patterns per view: flood filled white, flood filled black, and a lollipop pattern of random colors. The idea behind the flood filled white and black patterns was to easily allow registration to occur for extrinsics estimation (since every 3 such patterns share a view), and also for scheduling the optimizing as described in the next section. We used COLMAP \cite{colmap1, colmap2}. To further supervise the optimization, we also automatically mask the foreground for all input images using a pretrained U2-Net \cite{u2net} on the DUTS-TR dataset \cite{duts}. For synthetic scenes, since the projector projects a single pattern per view, we opted for iterating between the aforementioned patterns. As mentioned before, lollipop textured views (those for which the projector is augmenting the scene) are very hard to register to each-other and to other views using a standard SfM algorithm because the scene appearance is extremely different. To counter this, we use the flood-filled black views for the purpose of extracting the viewpoints (COLMAP) and interpolate the rest of the views. Then, we optimize the cameras extrinsics along with the rest of the system similarly to how the projector extrinsics are optimized. We note that this only worked if the interpolated views are close enough to the ground truth.

\subsection{Optimization objective}
\label{sec:optimize}
By defining the light transport model we can now describe the full optimization objective formally. Given $N$ images from different viewpoints and their camera parameters, projected with $M$ (known) different patterns by an unknown projector $Proj(K, T, \gamma, G_p)$, the goal is to minimize the image loss (\cref{eq:img_loss}) over samples $s_i \in \mathbb{R}^3$ generated by rays $R \in Cam$ from all cameras pixels, where $C(R)$ is the accumulated color along the ray $R$, $C_{GT}(R)$ is the ground truth captured color and $R_2$ is a secondary ray defined between $s_i$ and the (unknown) projector's location:

\begin{align}
C(s_i) &= (L_{proj}(s_i) + L_{co}(s_i)) \cdot BRDF_{a, n, \rho, R, R_2}(s_i)  \nonumber \\ 
C(R) &= \sum_{s_i\in R} W(s_i)\cdot C(s_i)\nonumber \\ 
Loss_{img} &= \sum_{R \in Cam}||C(R) - C_{GT}(R)||_{2}^{2}
\label{eq:img_loss}
\end{align}

We use one neural network $N_{\phi}^{geo}$ for predicting $(W, n)$, another one $N_{\phi}^{mat}$ for $(\rho, a)$ and a third one $N_{\phi}^{tr}$ for $\tau$, so the parameters of the optimization are $\theta = \{\phi_{geo}, \phi_{mat}, \phi_{tr}, K, Rt, \gamma, G_p, G_{co}\}$. 
We use the following loss term to enable the transmittance network to predict values close to the primary rays transmittance:
\begin{align}
Loss_{tr} &= \sum_{R} \sum_{s_i \in R}||N_{\phi}^{tr}(s_i) - \tau(s_i)||_2^2
\label{eq:vis_loss}
\end{align}

To encourage predicted normals to orient themselves similarly to density distribution and to not face backwards from the camera, we regularize them using the normals derived by differentiating the density values $\sigma$ with respect to $s_i$ (\cref{eq:normal_loss}, $Loss_{n_{1}}$) and penalize them if they are oriented away from the primary rays ($Loss_{n_{2}}$) \cite{refnerf}:
\begin{align}
\hat{n}(s_i) &= \frac{\nabla_{\sigma} s_i}{||\nabla_{\sigma}(s_i)||} \nonumber \\
Loss_{n_{1}} &= \sum_{R}\sum_{s_i \in R}W(s_i)\cdot||n(s_i) - \hat{n}(s_i)||_2^2 \nonumber \\
Loss_{n_{2}} &= \sum_{R}\max((\sum_{s_i \in R}W(s_i)\cdot n(s_i))\cdot R_{dir}, 0) \nonumber \\
Loss_n &= Loss_{n_{1}} + Loss_{n_{2}}
\label{eq:normal_loss}
\end{align}

Additionally, since we are capturing only scenes with solids, we add a "fog loss", penalizing solutions which consist of densities that are not either 0 or 1. This prevents explaining any visible signal as fog (densities of non solids). We do this by imposing a parabolic function over the transmittance of all samples (i.e. transmittance is encouraged to describe a solid or empty space):

\begin{align}
Loss_{fog} &= -b \sum_{s_i} \tau(s_i)^2 - \tau(s_i)
\label{eq:fog_loss}
\end{align}
where $b$ is a hyper parameter controlling the parabolic curve.

\begin{figure*}[ht]
\setlength{\tabcolsep}{1pt}
\centering
\begin{tabular}{ccccc}
Ground Truth & NeRF \cite{nerf} & NeRF AA \cite{nerfaa} & Ours & \begin{tabular}[c]{@{}c@{}} 16 out of 36 \\novel views and projections \end{tabular}\\
 \includegraphics[width=0.16\textwidth, trim={0 7cm 0 0},clip]{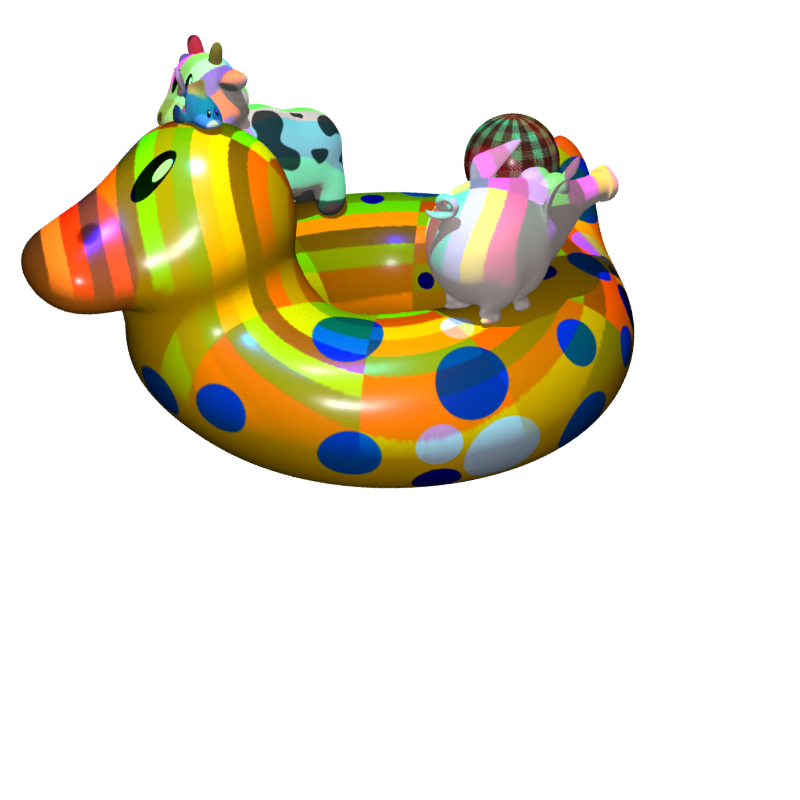} & \includegraphics[width=0.16\textwidth, trim={0 3.5cm 0 0},clip]{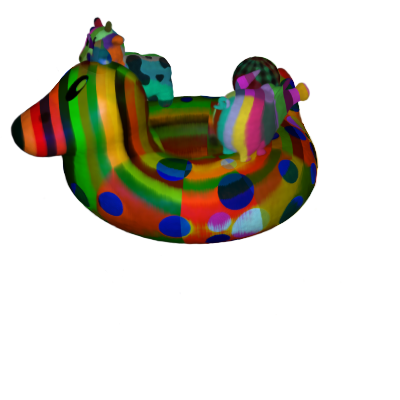} & \includegraphics[width=0.16\textwidth, trim={0 3.5cm 0 0},clip]{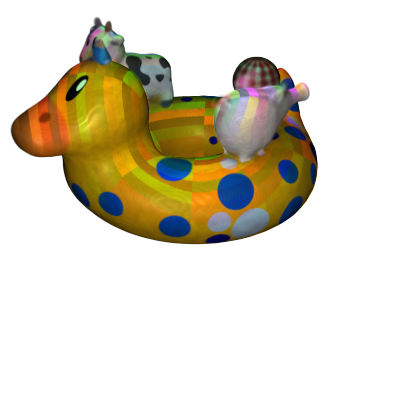} & \includegraphics[width=0.16\textwidth, trim={0 3.5cm 0 0},clip]{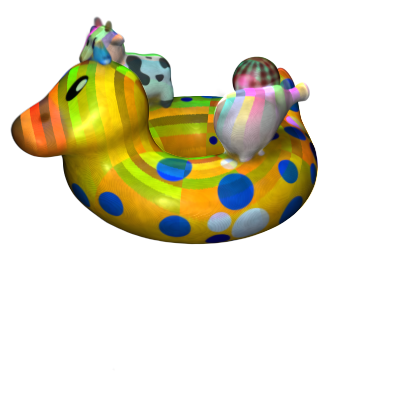} & \includegraphics[width=0.16\textwidth]{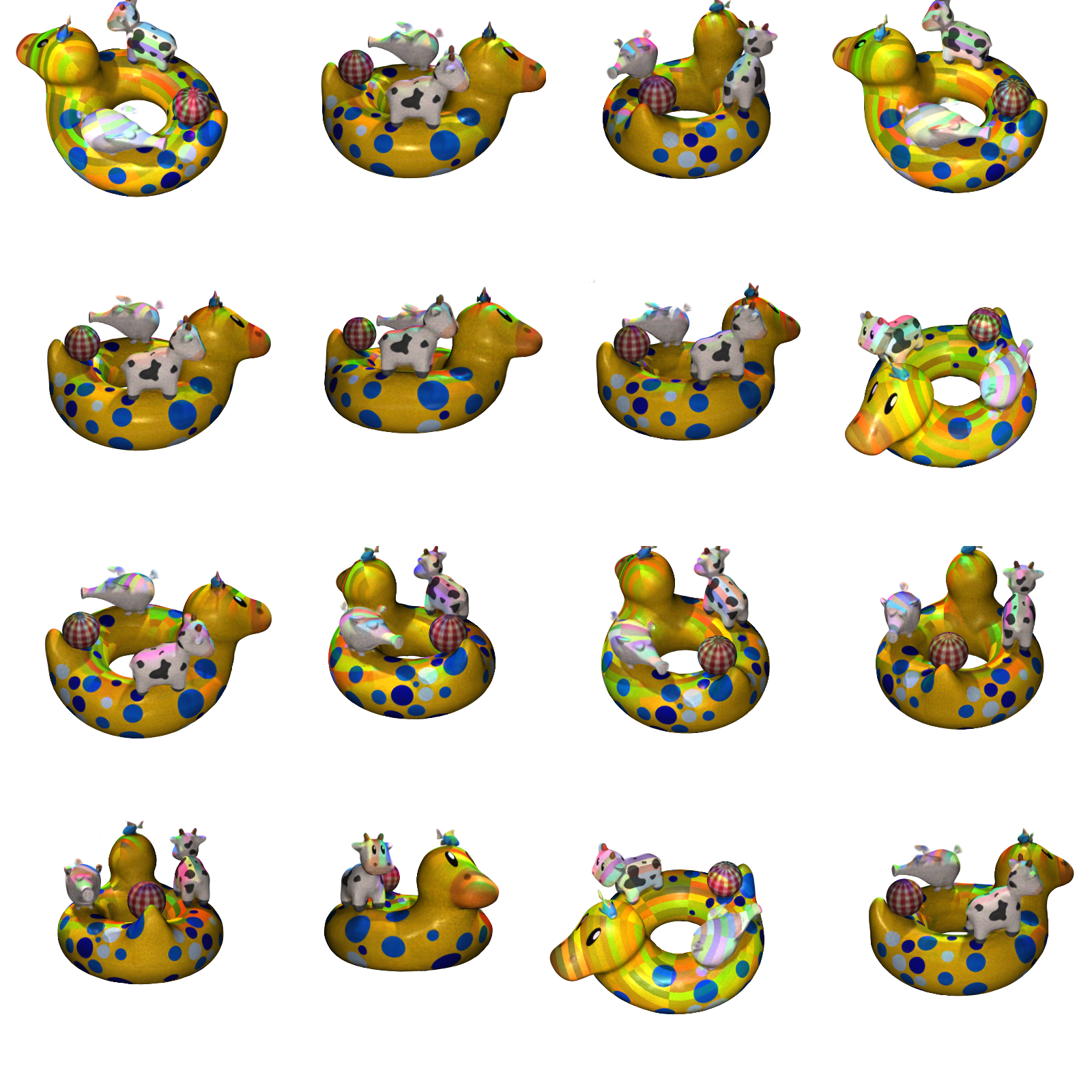}\\
  \includegraphics[width=0.16\textwidth]{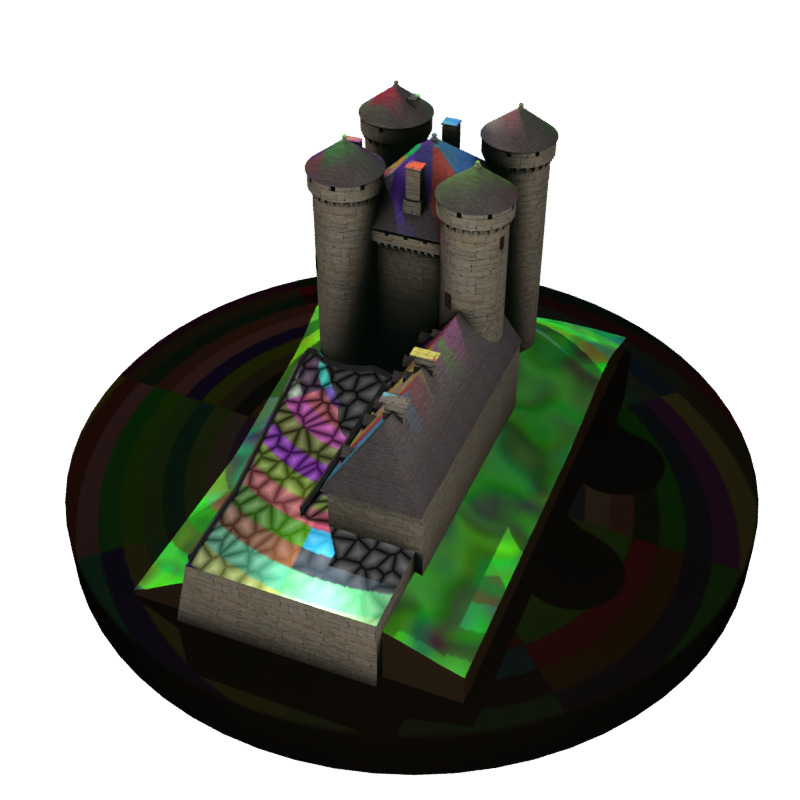} & \includegraphics[width=0.16\textwidth]{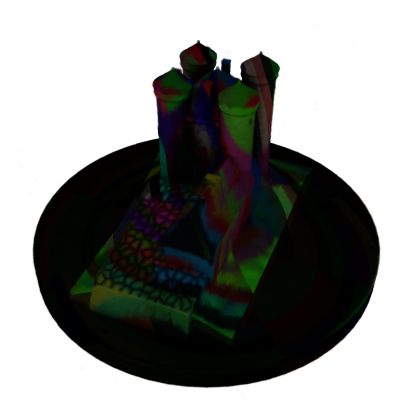} & \includegraphics[width=0.16\textwidth]{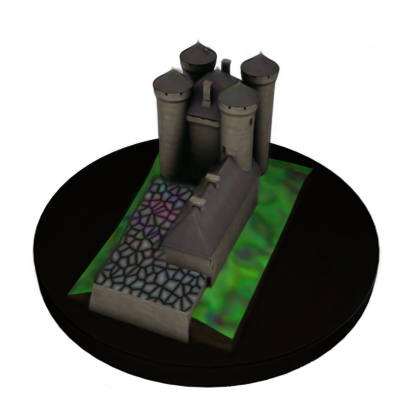} & \includegraphics[width=0.16\textwidth]{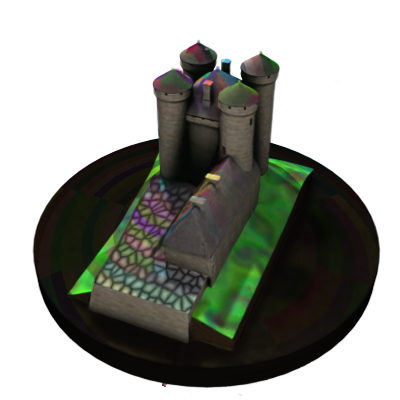} & \includegraphics[width=0.16\textwidth]{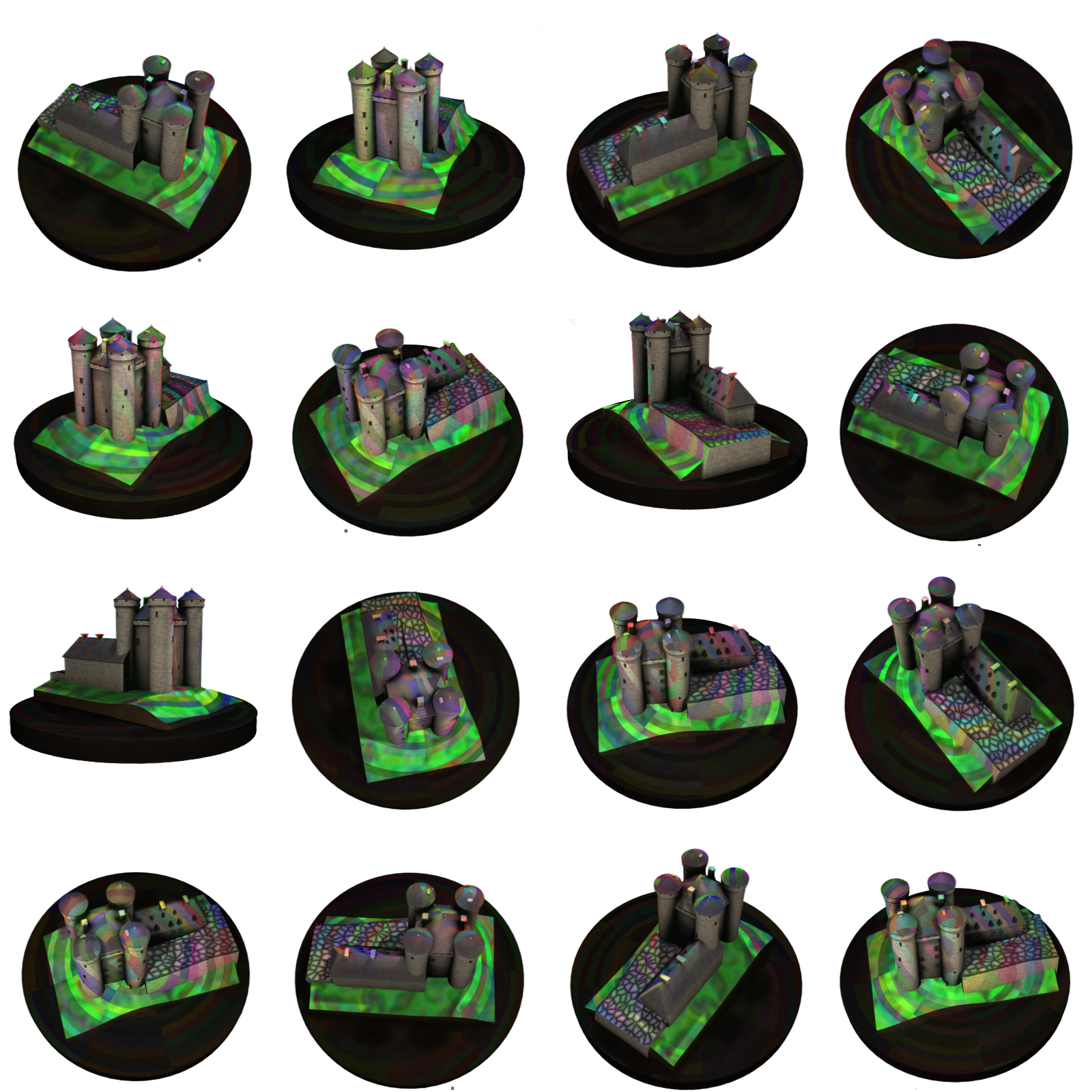}\\
    \includegraphics[width=0.16\textwidth]{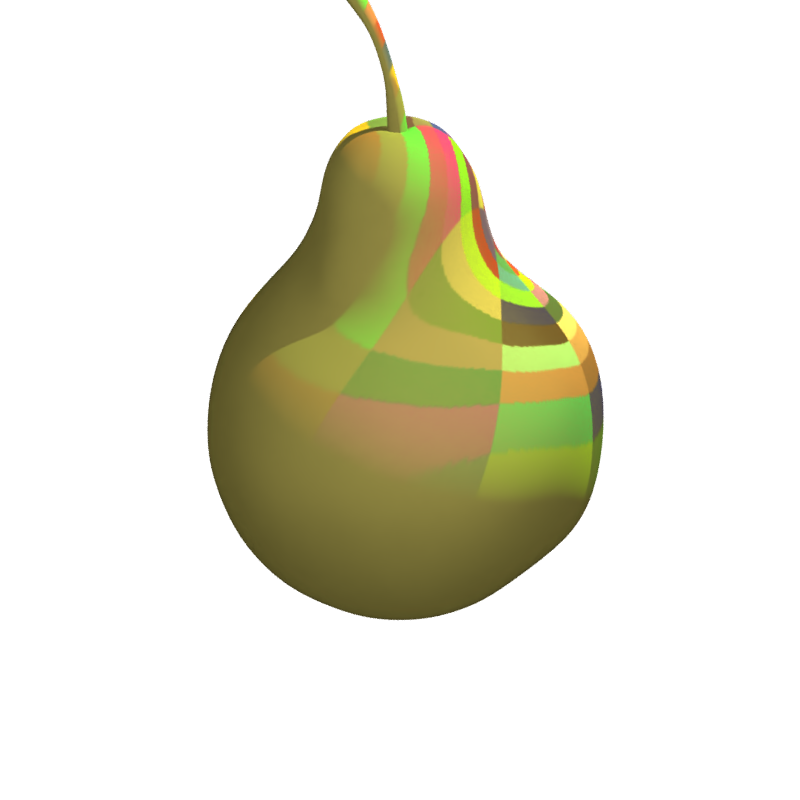} & \includegraphics[width=0.16\textwidth]{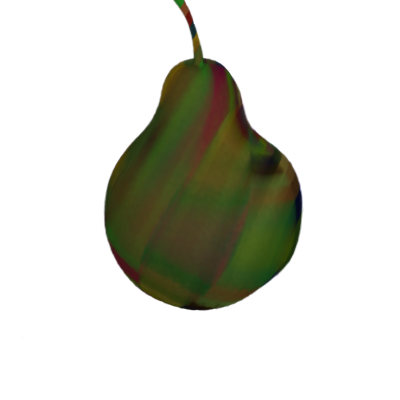} & \includegraphics[width=0.16\textwidth]{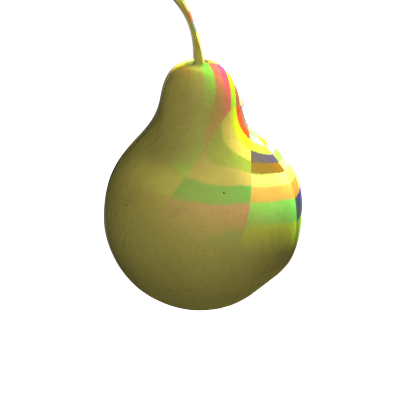} & \includegraphics[width=0.16\textwidth]{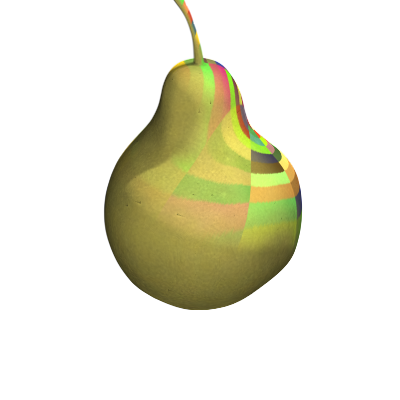} & \includegraphics[width=0.16\textwidth]{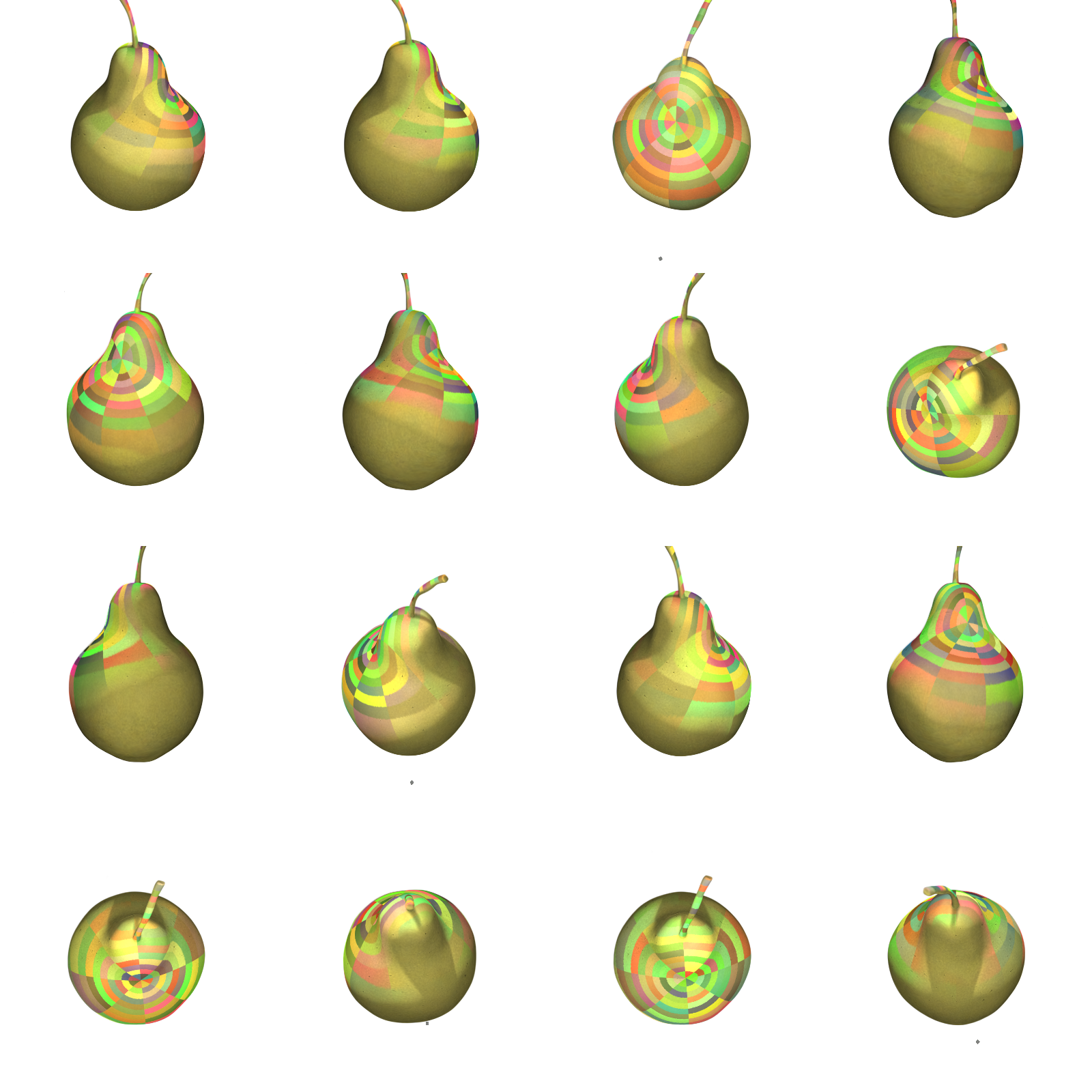}\\
 \hline
 \multicolumn{5}{c}{Average PSNR over 36 novel views and novel projections} \\
\hline
 - & 17.0 & 23.5 & 27.8 & -
 
\end{tabular}

\caption{Synthetic scene results. Each image was generated by multiplying the projector's irradiance and the surface geometry and material inferred by the networks of NeRF, NeRF AA, and ours. Since NeRF and NeRF AA do not train the projector's intrinsics and extrinsics, the ground truth parameters are injected in the computation. The right column shows 16 different views out of 36 novel views generated by our method.}
\label{fig:synth_results}
\end{figure*}

\subsection{Optimization scheduling}
\label{sec:optimization_schedule}
The strong dependency between the projectors parameters and the appearances of the object mean that optimizing them jointly negatively influences the result. This is because when the projector is misplaced or ill-oriented (a typical initialization scenario), the geometry is deformed (and materials are distorted) to comply with the target appearance (\cref{eq:img_loss}), until the projector reaches its correct orientation. Even worse, while the geometry is deforming, calibrating the projector is highly challenging since this process relies on correct visual cues to move the projector towards the correct place. Indeed, we found that jointly optimizing for all parameters did not converge to a good solution under all experimental setups. To resolve this issue, we split the optimization into 3 individual steps:
\begin{itemize}
  \item \textbf{Scene:} where we first optimize an accelerated version of NeRF \cite{nerfacc} over black flood-filled views, and use the resulting occupancy grid to only optimize $N_{\phi}^{geo}$, $N_{\phi}^{mat}$ and $N_{\phi}^{tr}$ (since the projector parameters are unknown at this point) using the all loss terms described in \cref{sec:optimize}.
  \item \textbf{Projector:} where we use $Loss_{img}$ to optimize projector parameters (and cameras in synthetic scenes) over all views.
  \item \textbf{Fine-tune:} where we refine all parameters over all views using all loss terms.
\end{itemize}
The first step allows coarse material and geometry to be represented, the second step calibrates the projector using this representation, which succeeds now that geometry is roughly correct. The last step uses this calibration to improve results. For further detail about the training schedule, see supplementary material.

\section{Results}

\begin{figure}[t]
\centering
    \includegraphics[width=0.98\hsize]{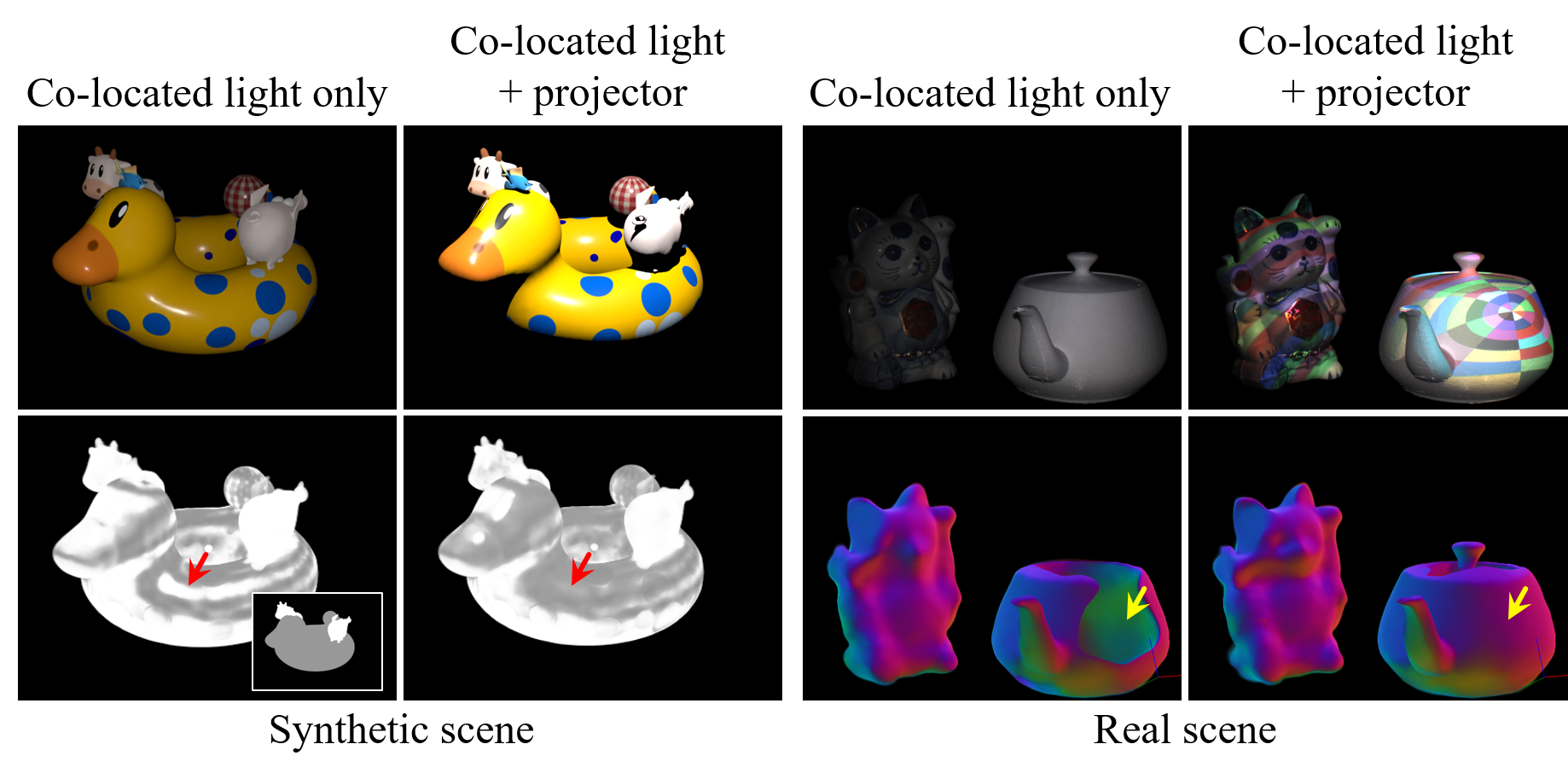}
  \caption{Material and geometry estimation. Top: two different scenes (one synthetic, and one real) are shined with either only colocated light, or both colocated light and projector patterns. Bottom: After running a full optimization of the scene, we perform a decomposition of the roughness (left) and normals (right), for either only colocated light viewpoints, or when adding the projector. Notice how using a projector significantly improves both the roughness estimation (red arrow) and the normals for any area exposed to projector illumination (yellow arrow). Inset in the bottom-left image contains ground truth roughness where higher pixel values are more shiny areas.}
  \label{fig:roughness}
\end{figure}

\begin{figure*}[h]
\setlength{\tabcolsep}{1pt}
\centering
\begin{tabular}{ccccccc}
Albebdo & Normals & Roughness & Transmittance & $L_{proj}$ & Final image & Ground truth\\
  \includegraphics[width=0.14\textwidth]{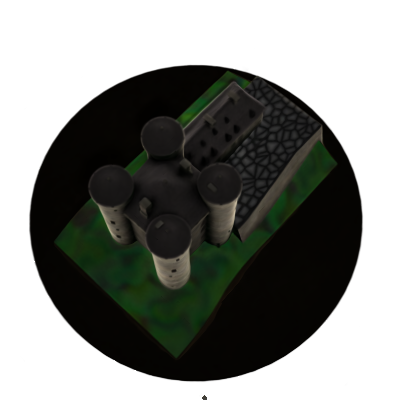} & \includegraphics[width=0.14\textwidth]{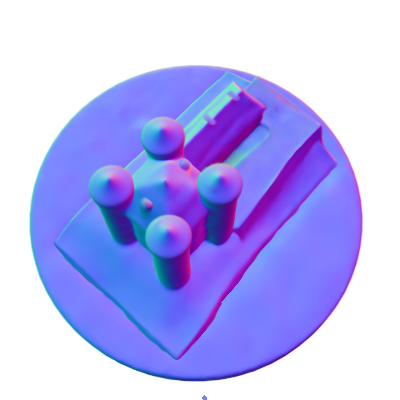} & \includegraphics[width=0.14\textwidth]{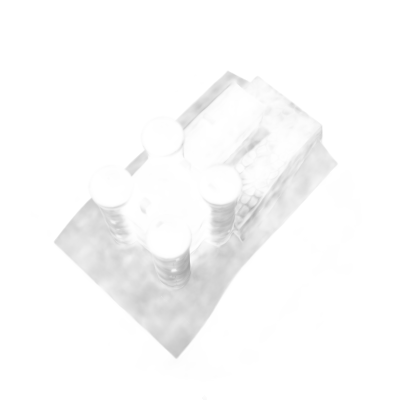} & \includegraphics[width=0.14\textwidth]{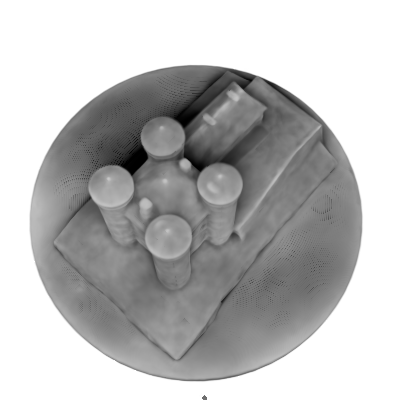} & \includegraphics[width=0.14\textwidth]{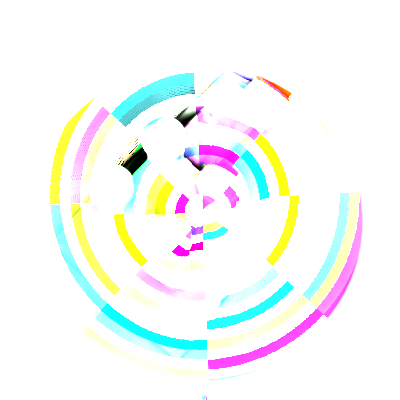} & \includegraphics[width=0.14\textwidth]{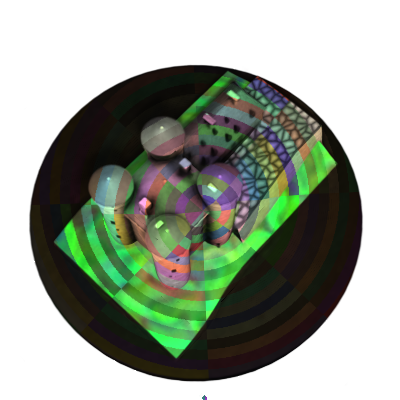} &
  \includegraphics[width=0.14\textwidth]{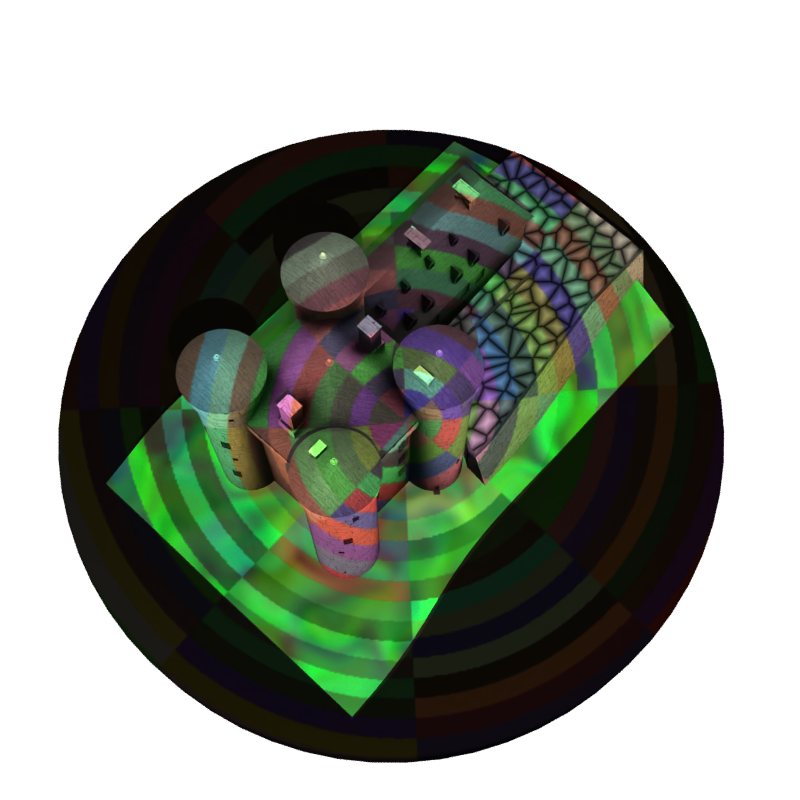}\\
 \includegraphics[width=0.14\textwidth]{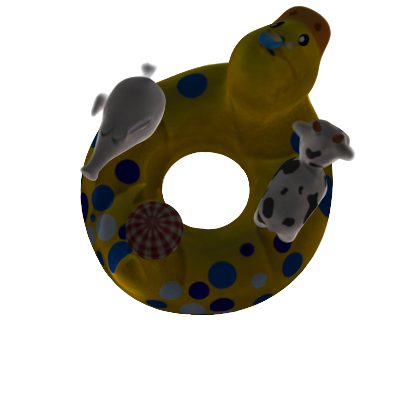} & \includegraphics[width=0.14\textwidth]{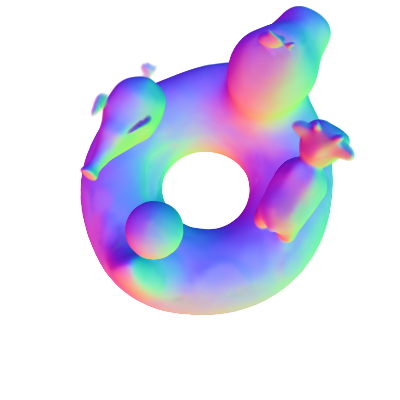} & \includegraphics[width=0.14\textwidth]{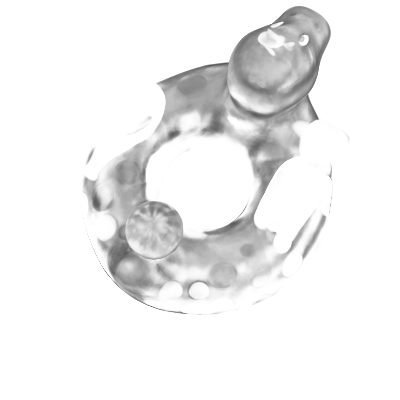} & \includegraphics[width=0.14\textwidth]{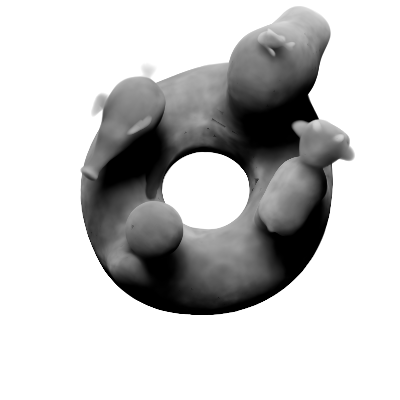} & \includegraphics[width=0.14\textwidth]{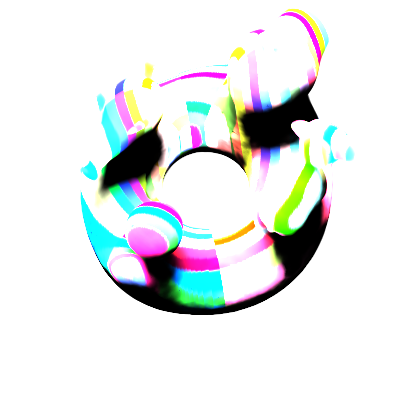} & \includegraphics[width=0.14\textwidth]{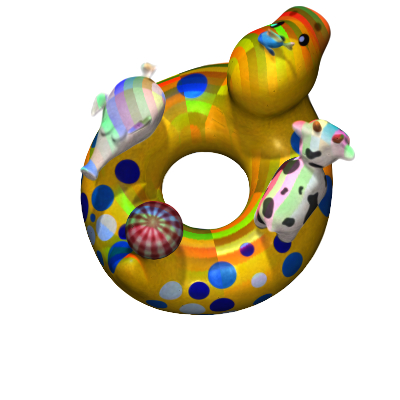}&
 \includegraphics[width=0.14\textwidth]{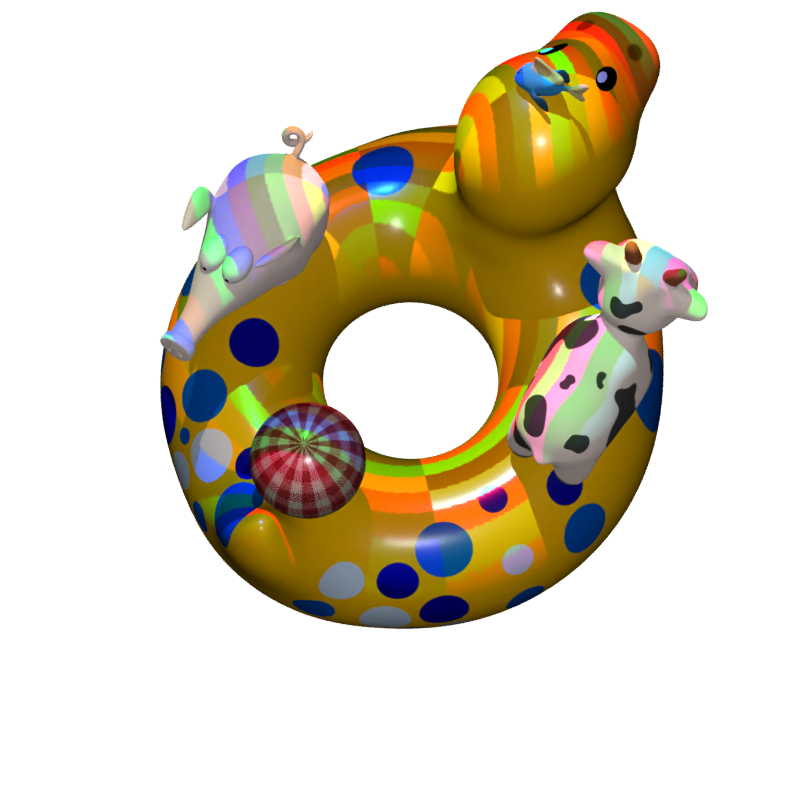}\\ 
 \hline
 \includegraphics[width=0.14\textwidth]{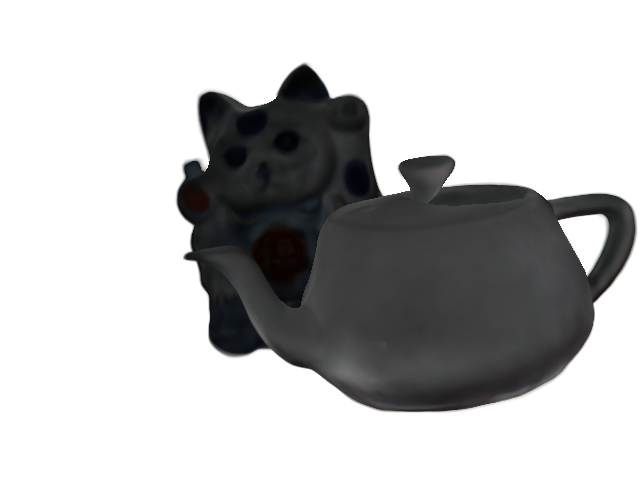} & \includegraphics[width=0.14\textwidth]{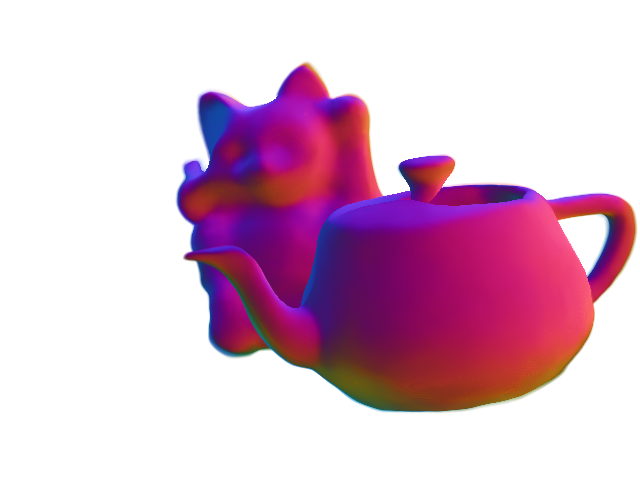} & \includegraphics[width=0.14\textwidth]{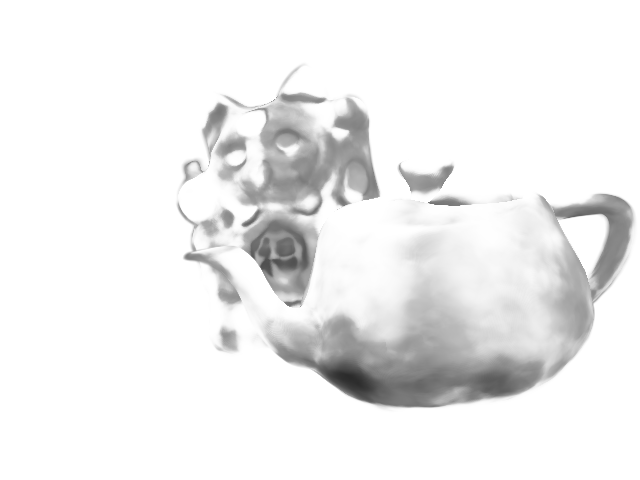} & \includegraphics[width=0.14\textwidth]{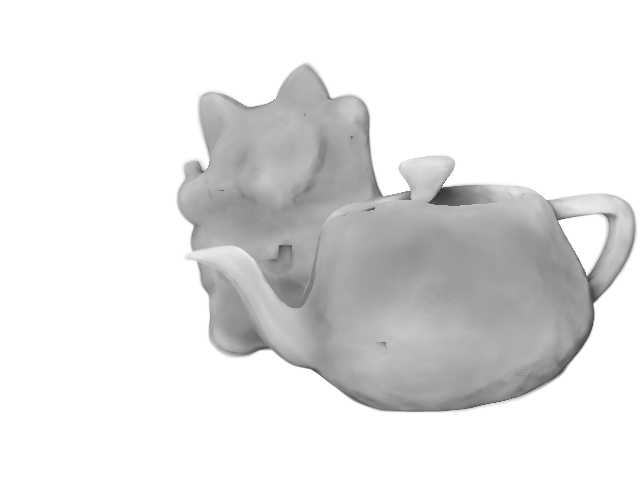} & \includegraphics[width=0.14\textwidth]{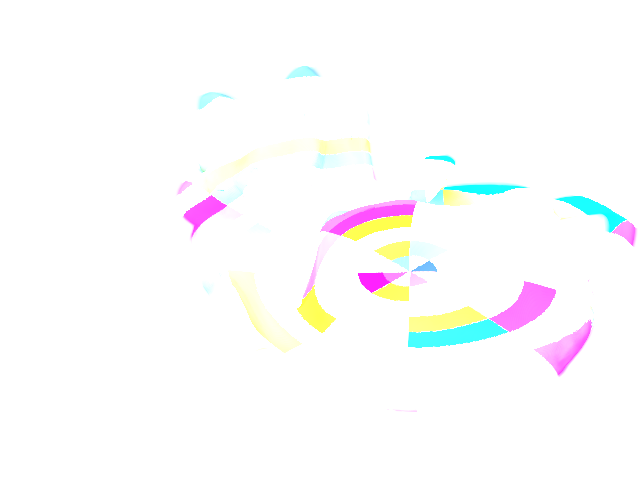} & \includegraphics[width=0.14\textwidth]{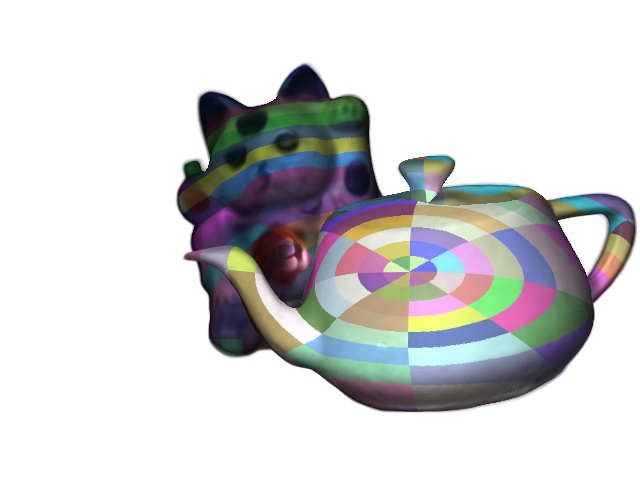} &
\includegraphics[width=0.14\textwidth]{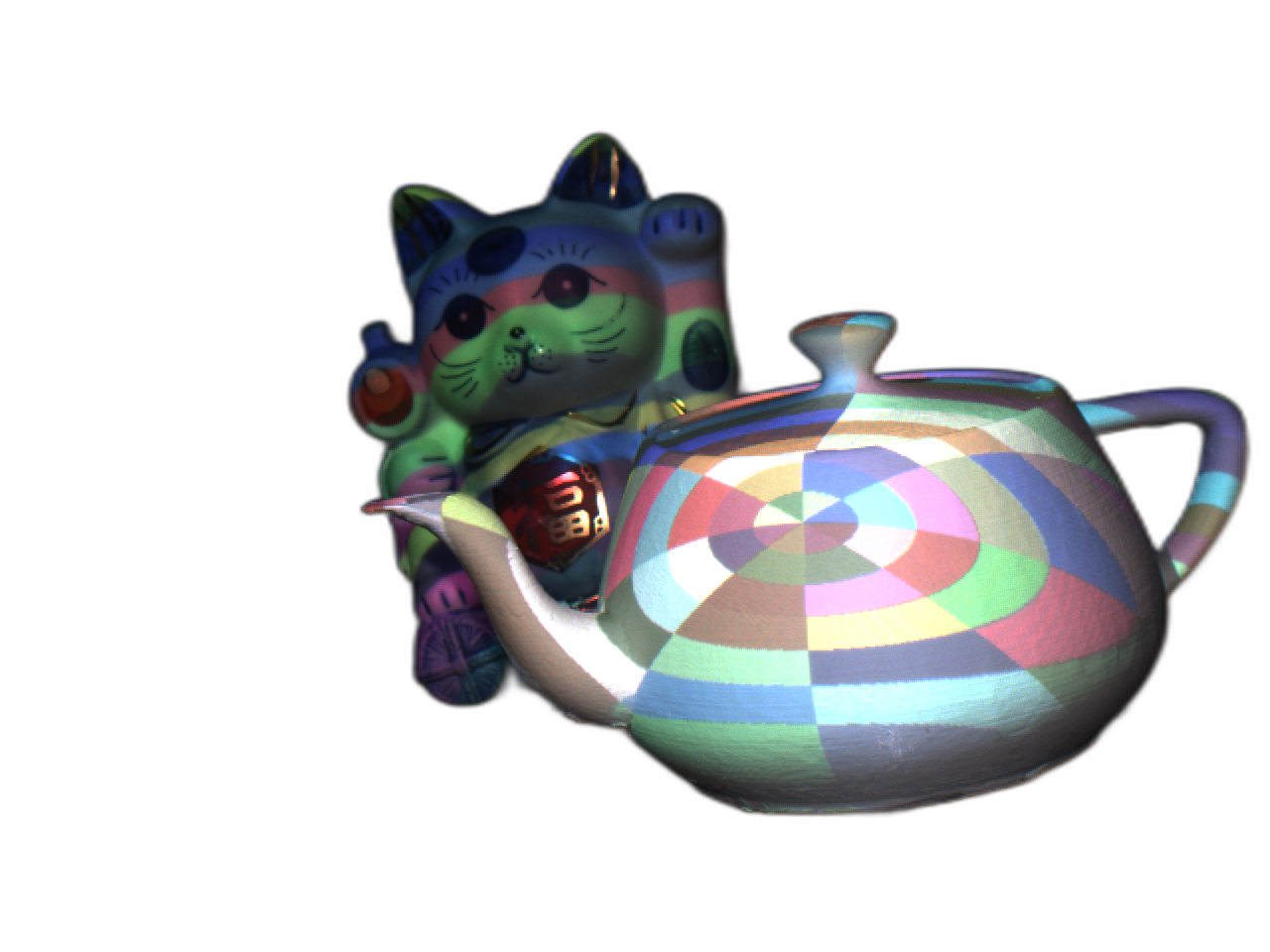}\\
\includegraphics[width=0.14\textwidth]{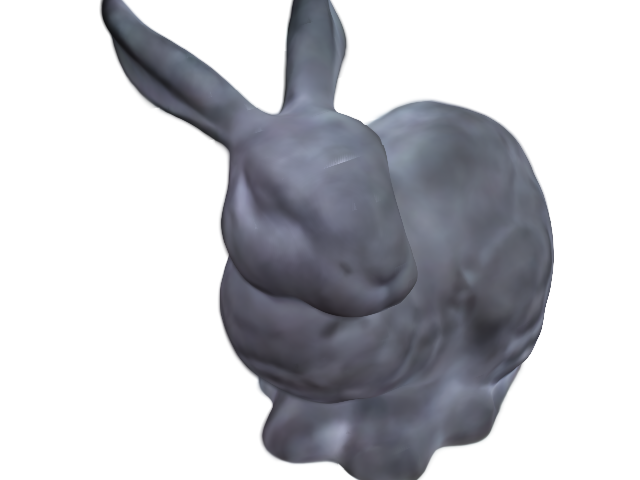} & \includegraphics[width=0.14\textwidth]{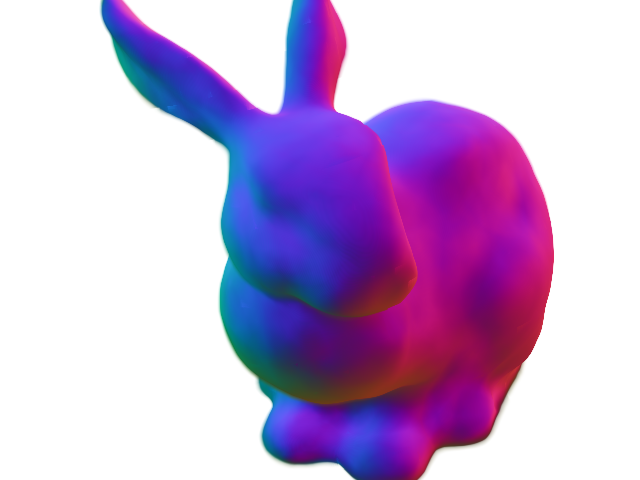} & \includegraphics[width=0.14\textwidth]{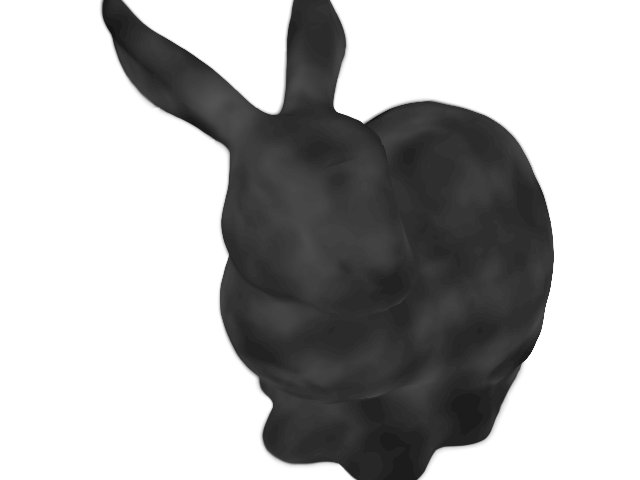} & \includegraphics[width=0.14\textwidth]{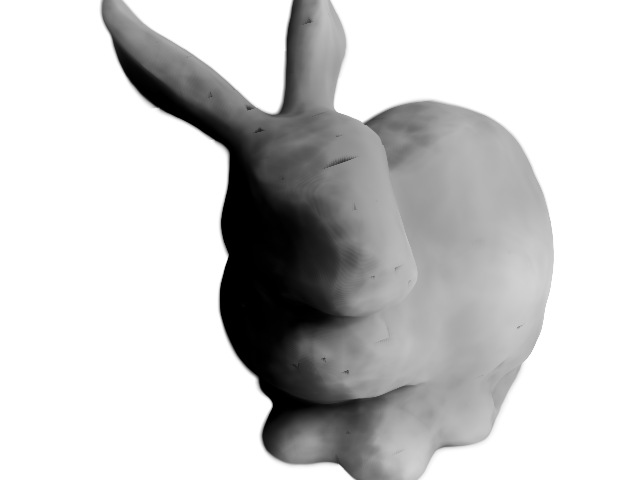} & \includegraphics[width=0.14\textwidth]{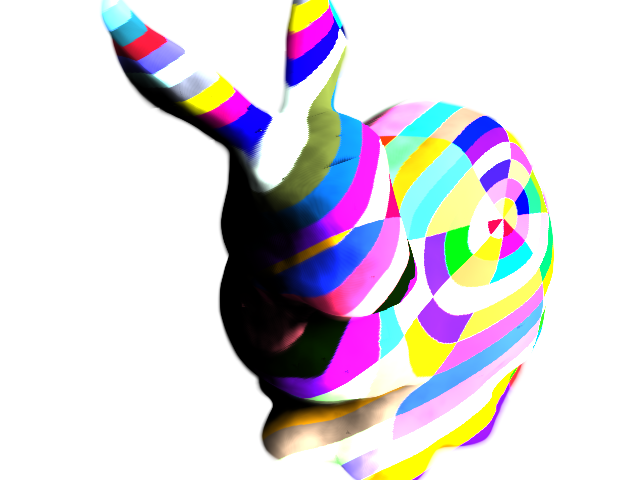} & \includegraphics[width=0.14\textwidth]{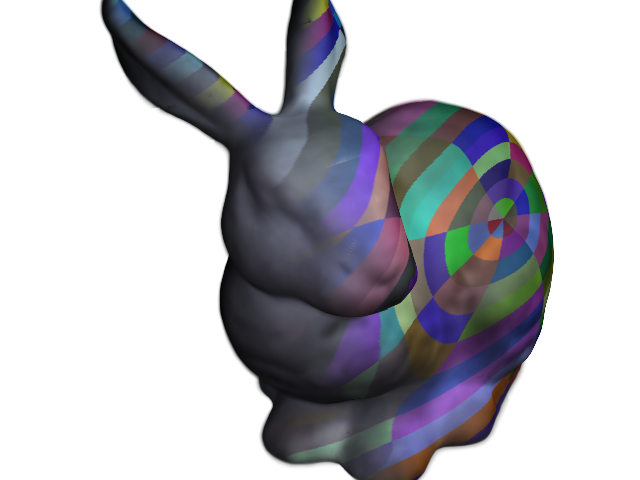} &
\includegraphics[width=0.14\textwidth]{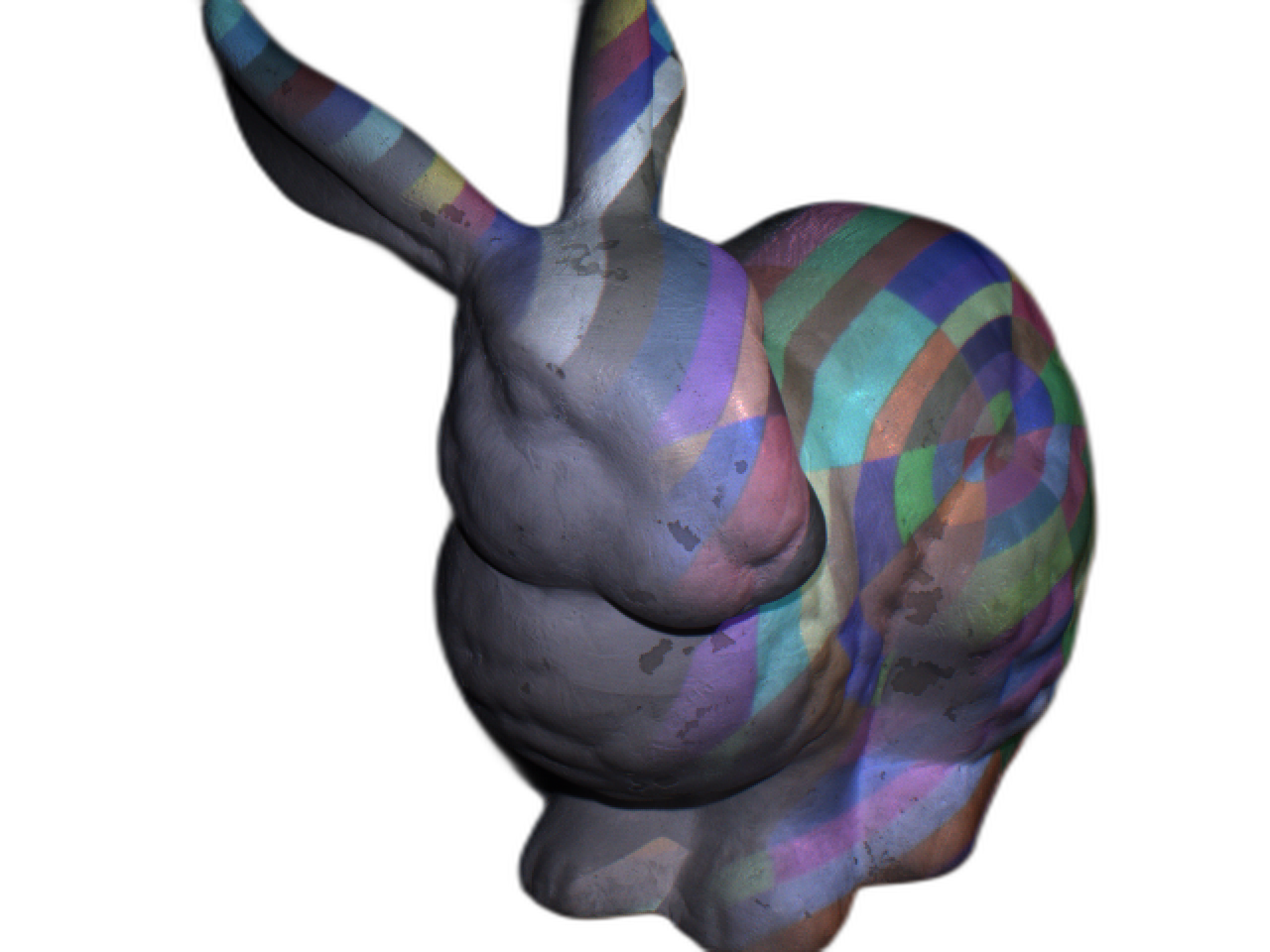}\\

\end{tabular}

\caption{Decomposition. Top 2 tows: synthetic scenes. Bottom two rows: real scenes. Our method decomposes the scene into interpretable quantities which are passed into a Macrofacet BRDF function \cite{microfacet} to form the final image.}
\label{fig:decomposition}
\end{figure*}

\subsection{Synthetic scenes}
We performed quantitative evaluation of our method over 3 synthetic scenes created in Blender: Zoo, Castle and Pear, and compared to two baselines, NeRF \cite{nerf} and NeRF AA \cite{nerfaa}, results can be seen in \cref{fig:synth_results}. We captured 297 images of each scene from different viewpoints using a camera with a co-located point light. All methods were trained on the same views and tested on the same novel views. In our method, the scene was illuminated by a synthetic projector either with a white flood-filled texture, a black flood-filled texture, or a random lollipop pattern. Then, using the trained networks, we synthesized each scene under projection of new lollipop patterns from 36 novel viewpoints. All training and inference were performed using a RTX A6000 Nvidia GPU. Training took roughly 3 hours per scene, while inference took 2 to 10 seconds per view depending on the complexity of the observed geoemtry (for a 400x400 resolution image).

NeRF was trained on the scene with the co-located light on, and the projector was injected in inference using ground truth projector parameters. It fails to explain the scene because of its underlying absorption-emission model, which does not account for different lighting or shadows, and wrong normals. NeRF AA was also trained in this manner and explains the scene better because it consists of an absorption-reflection model. Our method achieves higher reconstruction quality through better material and geometry estimations, even though our projector parameters were not the ground truth ones but optimized along with the network training. The improvements are quantitatively measured using PSNR values ($10*log(\frac{max(Im)}{Loss_{img}})$). Importantly, optimizing jointly with a projector not only improves results in terms of material and geometry, but allows automatic calibration of the projector to occur.

\cref{fig:roughness} depicts the advantage of our method in material and geometry estimation. In particular, roughness estimation is highly dependent on where the specular lobe was seen during training, and for a co-located light setup is limited to the camera viewpoints used. Note improving material estimation cannot be achieved by merely placing another point light source instead of the projector, since its position is unknown and cannot be resolved easily, as opposed to the projector which self-calibrates.
On top of this, the projector shines light from a different vantage point, which means the relationships between view, light and surface normal directions are better captured for the BRDF calculations, since the network is forced to explain much wider angels of the specular lobe. For the geometry improvement, a hole was covered up only with a projector being optimized as well, as the co-located light views were explained by the geometry network as other side of the teapot. We note that that geometry and material did not improve if the projector is not calibrated and in-fact their estimation deteriorates. For a decomposition of synthetic scenes, see \cref{fig:decomposition}, top.

\subsection{Real scenes}
We captured 2 real scenes (Bunny, Teapot and Maneki-neko) using a projector (EPSON EH-TW5350, 1920$\times$1080) and an RGB camera (Point Grey FL3-U3-13S2C-CS, 1280$\times$960) in a dark room. We captured 306 images from 102 different viewpoints using the camera with a co-located LED light (\cref{fig:teaser}). For each viewpoint, we illuminated the scene using the projector with a white flood-filled texture, a black flood-filled texture, and a random lollipop pattern. Using the captured images, we extracted the camera parameters using COLMAP \cite{colmap1} and trained our networks. The decomposition of these scenes can be seen in \cref{fig:decomposition}, bottom. The challenge associated with real scenes compared to synthetic ones are the unaccounted for effects that arise due to global illumination, equipment noise, and optical effects such as defocus blur and finite aperture size. Despite this, we manage to extract decent scene representations.

\begin{figure}[t]
\setlength{\tabcolsep}{1pt}
\centering
\begin{tabular}{cccc}
 \includegraphics[width=0.12\textwidth]{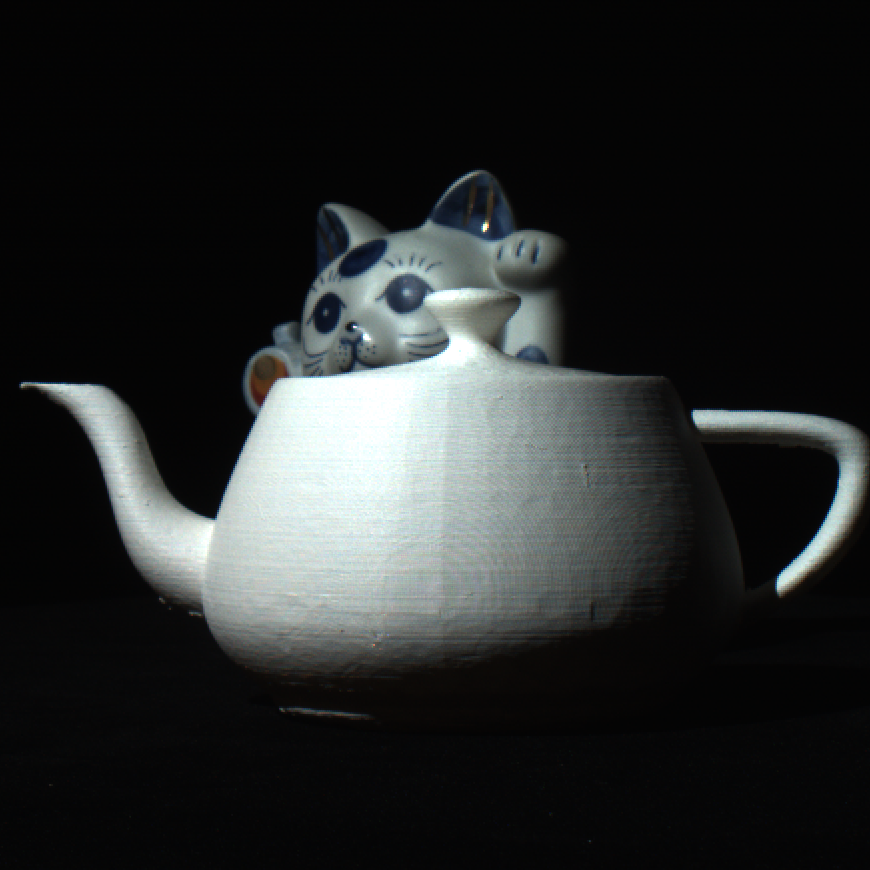} &
  \includegraphics[width=0.12\textwidth]{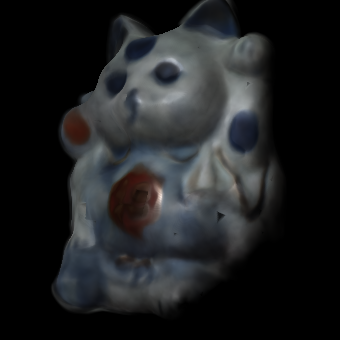} &
  \includegraphics[width=0.12\textwidth]{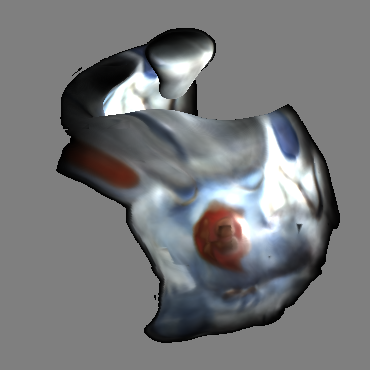} &
  \includegraphics[width=0.12\textwidth]{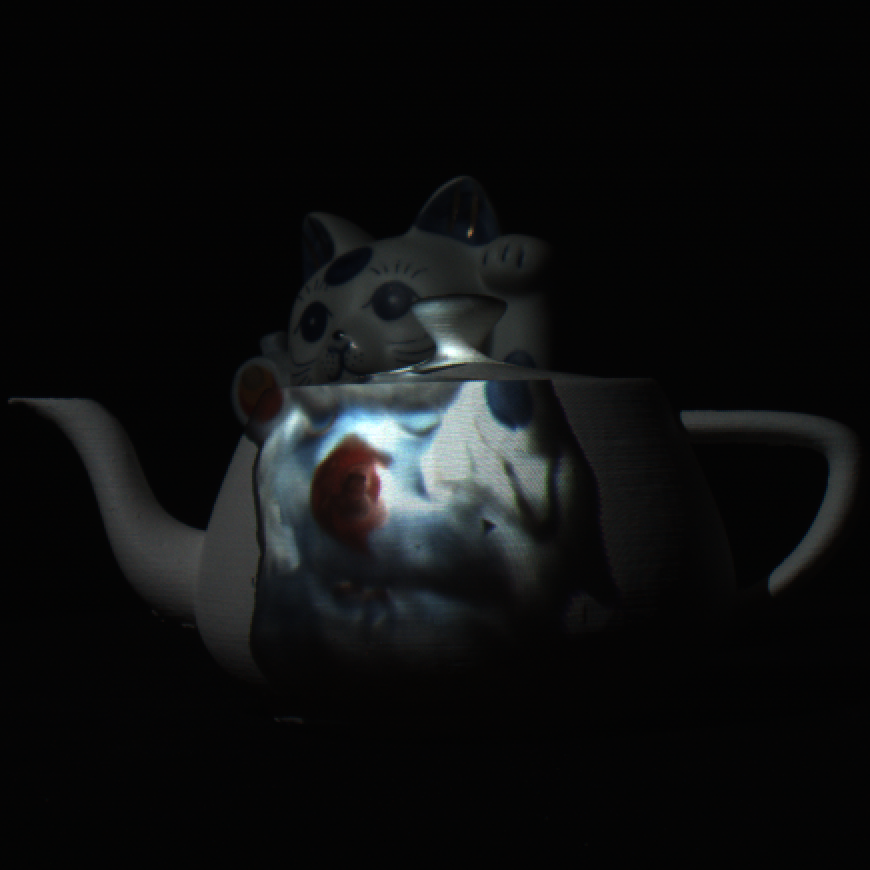}
  \end{tabular}
  \caption{XRAY. From left to right: The scene is shown with flood filled white being projected; A desired appearance is obtained by placing a virtual camera beyond an occluder (the teapot) and rendering; Using a 2-pass rendering technique, we obtain a projection image such that the desired view is achieved (without optimization); The result is reprojected onto the real scene, creating the illusion of a transparent occluder.}
  \label{fig:xray}
\end{figure}

\subsection{Applications}
\label{sec:applications}
\begin{figure*}[ht]
\setlength{\tabcolsep}{1pt}
\centering
\begin{tabular}{cccccc}
\begin{tabular}[c]{@{}c@{}}Desired \\Appearance \end{tabular} & \begin{tabular}[c]{@{}c@{}}SL Calib. + \\Gamma Cor. + \\Col. Comp. \cite{naive_compensate} \end{tabular} & \begin{tabular}[c]{@{}c@{}} No \\Calib. \end{tabular} & \begin{tabular}[c]{@{}c@{}} No \\Comp. \end{tabular} & CompenNeSt++ \cite{compensnet_pp} & Ours \\
 \includegraphics[width=0.16\textwidth]{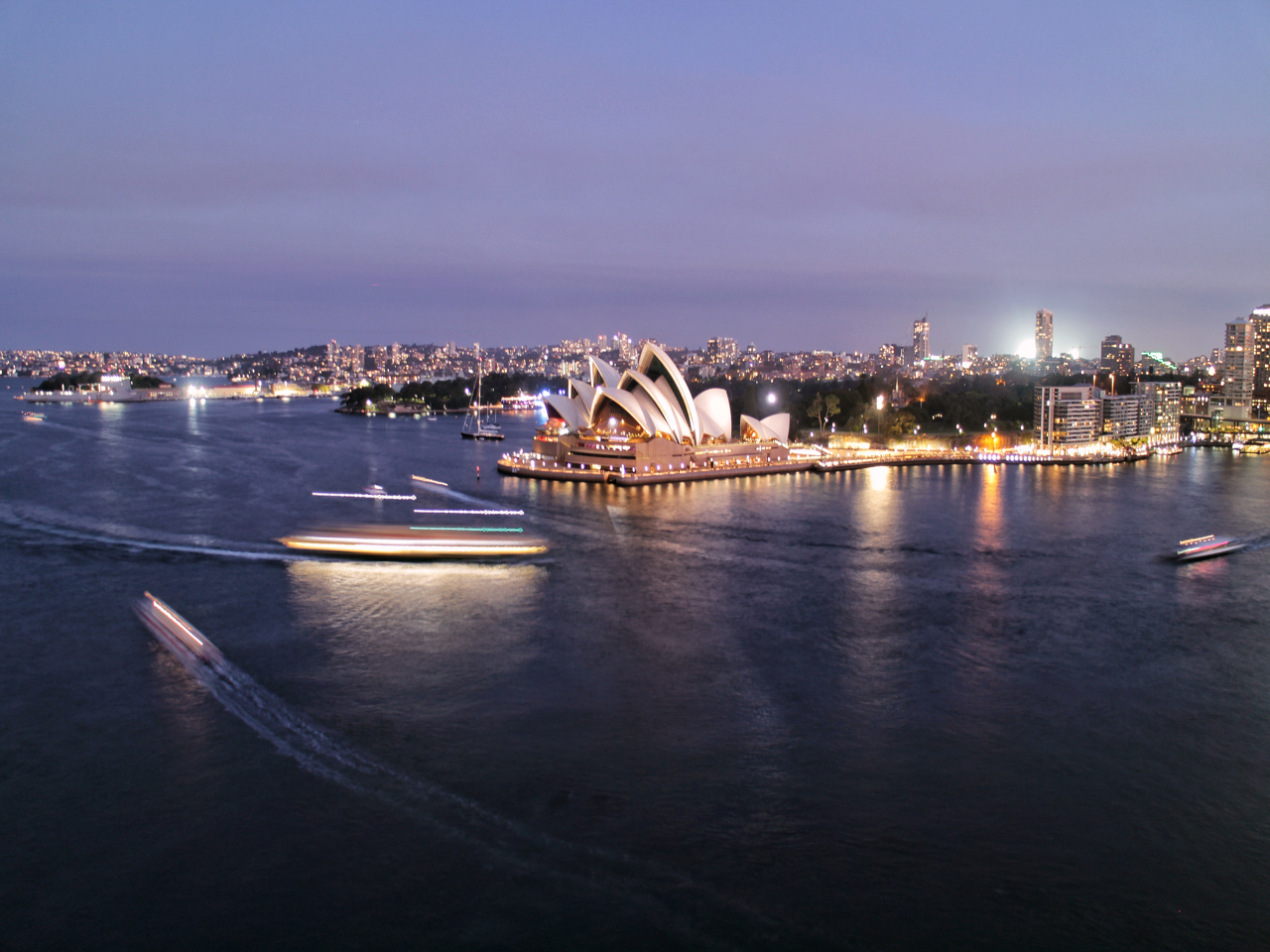} & \includegraphics[width=0.16\textwidth]{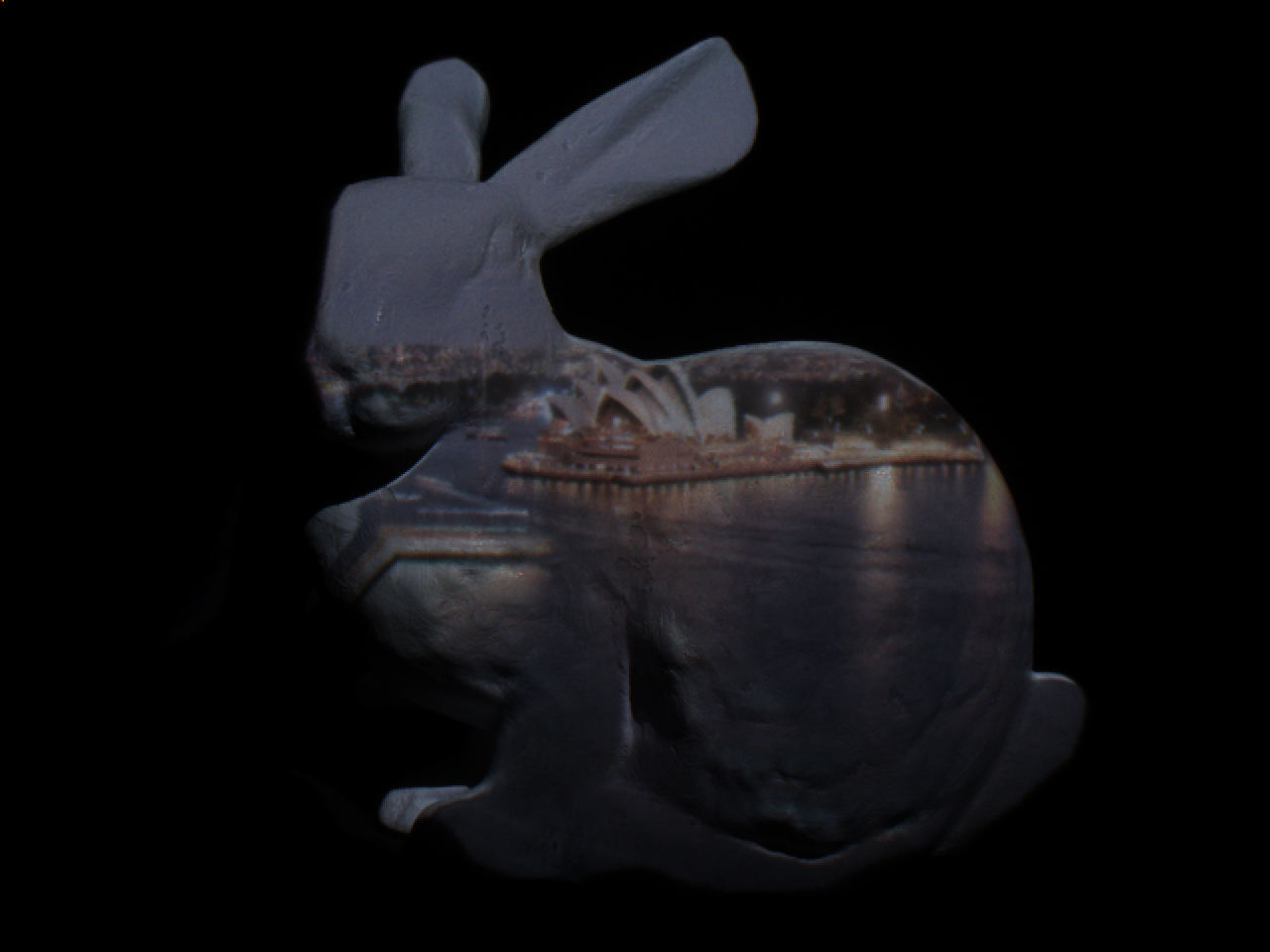} & \includegraphics[width=0.16\textwidth]{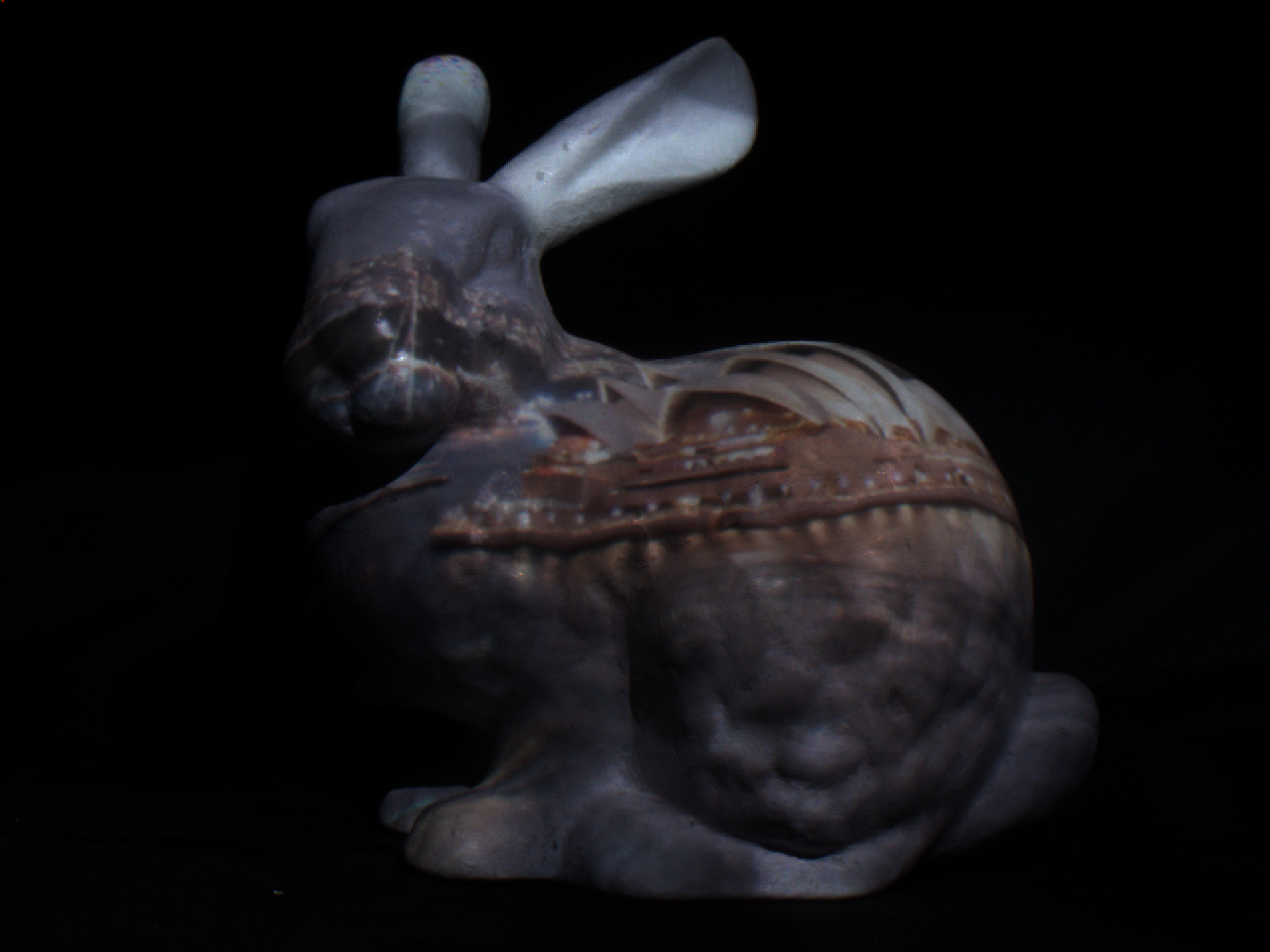} & \includegraphics[width=0.16\textwidth]{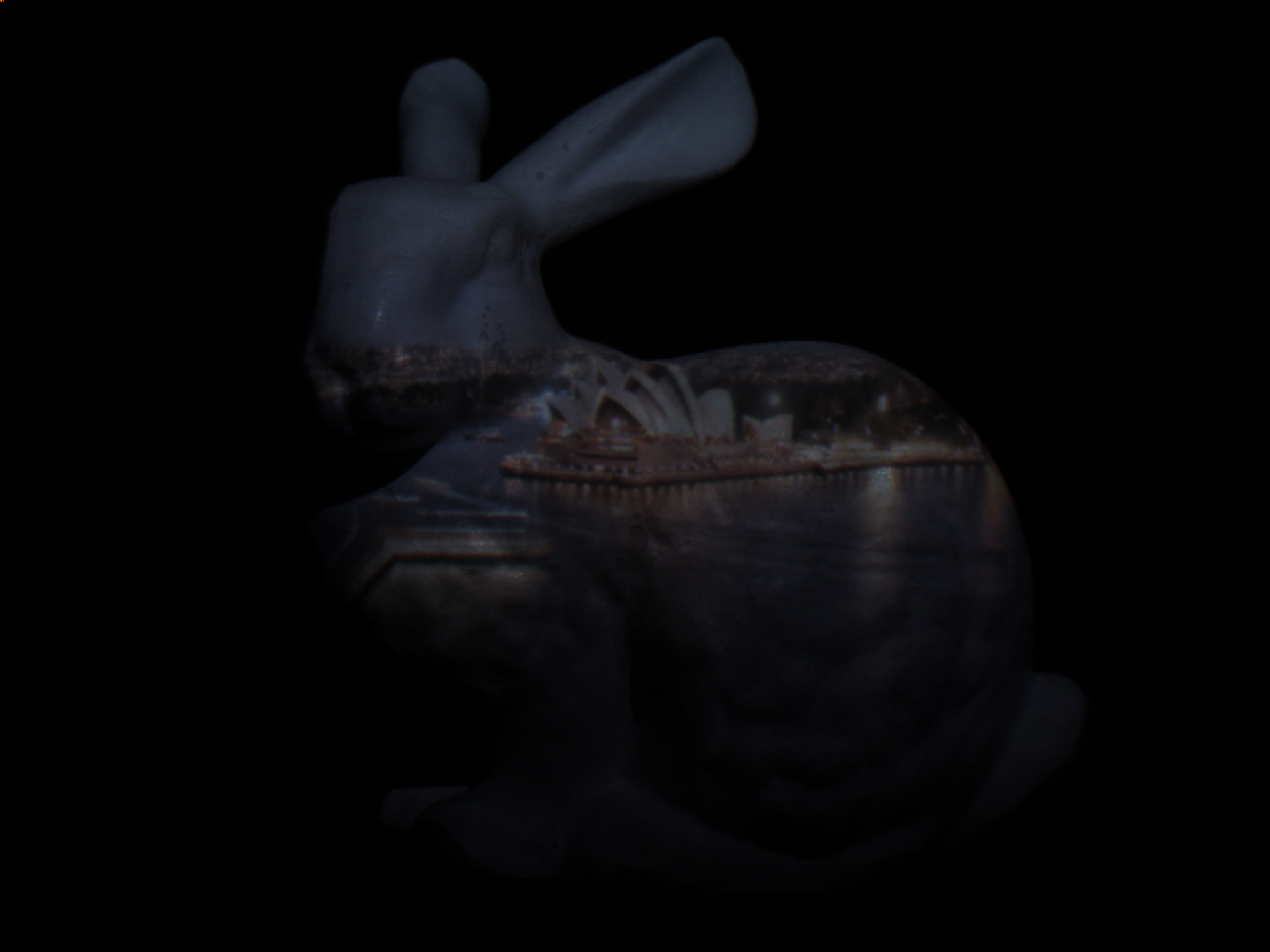} &
 \includegraphics[width=0.16\textwidth]{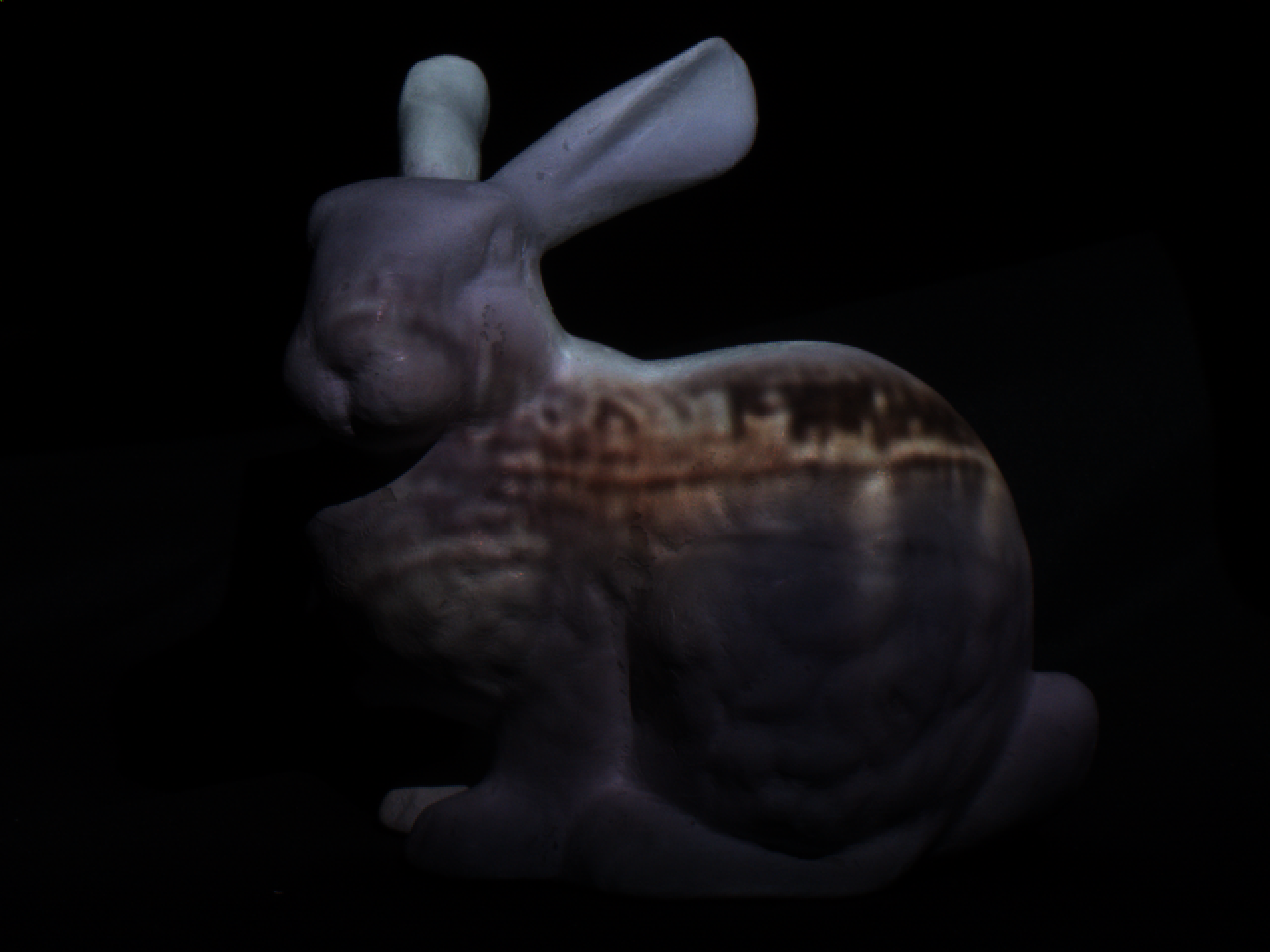} & \includegraphics[width=0.16\textwidth]{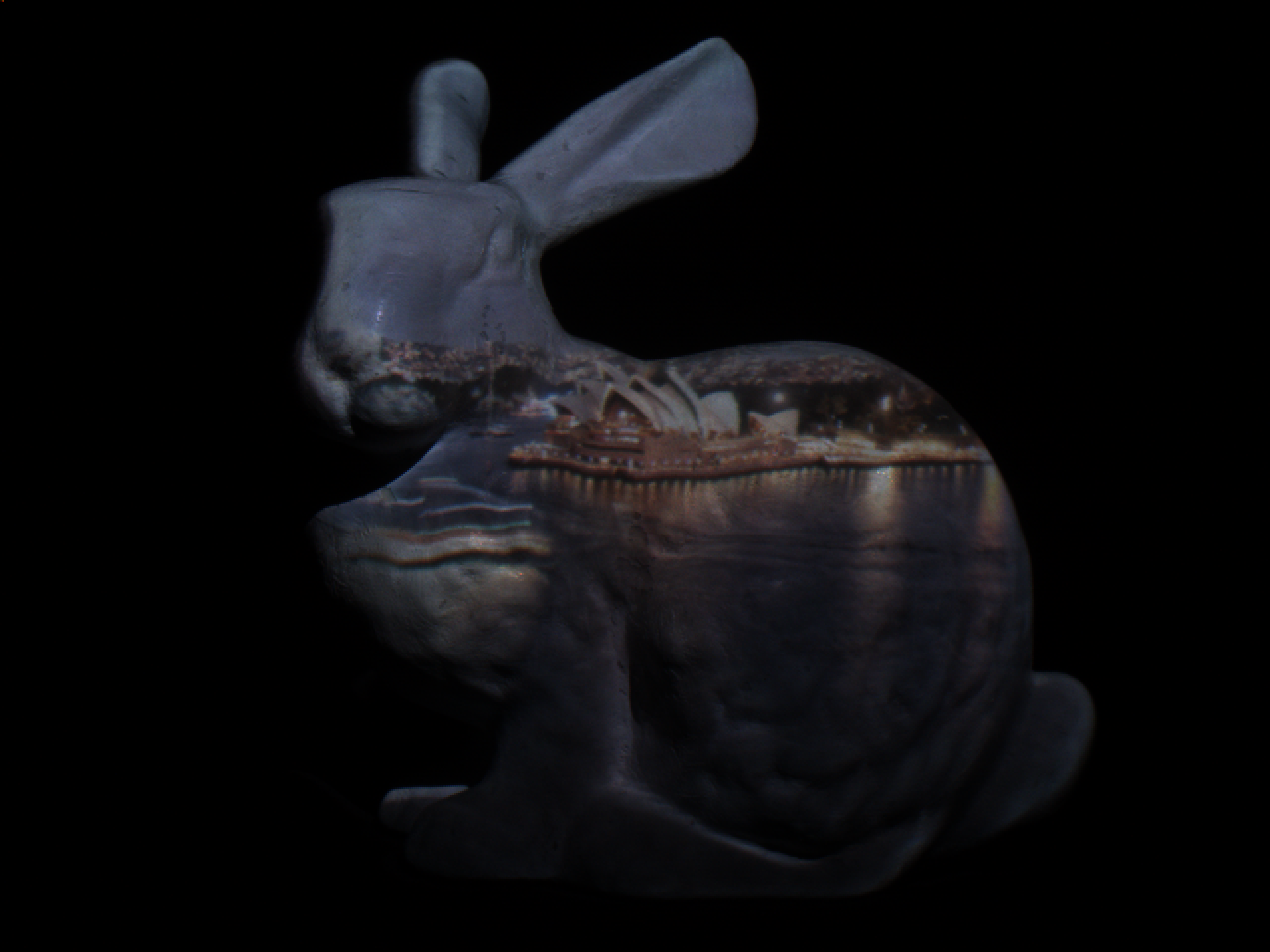} \\  \includegraphics[width=0.16\textwidth]{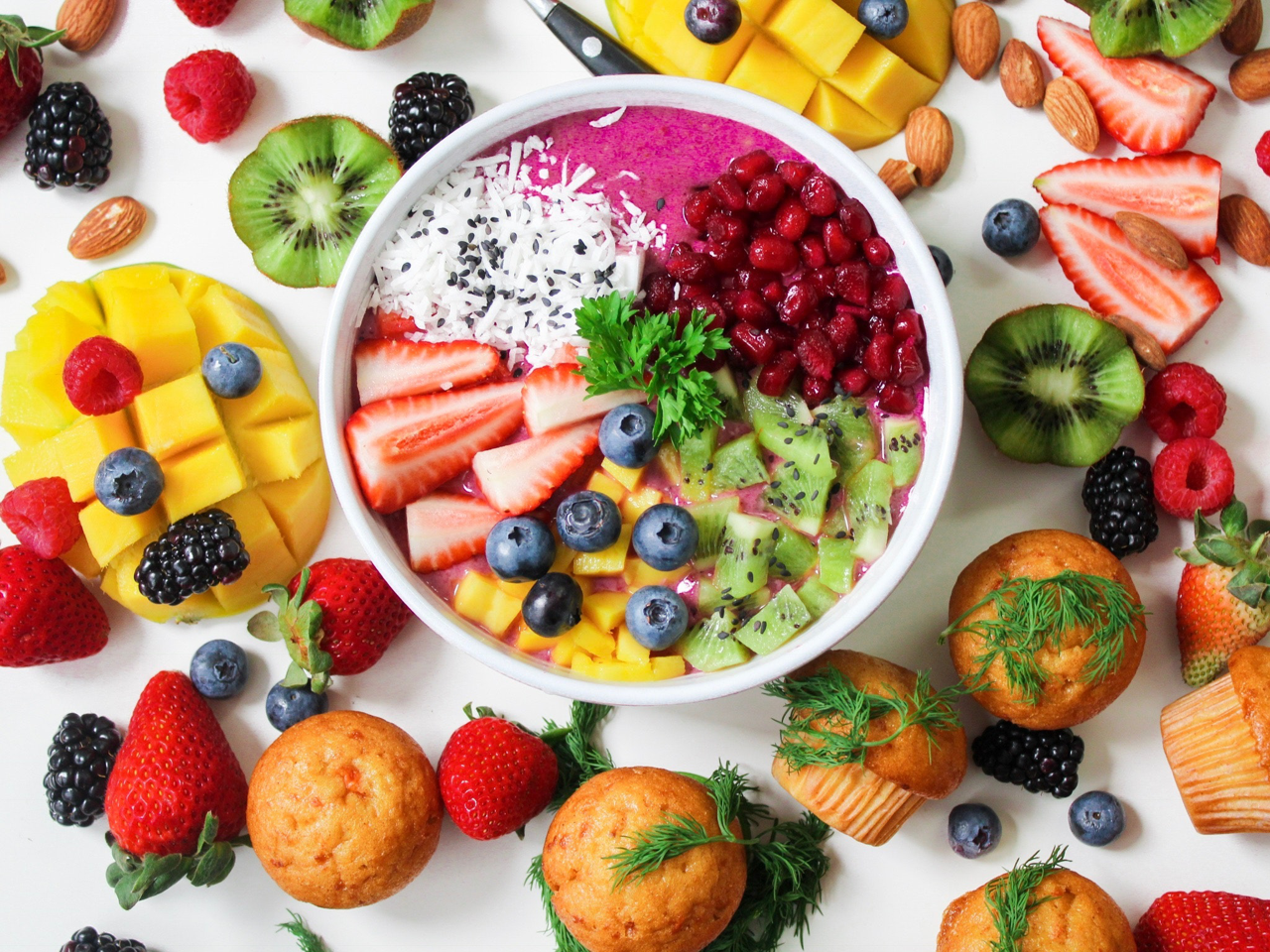} & \includegraphics[width=0.16\textwidth]{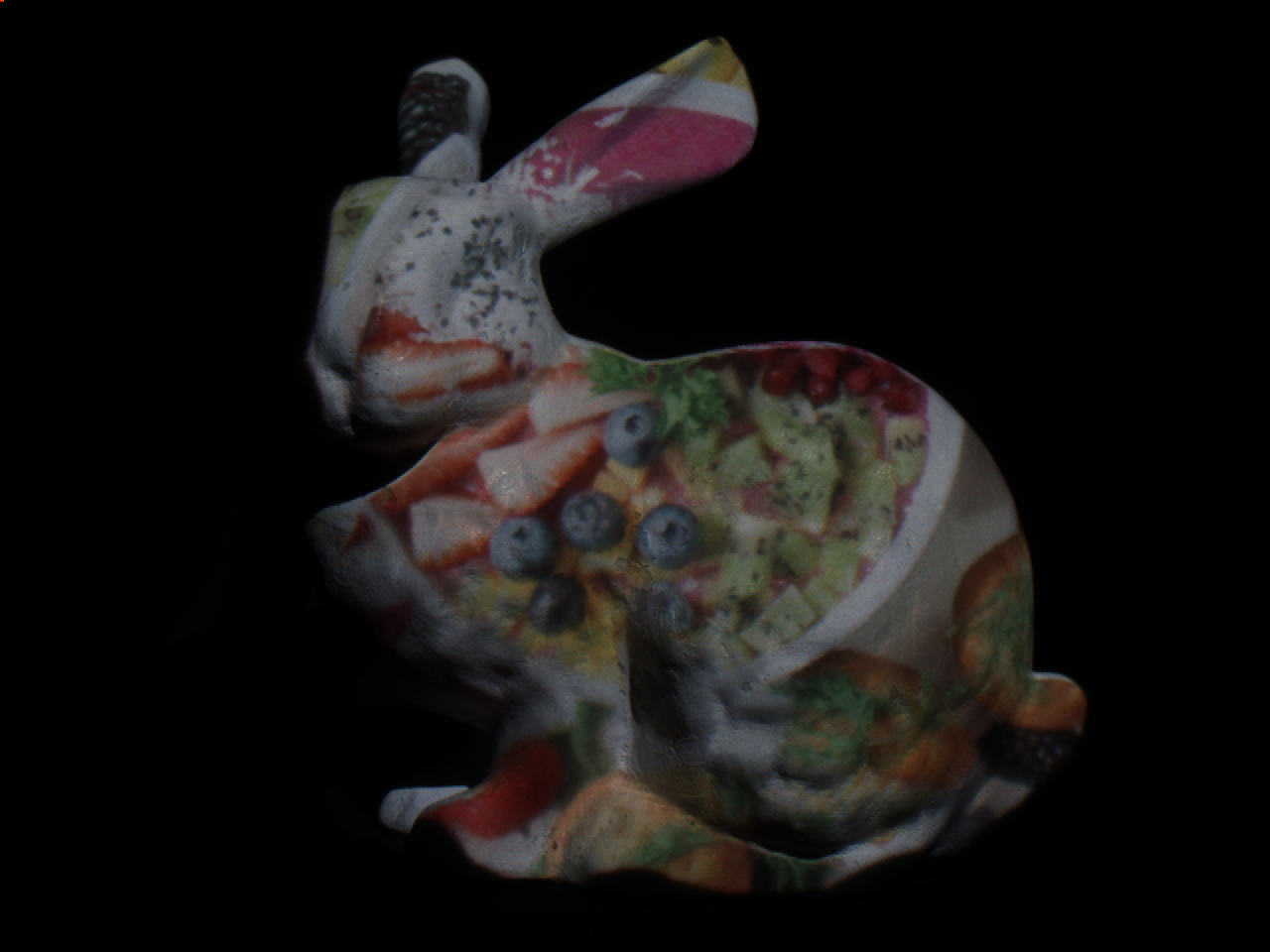} & \includegraphics[width=0.16\textwidth]{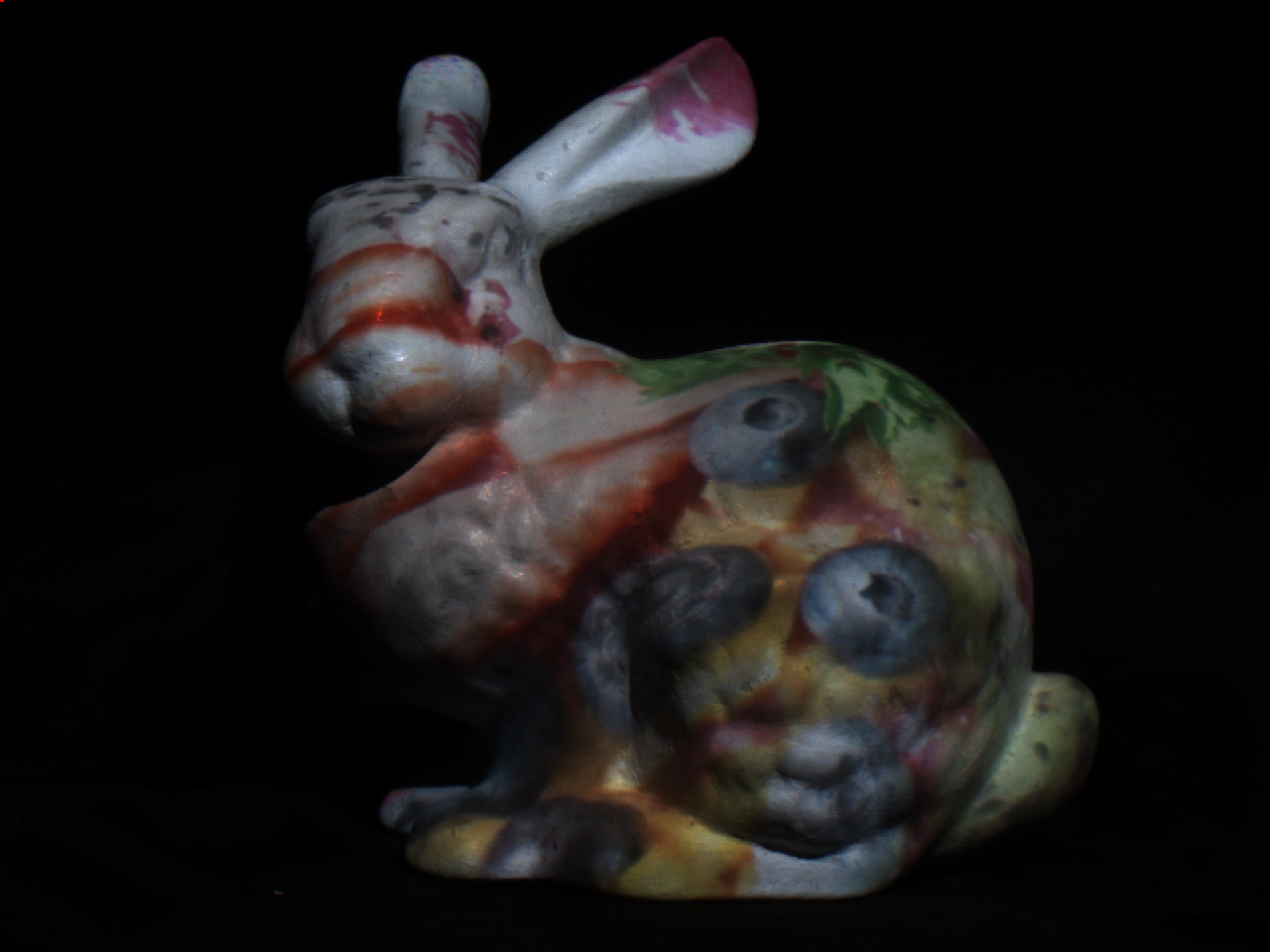} & \includegraphics[width=0.16\textwidth]{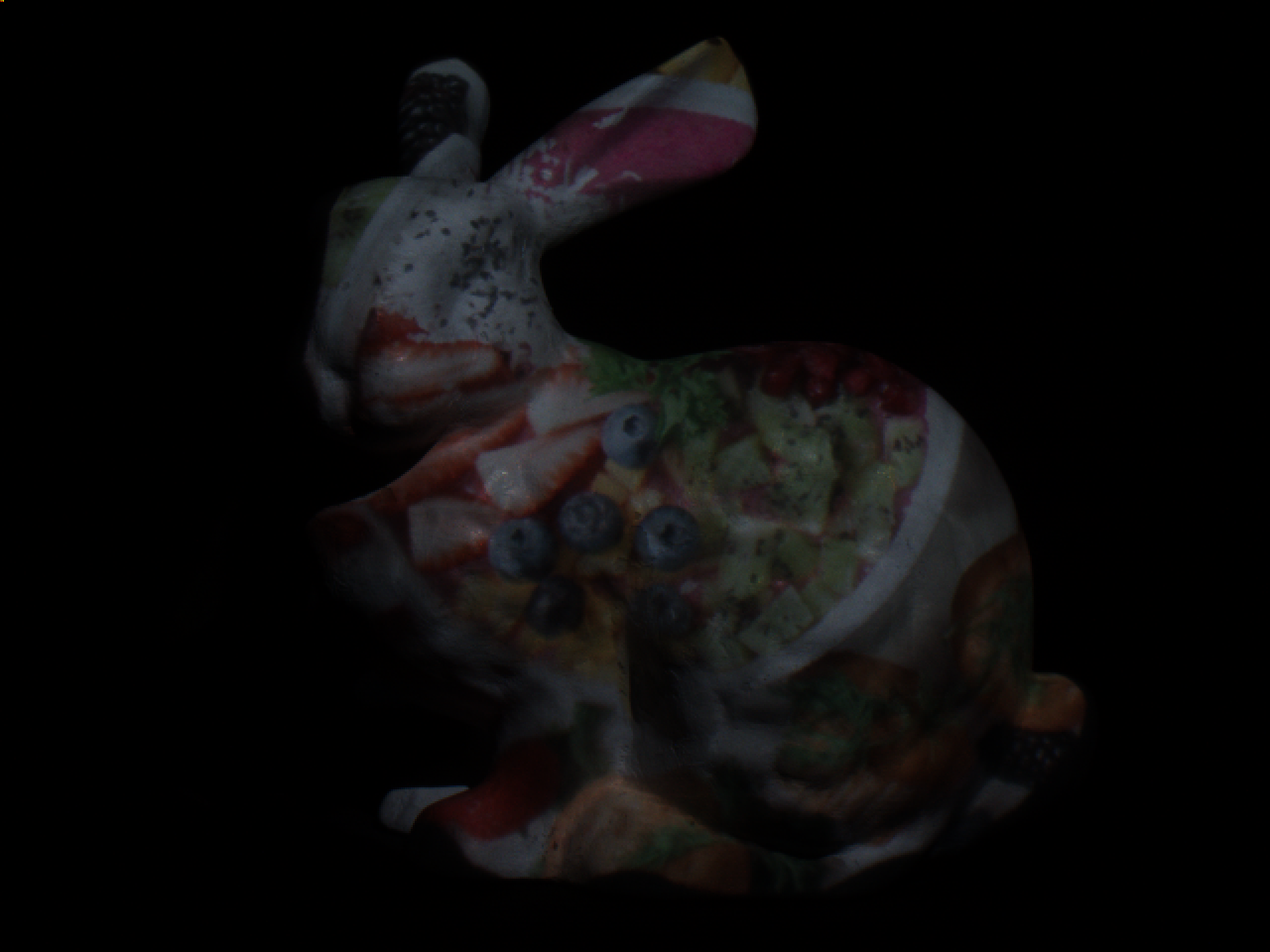} &
 \includegraphics[width=0.16\textwidth]{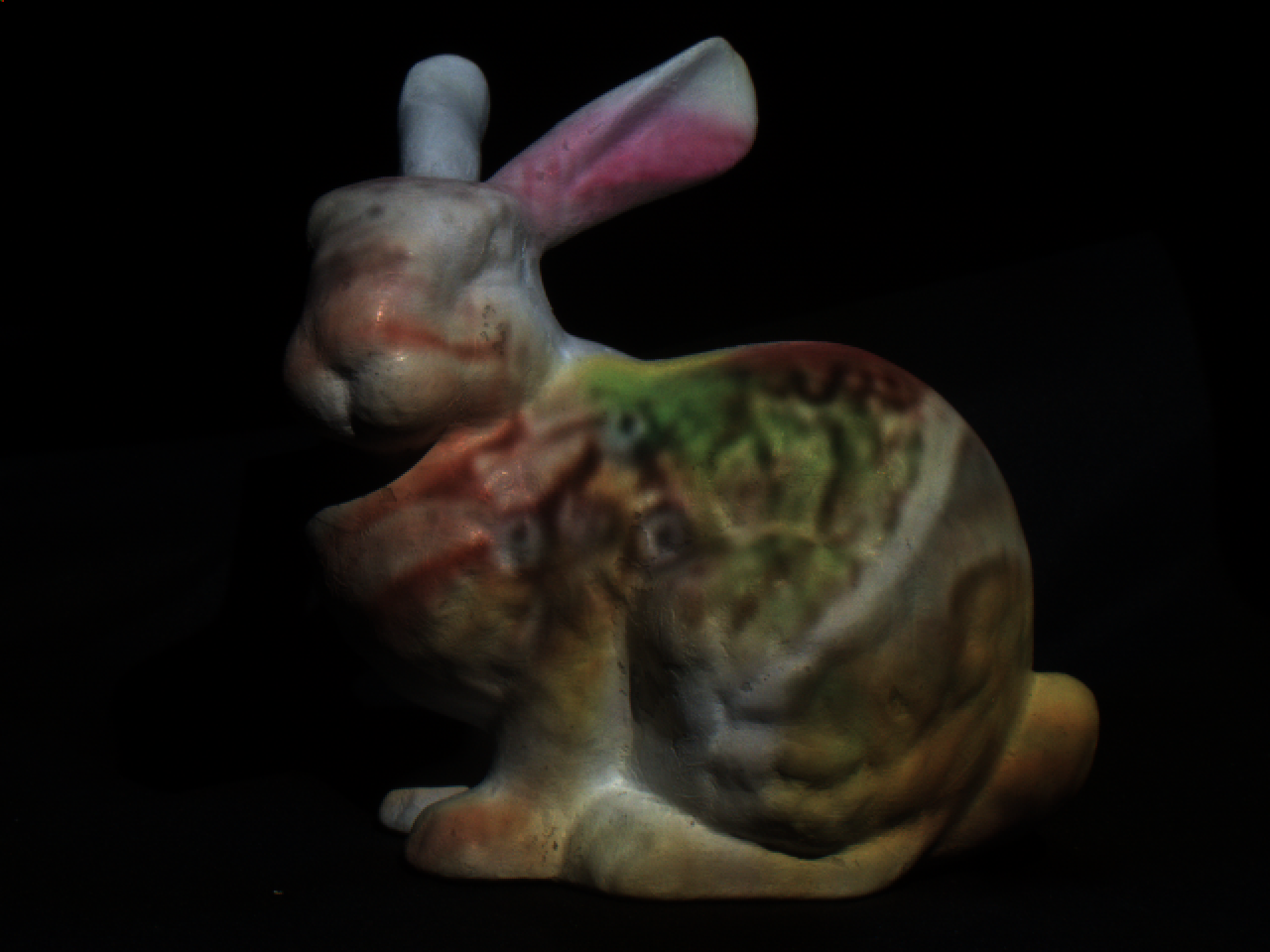} & \includegraphics[width=0.16\textwidth]{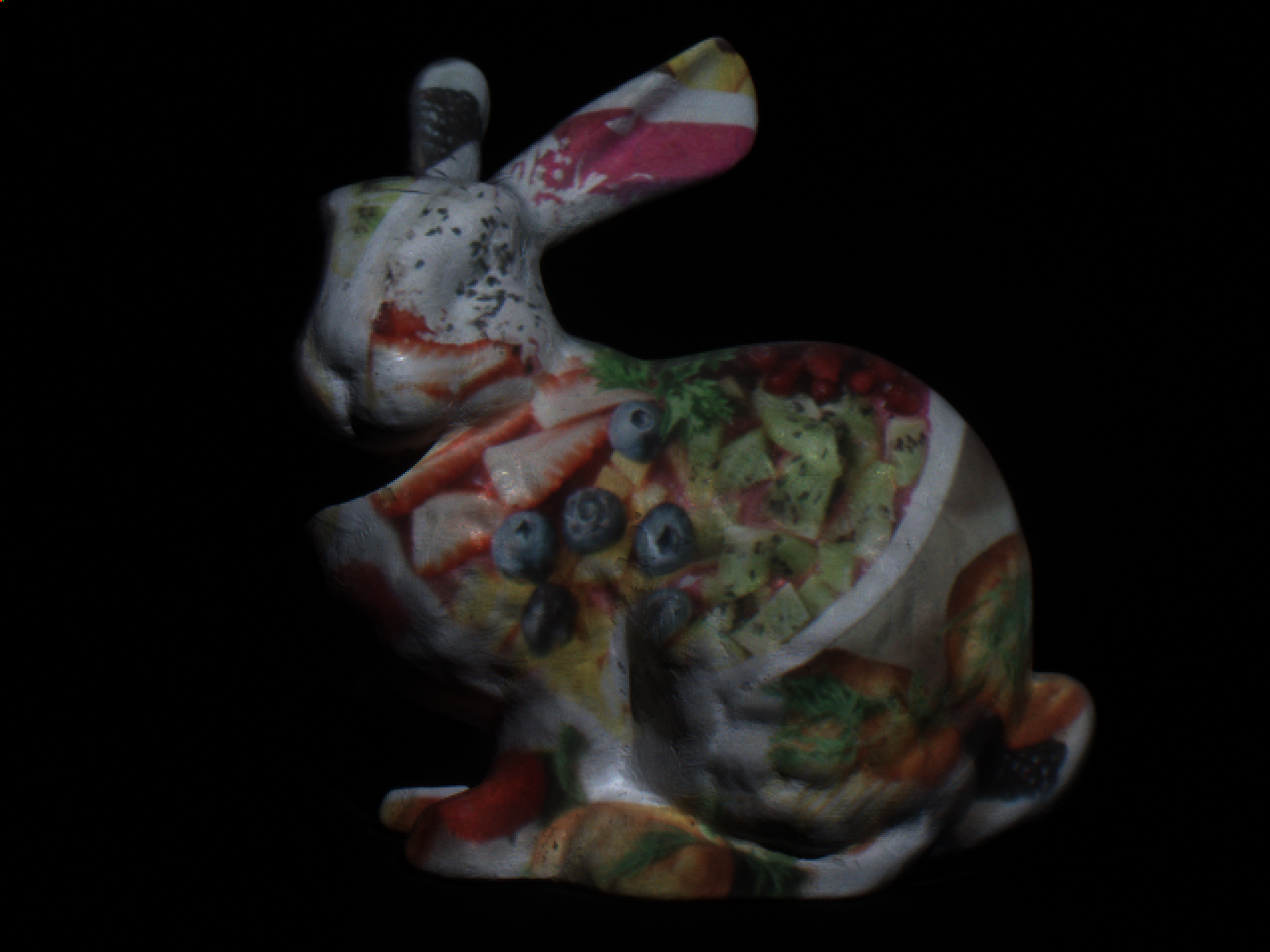} \\
\includegraphics[width=0.16\textwidth]{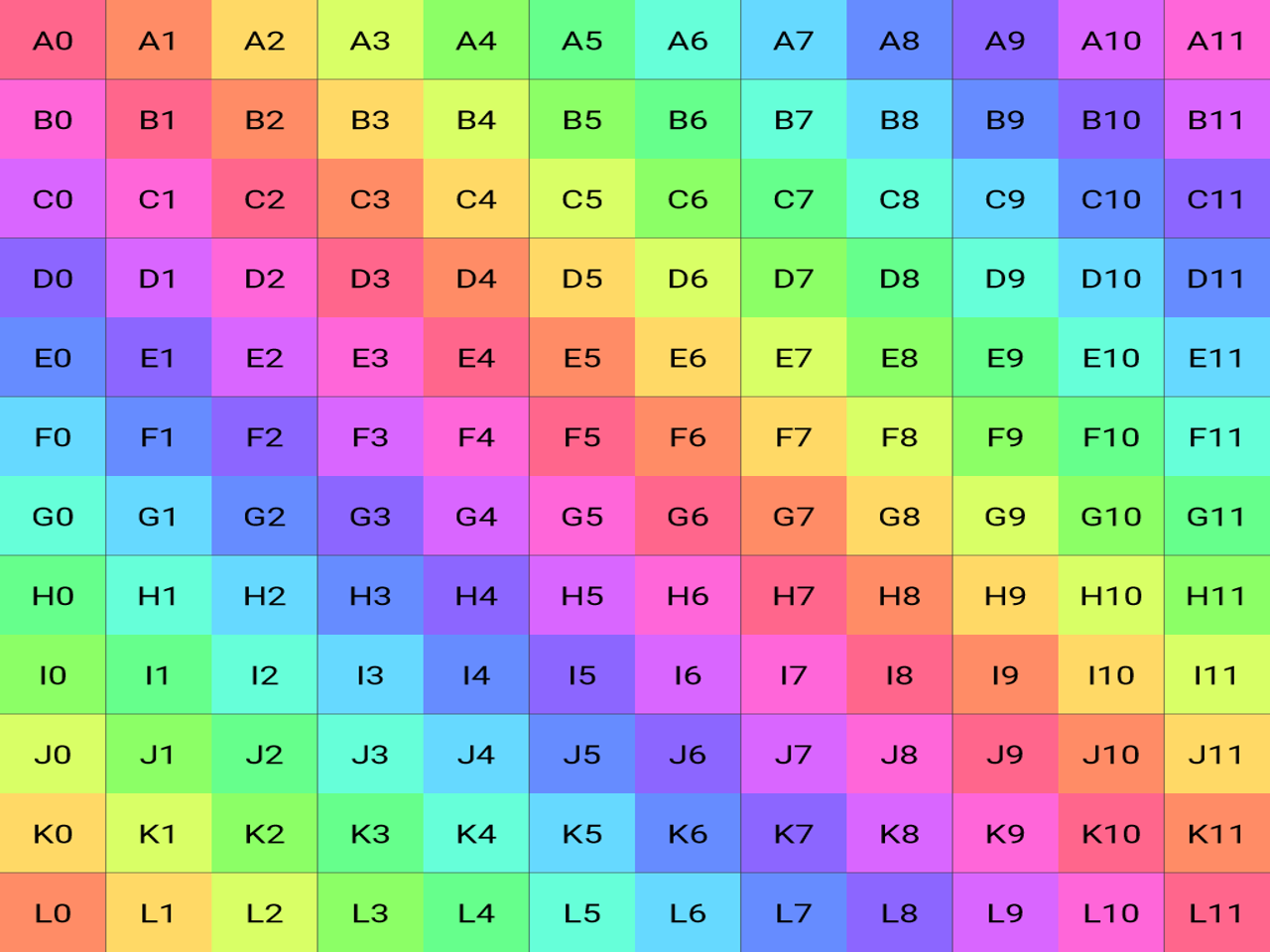} & \includegraphics[width=0.16\textwidth]{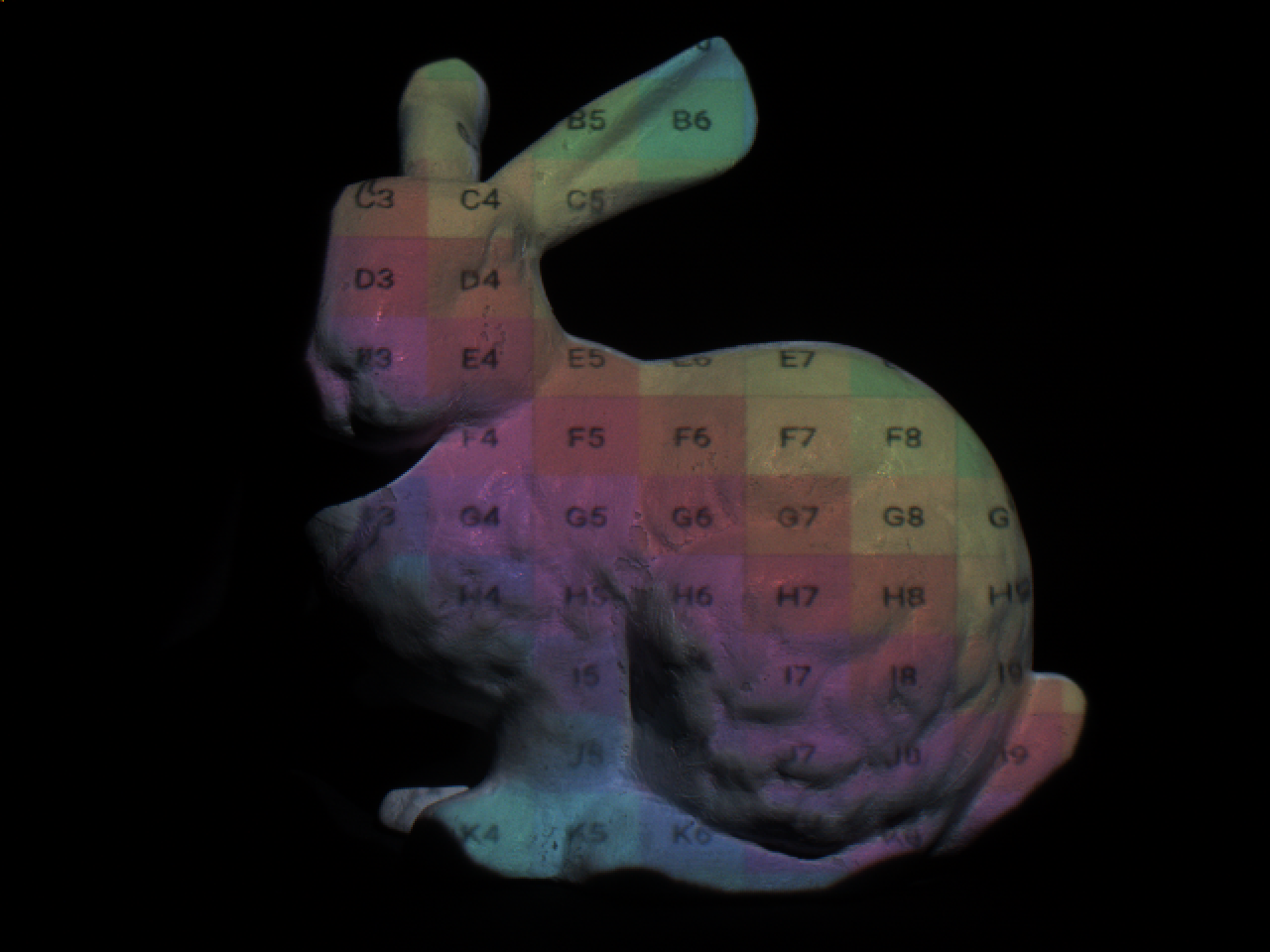} & \includegraphics[width=0.16\textwidth]{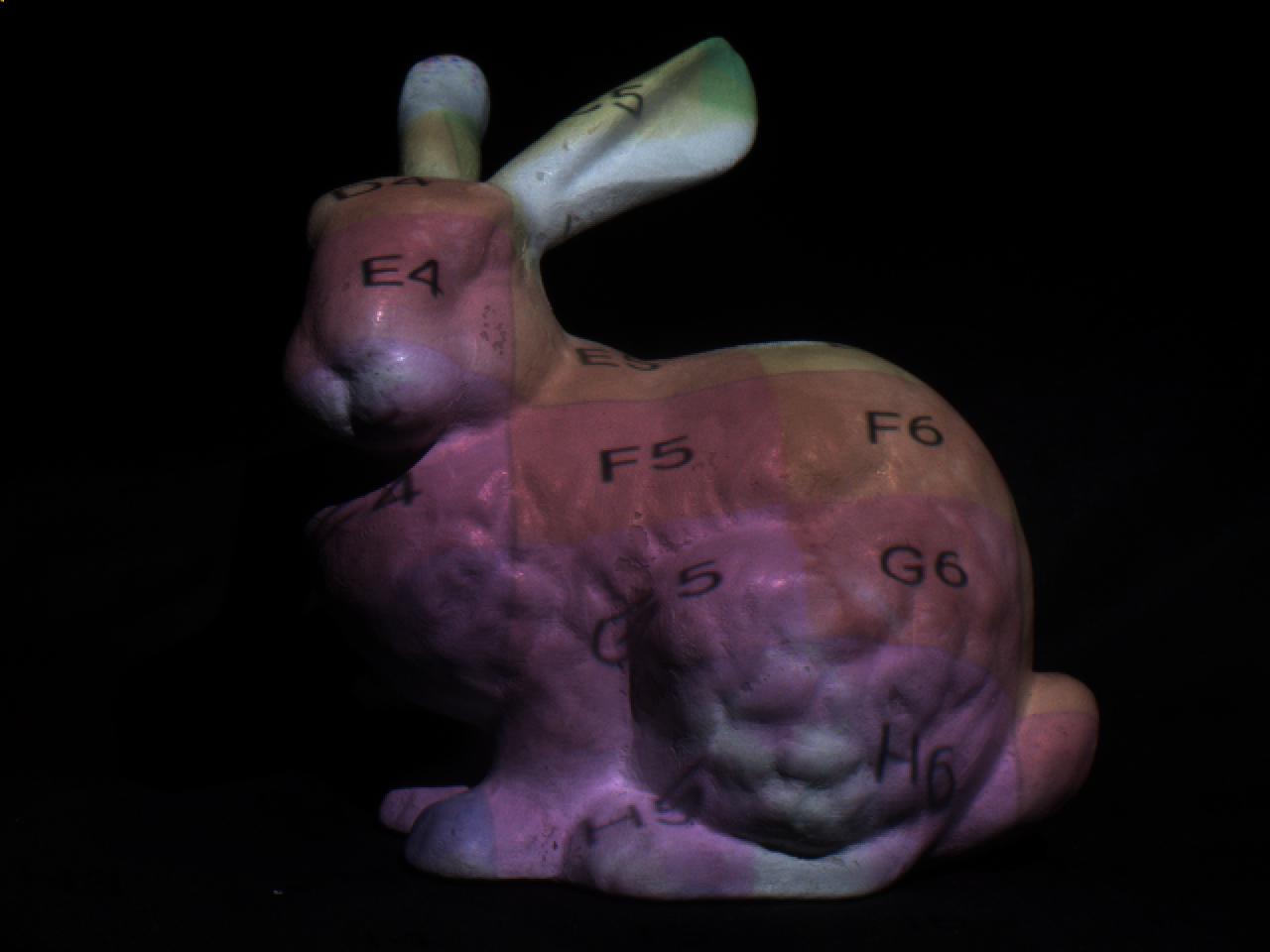} & \includegraphics[width=0.16\textwidth]{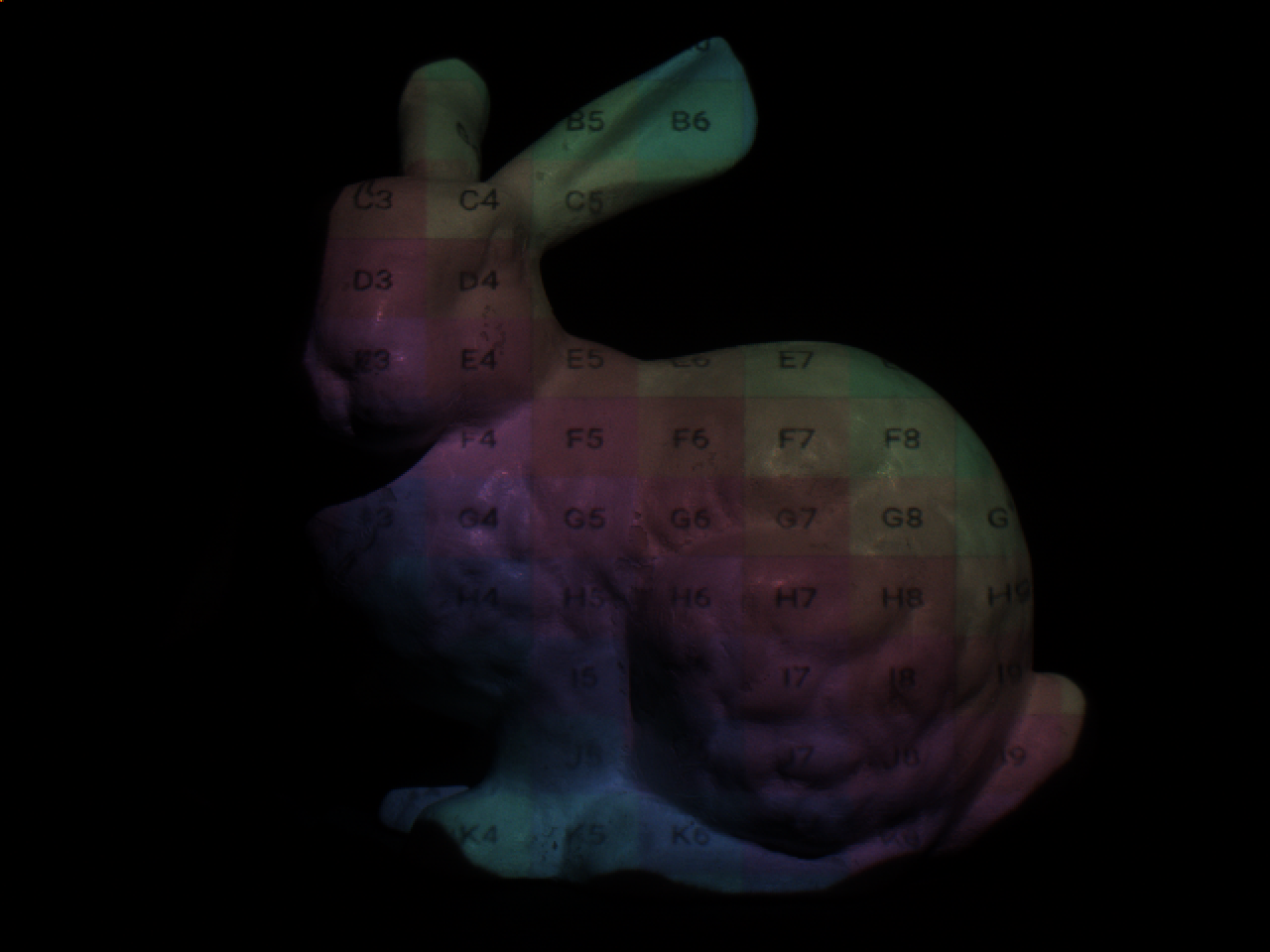} &
 \includegraphics[width=0.16\textwidth]{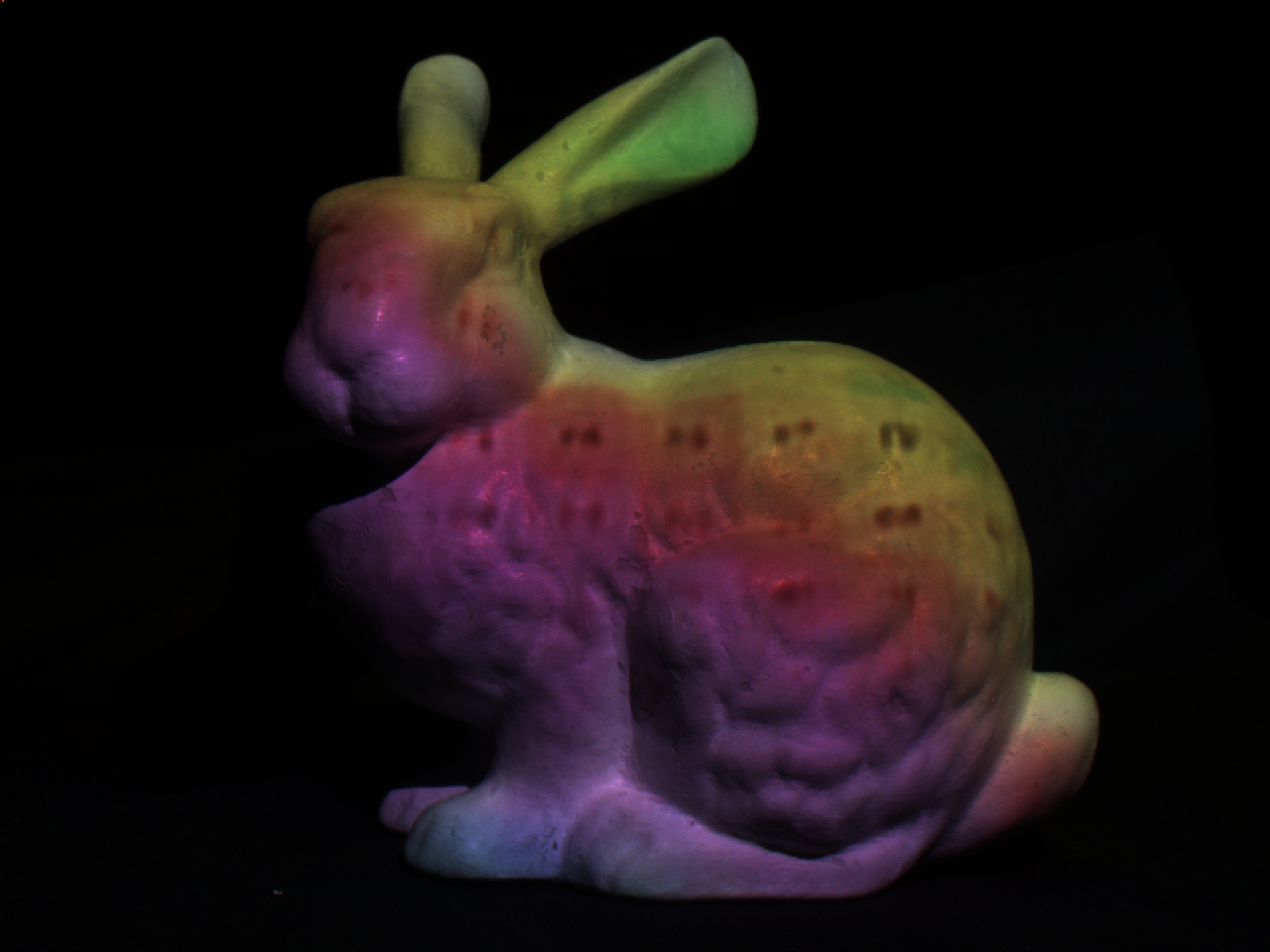} & \includegraphics[width=0.16\textwidth]{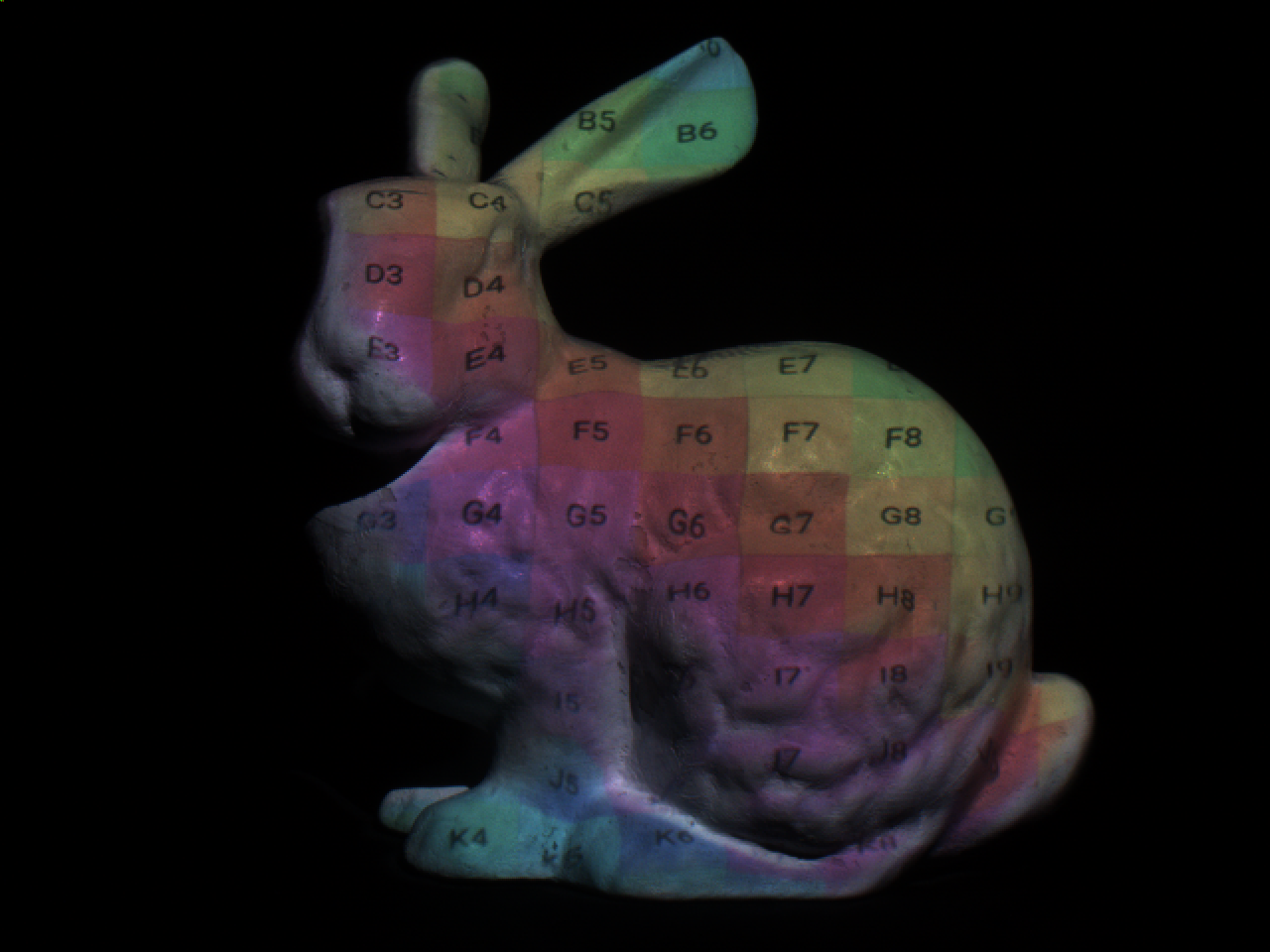} \\
 \includegraphics[width=0.16\textwidth]{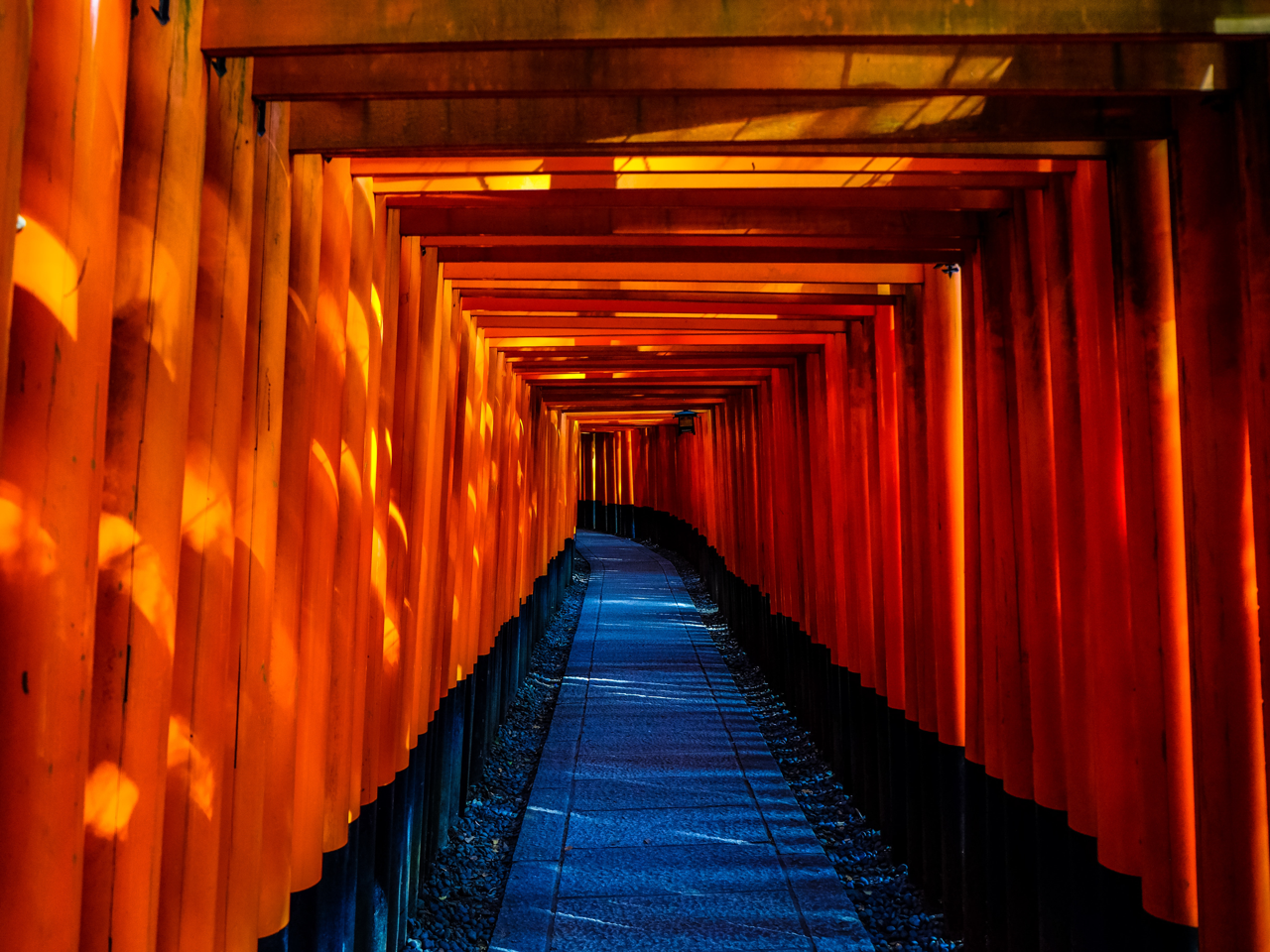} & \includegraphics[width=0.16\textwidth]{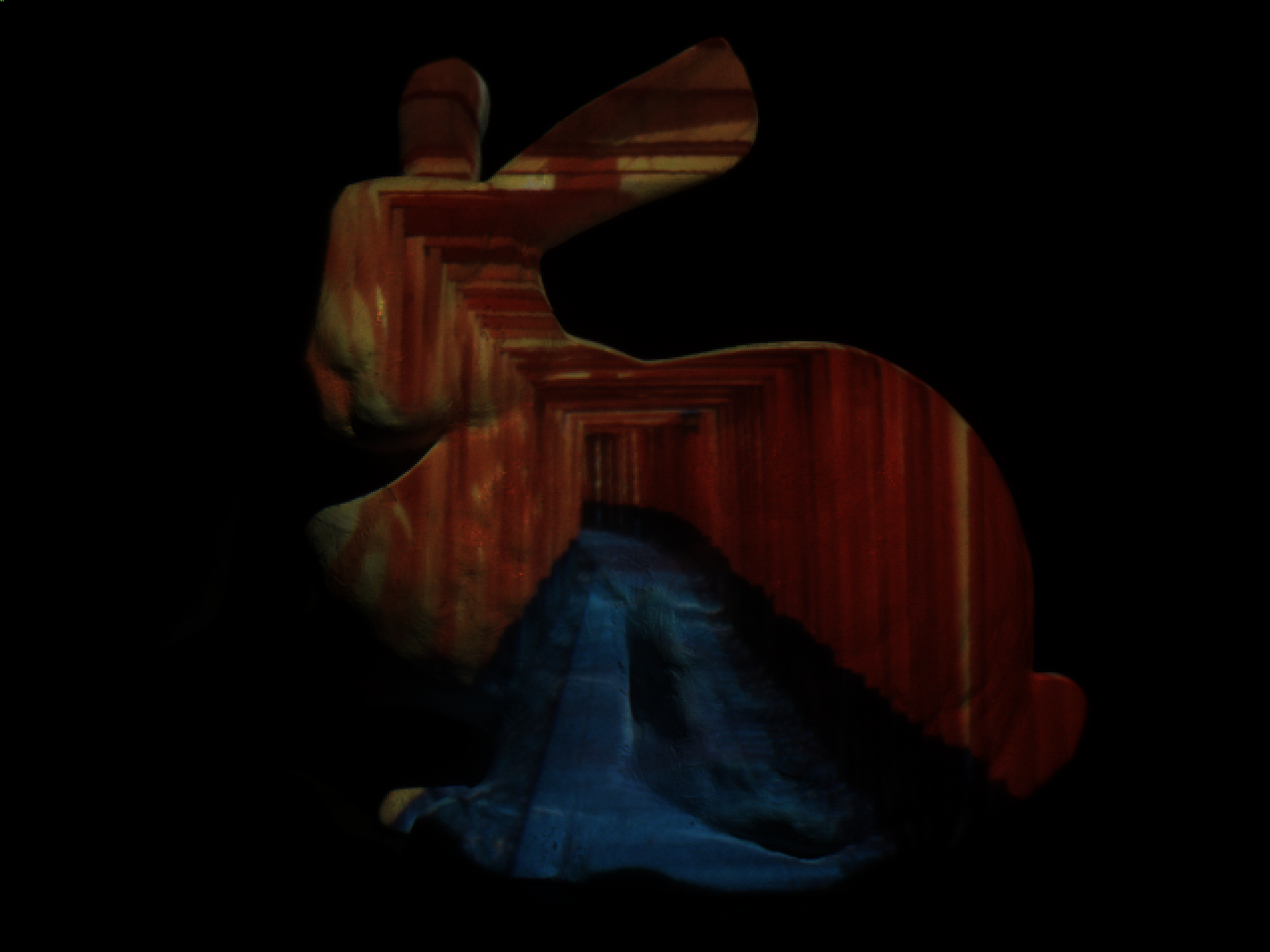} & \includegraphics[width=0.16\textwidth]{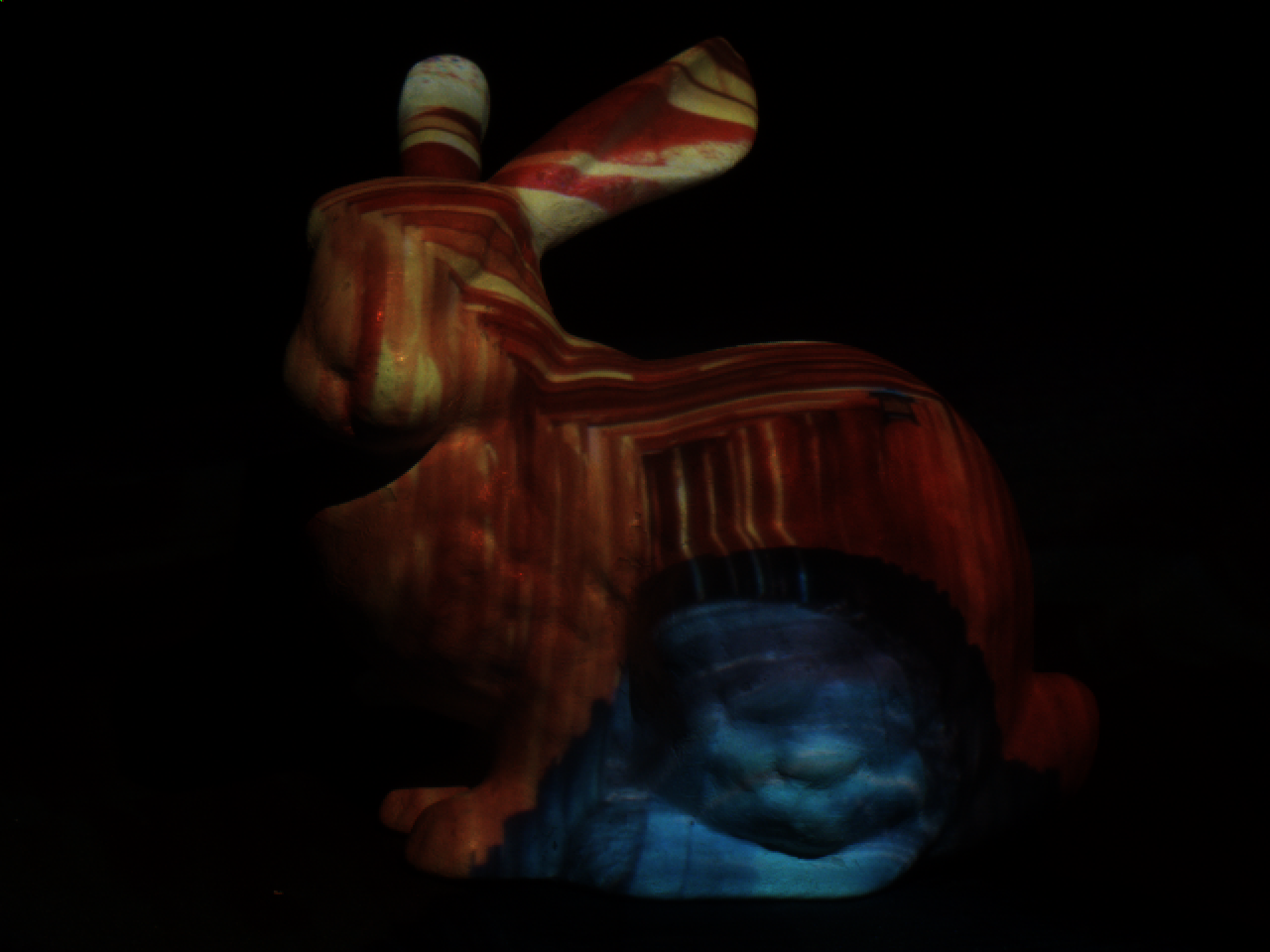} & \includegraphics[width=0.16\textwidth]{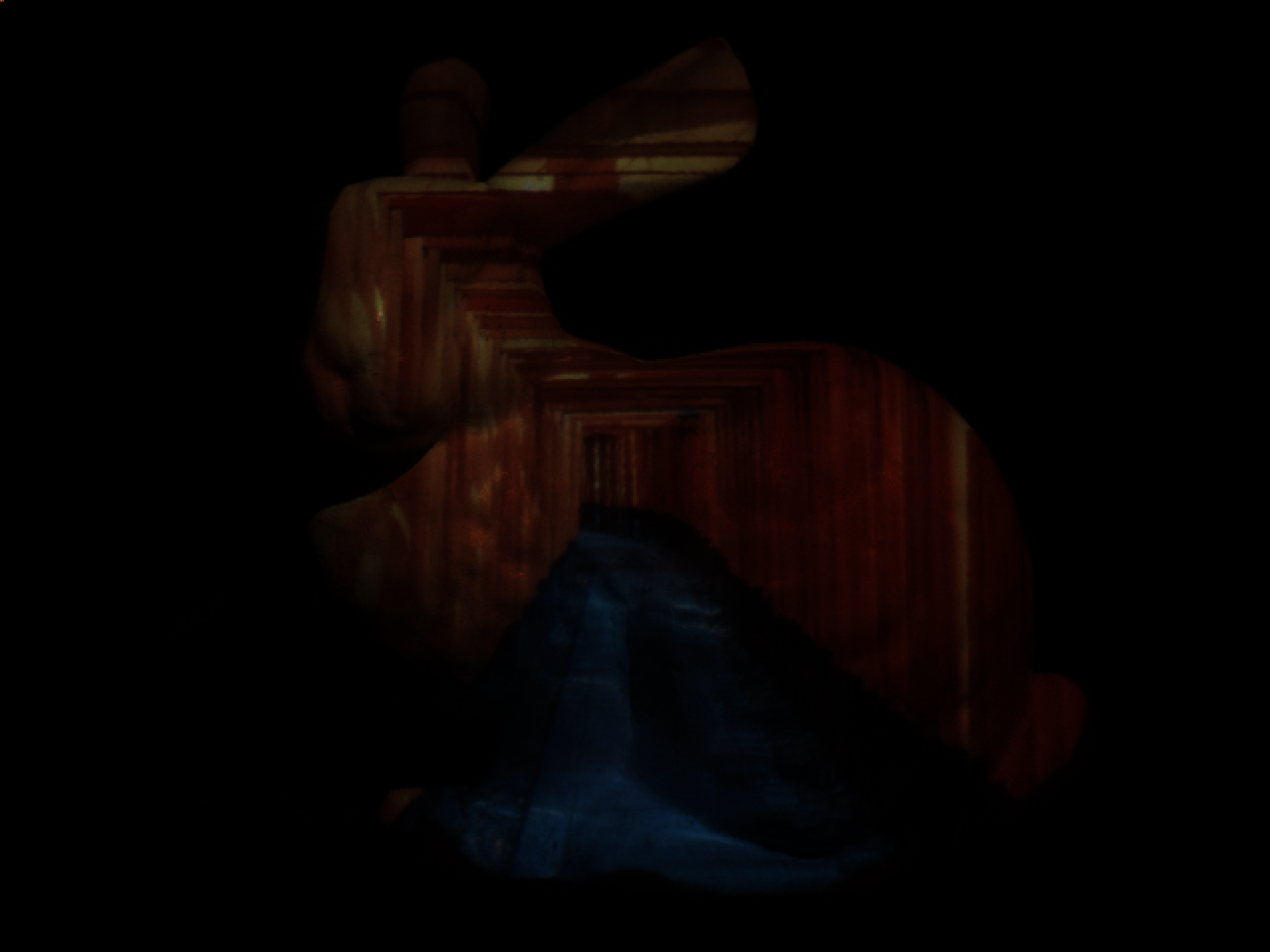} &
 \includegraphics[width=0.16\textwidth]{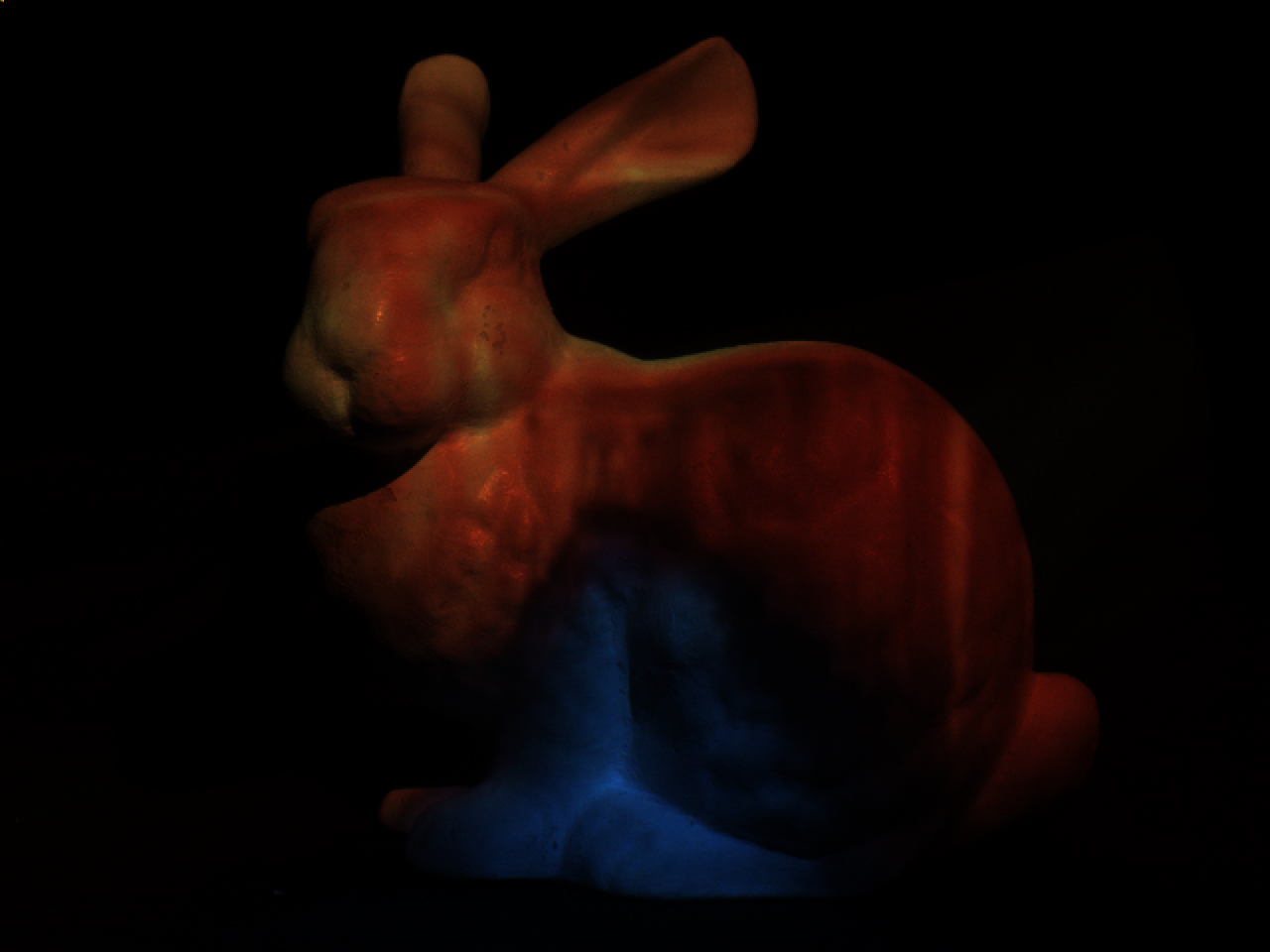} & \includegraphics[width=0.16\textwidth]{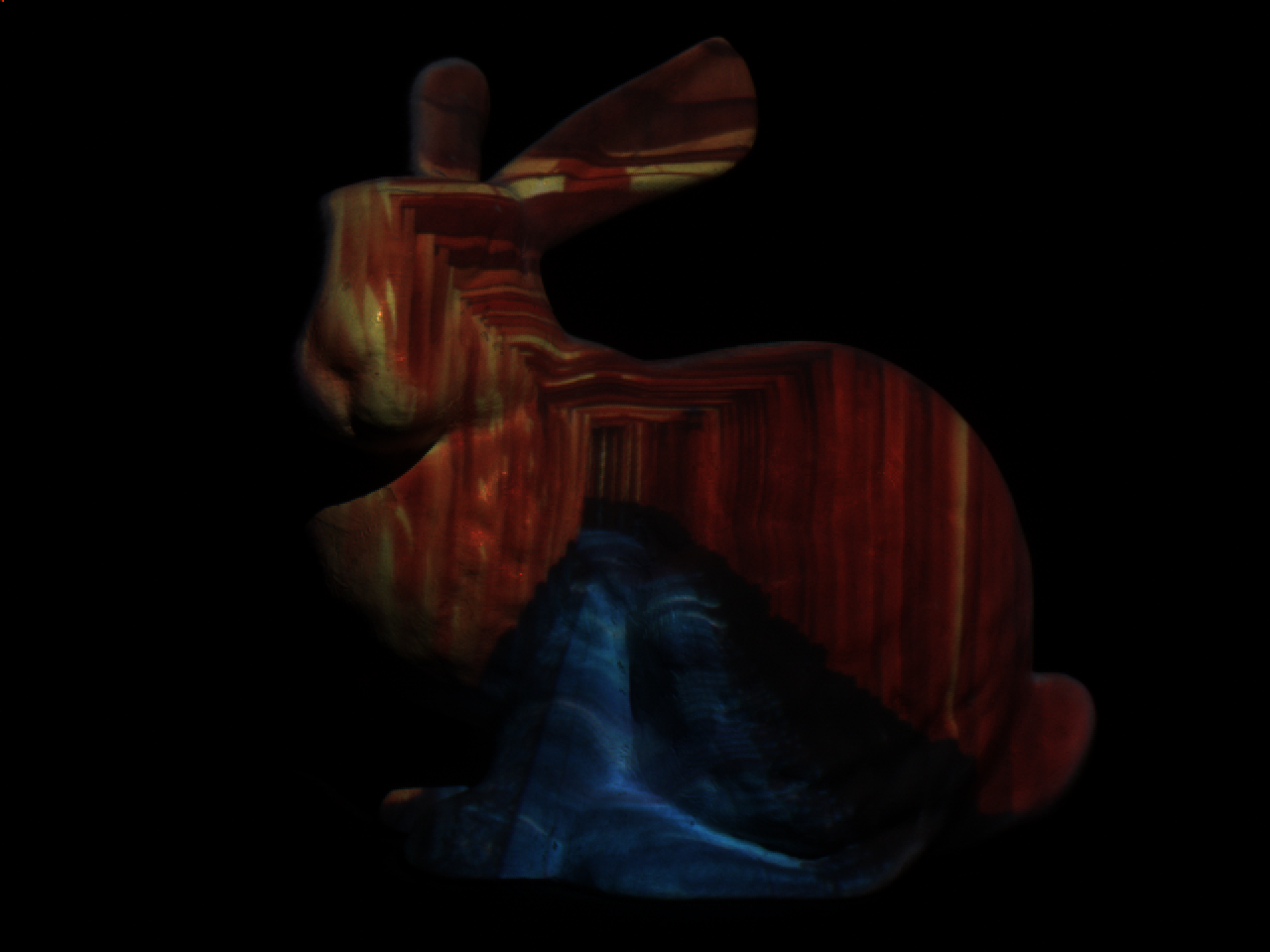} \\
 \hline
 \multicolumn{6}{c}{PSNR over 14 desired appearances} \\
\hline
 - & 17.8 & 13.2 & 15.8 & 16.7 & 17.5
 
\end{tabular}

\caption{Image compensation on a real scene. Desired Appearance: the desired appearance for the Bunny scene. SL Calib. + Gamma Cor. + Col. Comp: A structured light geometric calibration was performed to establish pixel to pixel correspondence, the desired image was then gamma corrected, and a classical color compensation technique \cite{naive_compensate} was used to achieve the desired appearance (for one particular view). No Calib.: no geometric calibration was performed, but color compensation was used to adjust the colors. No Comp.: A geometric calibration was performed, but without color compensation. CompenNeSt++\cite{compensnet_pp}: a state of the art framework for jointly estimating the geometric distortion and color compensation using deep learning.}
\label{fig:compensation}
\end{figure*}

\subsubsection{See through "XRAY"}

\label{sec:xray}
One immediate application enabled by our framework is the ability to edit the scene in a way where an occluder appears transparent (\cref{fig:xray}). A slightly na\"{i}ve (but feasible) way to do this would be to render the scene with a novel viewpoint placed beyond the occluder to obtain a desired view, and optimize the projector image to yield that exact image from the original cameras perspective. However, a much more efficient approach is to replace the optimization with a 2-pass render \cite{10.1145/280814.280861}: we swap the projector with a virtual camera having its exact intrinsics and extrinsics, and the actual camera with a virtual projector having the cameras parameters. Then we render the scene from the virtual cameras viewpoint (while the virtual projector is shining the desired view), and use the resulting render as the original projectors texture. This is achieved in $\sim20$ seconds with our implementation. Note that in \cref{fig:xray} we cropped the region of interest in the projected image for clarity, and artificially added a gray background such that the other elements of the scene can be observed.

\subsubsection{Projector compensation}
\label{sec:compnensation}
One important goal of a procam setup is to allow the viewer to experience a specific desired appearance even when projecting on non-flat, and not perfectly white surfaces. To achieve this, we seek an image, such that when projected the desired view will be obtained. 
As discussed in \cref{sec:relwork_procam}, both classical and recent deep neural network-based methods are constricted to a single viewpoint and projector location.

In our formulation, it is possible to achieve this for any view point and for any projector location. To do this, we define a new optimization (after the main optimization described in \cref{sec:optimize} converges), where the parameters for optimization are now an image $I\in \mathbb{R}^{n\times m\times3}$, which is the projector pattern, and the objective is to yield the closest possible result to a desired appearance using $Loss_{img}$, while keeping all the other parameters fixed. We initialize $I$ to all-zero, but pass it through a sigmoid function before it is projected into the virtual scene which we found helps convergence (i.e. the objective becomes to find an image $I$ such that its sigmoid yields the best result). The viewpoint is fixed and treated as an input for this optimization.

We compared the color compensation performances among our method and existing techniques.
We used the bunny scene and projected 14 desired appearances onto it with either a classical method \cite{naive_compensate}, state of the art method (CompenNeSt++) \cite{compensnet_pp}, and our technique.
In addition to this, we captured projected results without the geometric correction and color compensation as references.
The viewpoint used to capture the results was same among all methods and was used in calibration in all other methods.
However, it was a novel viewpoint for our method, which means that no image captured from this viewpoint was used in training our networks. The optimization took $\sim10$ minutes to converge using our implementation.

Results can be seen in \cref{fig:compensation}. The PSNR was computed over the entire image plane.
The relatively low PSNR values come from the black background in which no image can be displayed.
We can confirm that the three techniques improved the image quality over no geometric correction and no color compensation.
The classical technique provided the best results in this experiment.
On the other hand, CompenNeSt++ fails to capture detail due to reliance on a warp grid that only works for (roughly) planar surfaces.
Using our learned scene representation, we can optimize a projection image such that the desired appearance is achieved for any geometry, and with on par quality to a full geometric calibration and color compensation technique.
Again, we would like to emphasize that our result is computed for a novel viewpoint. Additionally, we found that the predicted normals decomposed from the scene had a strong effect on the final compensation quality from our method, which emphasizes the need for using $Loss_n$ defined in \cref{eq:normal_loss}.

\subsubsection{Text to projection}

\begin{figure*}[ht]
\setlength{\tabcolsep}{1pt}
\centering
\begin{tabular}{cccccc}
 User Input & Original View & Desired View & Projection Image & Simulated & Reprojected \\
 \begin{tabular}[c]{@{}c@{}}"Side profile \\ of The Batman" \end{tabular} & \raisebox{-.5\height}{\stackinset{l}{1pt}{b}{1pt}{\includegraphics[width=0.06\textwidth]{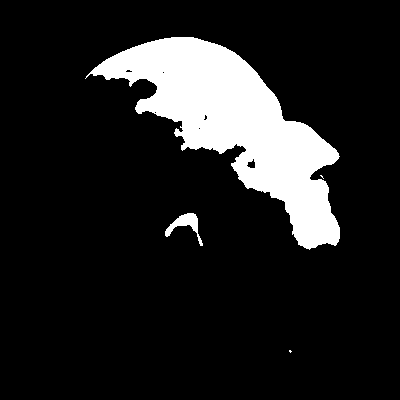}}{\includegraphics[width=0.15\textwidth]{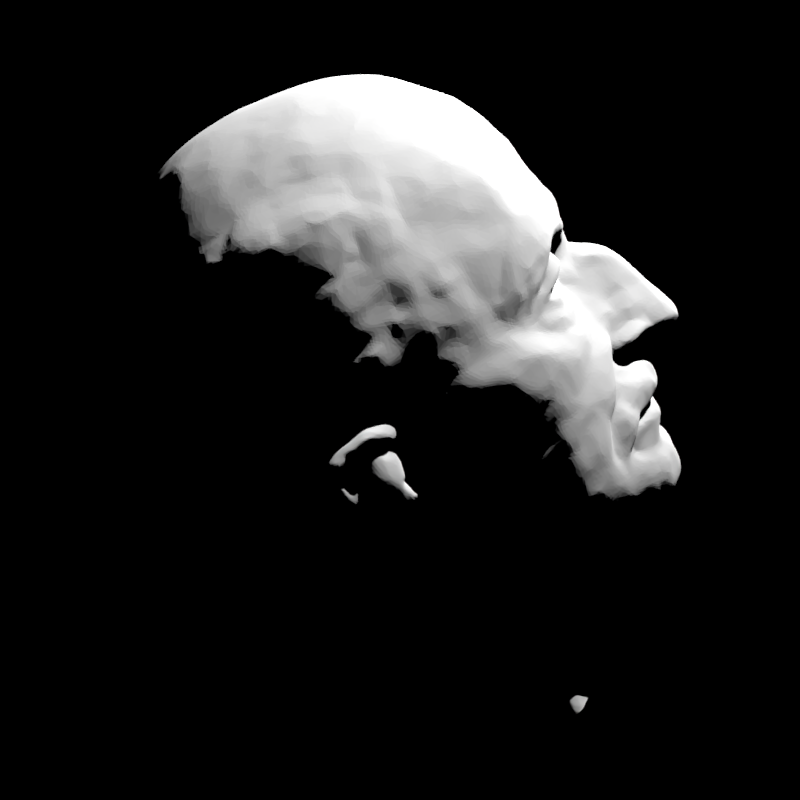}}} & \raisebox{-.5\height}{\includegraphics[width=0.15\textwidth]{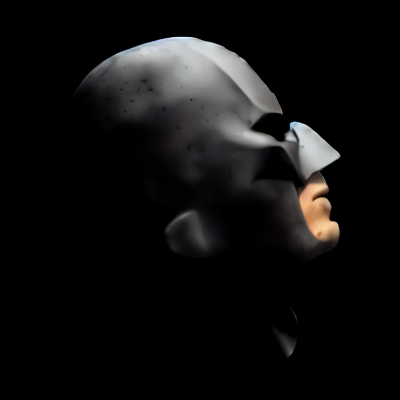}} & \multirow{2}{*}{\includegraphics[width=0.15\textwidth]{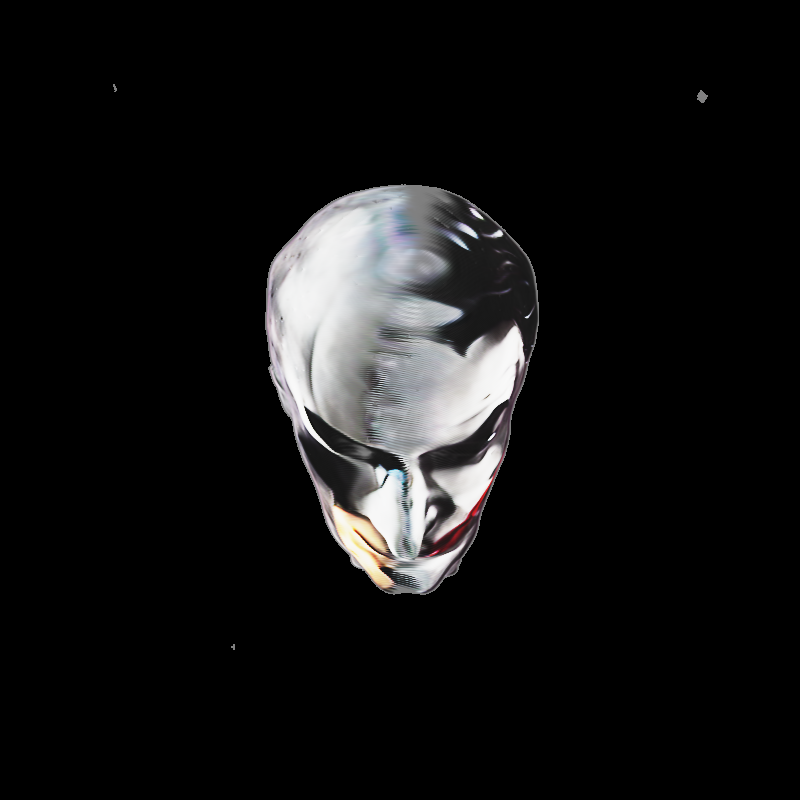}} & \raisebox{-.5\height}{\includegraphics[width=0.15\textwidth]{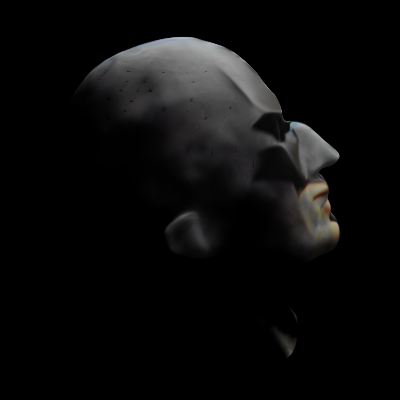}} & \raisebox{-.5\height}{\includegraphics[width=0.15\textwidth]{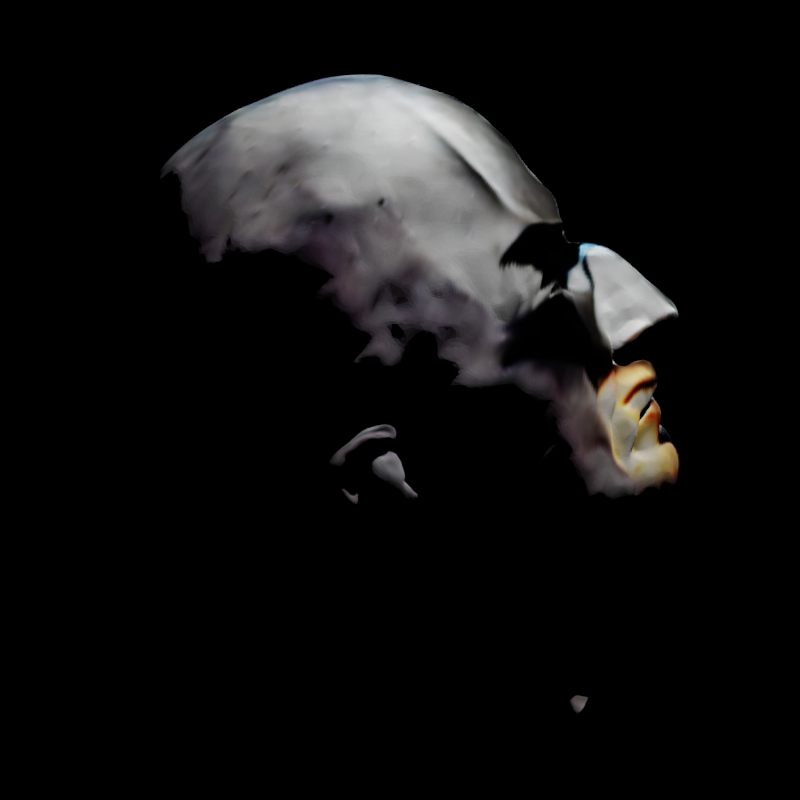}}\\
\begin{tabular}[c]{@{}c@{}}"Side profile \\ of The Joker" \end{tabular} & \raisebox{-.5\height}{\stackinset{r}{1pt}{b}{1pt}{\includegraphics[width=0.06\textwidth]{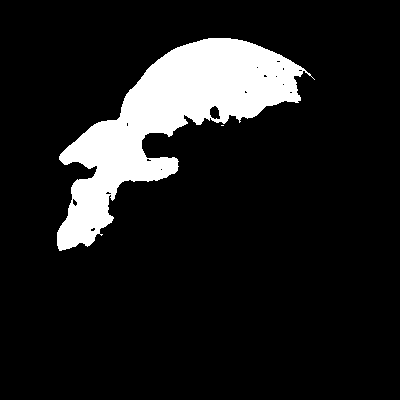}}{\includegraphics[width=0.15\textwidth]{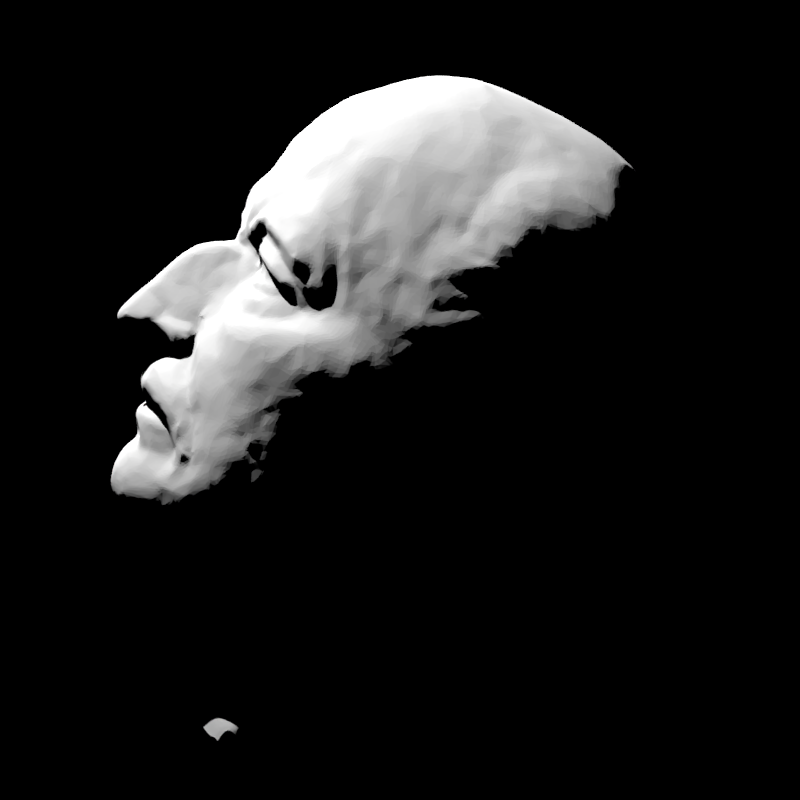}}} & \raisebox{-.5\height}{\includegraphics[width=0.15\textwidth]{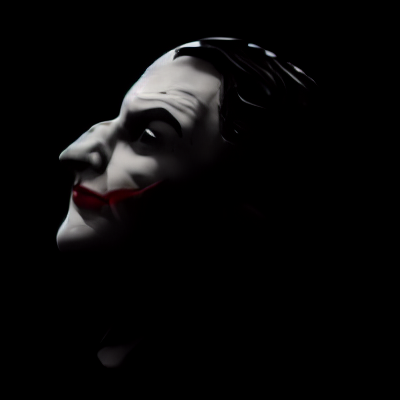}} && \raisebox{-.5\height}{\includegraphics[width=0.15\textwidth]{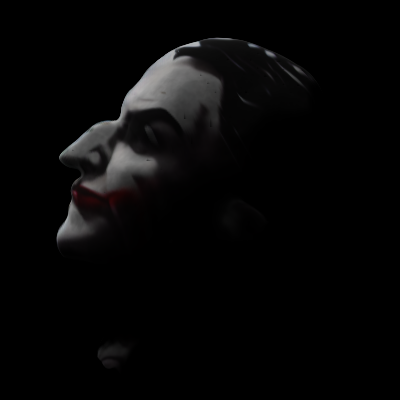}} & \raisebox{-.5\height}{\includegraphics[width=0.15\textwidth]{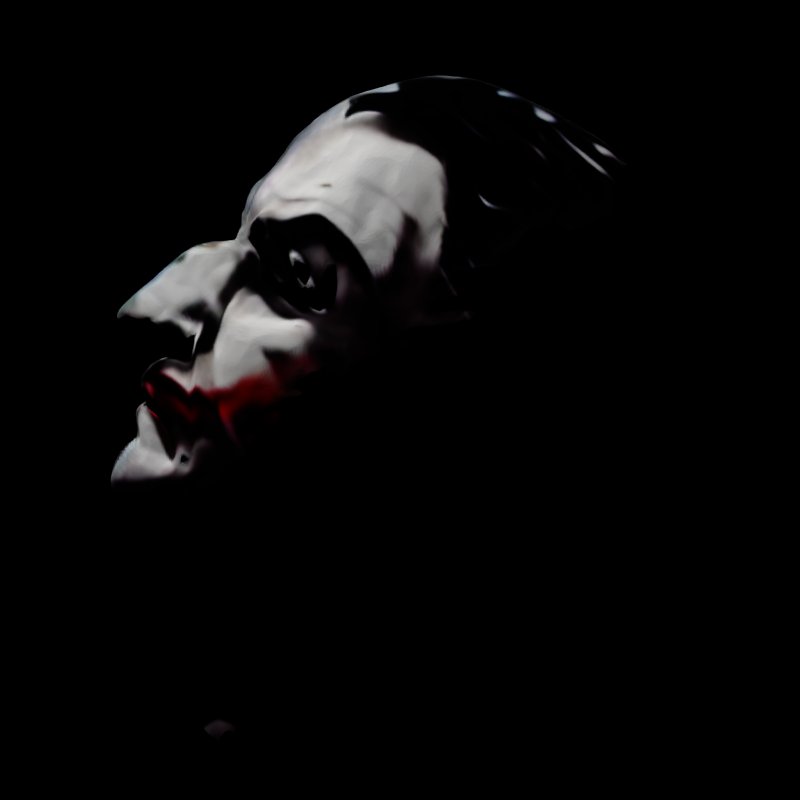}}
\end{tabular}
\caption{Multiview text to projection. Given a scene represented by our method, the \textit{User Input} is a text prompt set by the user per view, the \textit{Original View}s are rendered using our framework and together with automatically created masks (inset) sent to CDC \cite{cdc} for performing content-aware inpainting. After obtaining the \textit{Desired View}, an optimization commences which yields a \textit{Projection Image} best fit for all desired views simultaneously. \textit{Simulated} shows the resulting augmentation when the projection image is used in our framework, whereas \textit{Reprojected} shows the final augmentation when the projection image is used in the actual scene.}
\label{fig:sd}
\end{figure*}

It is a well-known fact among projection mapping enthusiasts and experts alike that immersive augmentations entail the creation of content that complements the scene. The applications in \cref{sec:xray} and \cref{sec:compnensation} demonstrate how our approach addresses traditional tasks, widely studied by prior art. In addition, current times offer exciting novel capabilities thought impossible not long ago, regarding the creation of the immersive content itself. Recent advances in image generation techniques, and specifically diffusion-based models \cite{denoising} have shown great success in creating convincing image content from simple user text prompts. Since our framework formulates the projected image as differentiable with respect to all scene parameters, we can employ it in a multi-view text-to-projection task. As far as we know, this is the first time such an application is possible in a fully automatic manner for projection mapping. Here, the user inputs a free text prompt of the desired view (optionally for multiple views), and the output is a photo-realistic augmentation of the scene that best describes all prompts. It is important to note that theoretically, this type of task may be solved using traditional projection mapping techniques for a single view only. The core contribution of using our framework here is allowing for a multi-view optimization and accounting for view dependent effects. To this end, we developed a pipeline that faithfully allows such an application.

Results can be seen in \cref{fig:teaser} and \cref{fig:sd}. The solution involves defining binary masks that correspond to editable regions in the camera image plane, and to use a stable-diffusion in-painting model \cite{stable_diffusion} to generate the content. However, merely in-painting all of the surface illuminated by the projector can have the undesired effect of ignoring the scene content, which can degrade the immersion of the augmentation in many situations. Better augmentation can be achieved by having the edit take into account the context of the scene. Content aware text driven image editing is a highly active area of research, we found CDC \cite{cdc} to yield faithful results, and we used it instead of directly using an in-painting stable diffusion model. CDC allows for controllable local changes in the input image by specifying a binary mask for regions to be edited and a single hyper parameter $T_{in}$ controlling the extent of the edit. This is highly suitable for our purposes as we wouldn't want the in-painted regions to differ strongly from the underlying surface.

The text-to-projection pipeline consists of the following steps: \textbf{Render}, where the scene is rendered from some user-defined set of viewpoints with the projector shining all-white. This is where the user can look on the renders, and can insert their prompts so they know what they are going to edit. \textbf{Mask}, where we automatically define a mask on the camera's image plane (per view) that is a logical \textit{AND} combination of the following two masks: an Opacity mask, which is 1 where the render had any density along primary rays ($W(s_{i}) > 0$), and a Transmittance mask, which is 1 where the transmittance with respect to the projector is larger than some fixed threshold. \textbf{Inpaint}, where the rendered images, together with the mask and user prompts are sent to CDC for generation using the user text prompt, the result is the desired appearance. Finally, we \textbf{Optimize} for a projection image by first reducing the brightness of the desired appearance to allow for the projectors dynamic range to capture its detail, and then we optimize the projectors texture by exactly the same procedure as in \cref{sec:compnensation}. The result is projected into the scene for the final effect. The full pipeline takes $\sim10$ minutes to converge, we used $T_{in} = 0.1$ for all experiments.

\begin{figure}[ht]
\setlength{\tabcolsep}{1pt}
\centering
\begin{tabular}{ccc}
 \includegraphics[width=0.16\textwidth]{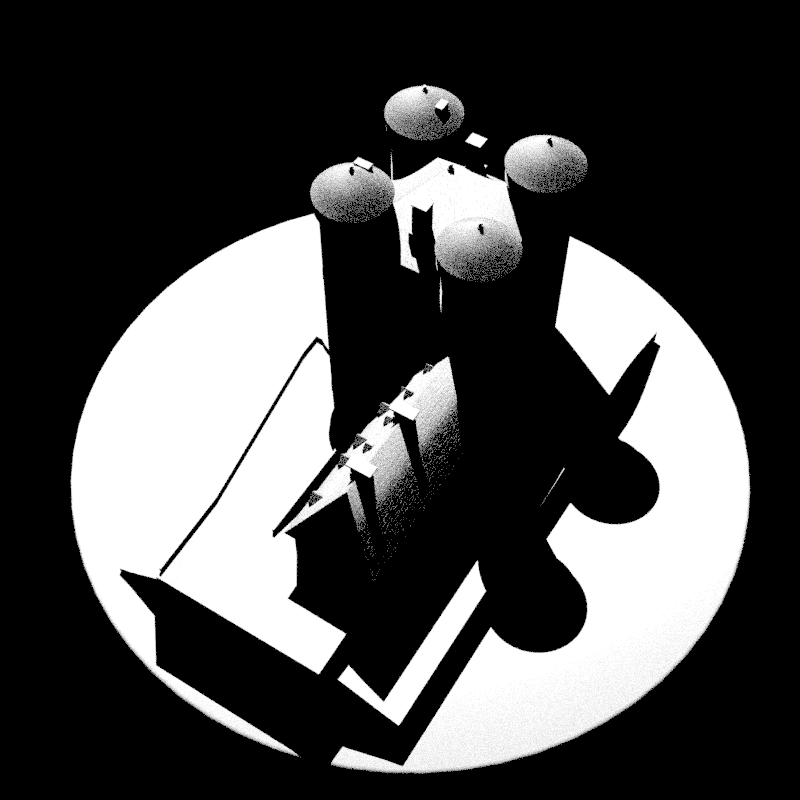} & \includegraphics[width=0.16\textwidth]{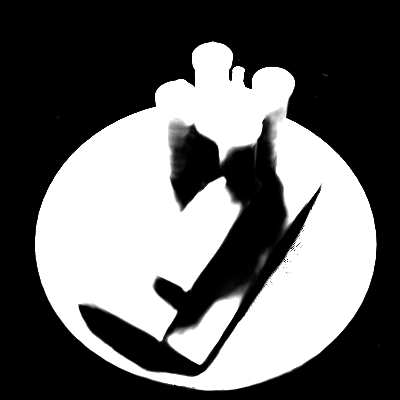} & \includegraphics[width=0.16\textwidth]{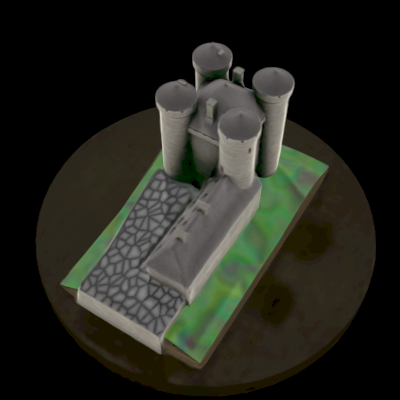} \\
\includegraphics[width=0.16\textwidth]{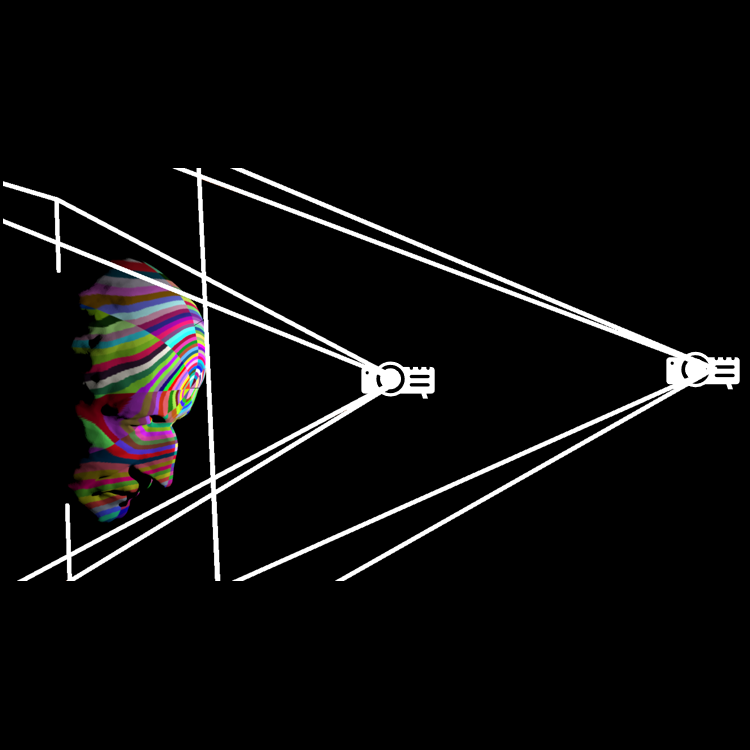} & \includegraphics[width=0.16\textwidth]{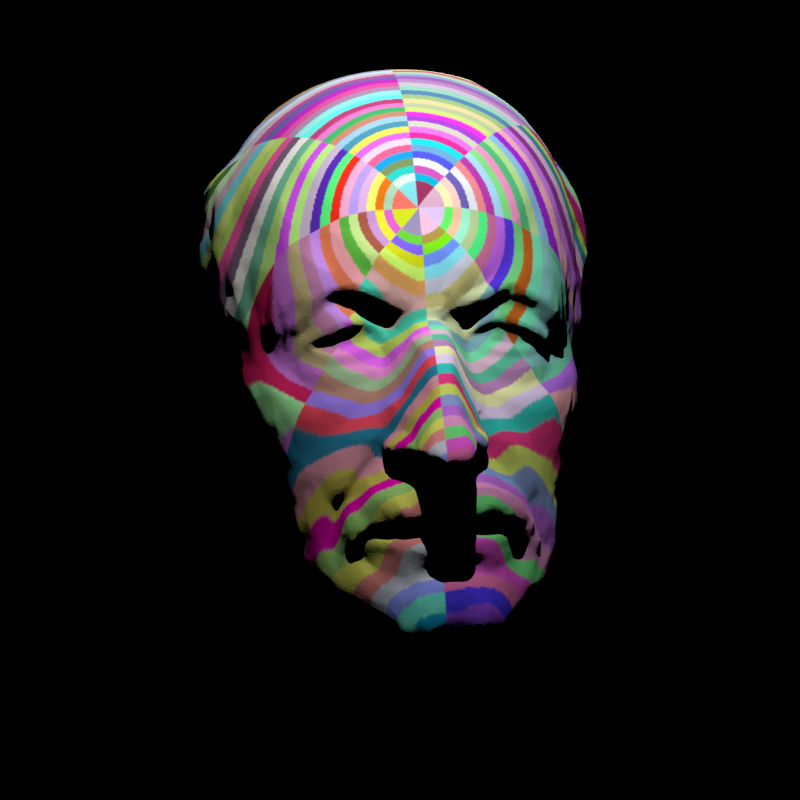} & \includegraphics[width=0.16\textwidth]{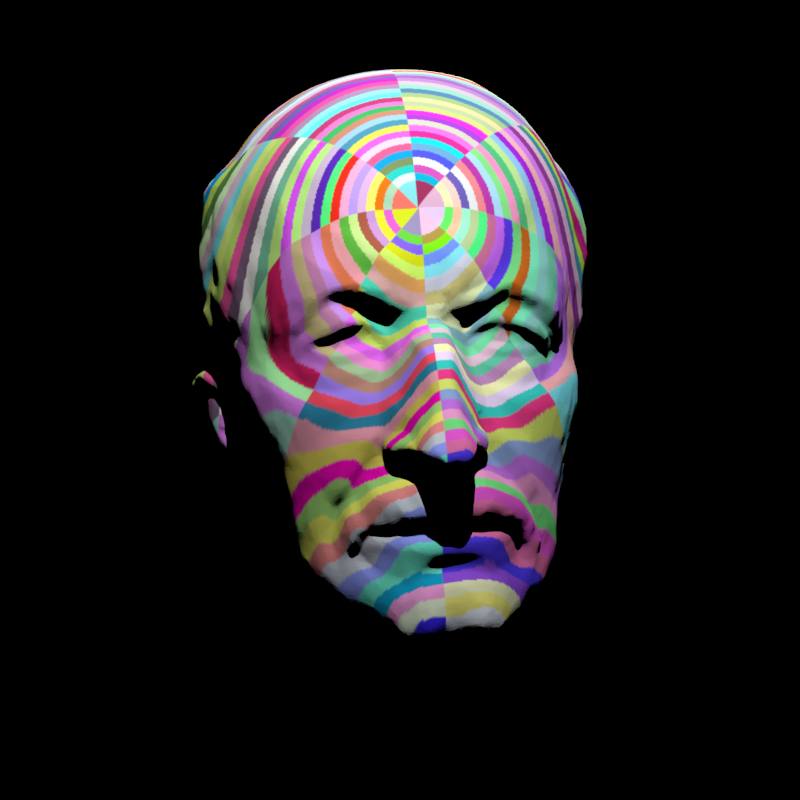}\\
\end{tabular}
\caption{Limitations. Top: the ground truth shadows (left) from the castle turrets are cast to a longer extent than the estimated shadows from transmittance network (middle), which results in darker albedo (right) for patches that were supposed to be in the shadow. Bottom: A scene is illuminated with two different projectors with different focal lengths and distances. They both yield very similar appearances, which sometimes traps the optimization in a local minima.}
\label{fig:limitations}
\end{figure}

\section{Discussion}

Through our experiments, we have confirmed that our networks outperform conventional NeRF-like networks in extracting scene geometry and material features. Specifically, we have demonstrated that while adding a differentiable projector post-training can achieve reasonable results over the right material model, incorporating the projector into the training process not only allows for projector calibration but also enhances the estimation of the scene's geometry and materials.

Our method offers significant potential for users in projection mapping applications, providing more degrees of freedom. For instance, our approach enables users to perform projector compensation for novel viewpoints without requiring additional calibrations, which is not feasible using conventional techniques. Moreover, we leverage the end-to-end neural representation of the projection mapping system, enabling users to edit the real-world appearance using natural language text prompts.

However, our method heavily relies on the simplifying assumption that direct illumination dominates the light field, which may not hold for all projection mapping environments. We discuss the limitations of our method and future directions based on these limitations in the following section.

\subsection{Limitations and future work}

As previously mentioned we do not consider global illumination effects except shadows, and they are estimated indirectly as well (\cref{fig:limitations}, top). Neglecting these is reasonable when the direct light component is strong relative to other effects, but this does not hold for many realistic scenes that exhibit diffuse inter-reflections, sub-surface scattering and caustics. These effects could be explained by the albedo of the surface, resulting in wrong color estimation. In addition, despite lengthy parameter tuning, we consistently found the final appearance estimation of our absorbtion-reflection model to be slightly inferior to the original absorption-emission NeRF \cite{nerf} (when using only frames where the projector shined all black). This could probably be improved by using a more expressing material model, especially with regards to the specular term \cite{nerfactor, psnerf}.

Furthermore, we observed an ambiguity when estimating the projector parameters: projector location and focal length can interchange, yielding similar results despite being quite different (see \cref{fig:limitations}, bottom). This leads to local minima that the optimization sometimes gets stuck on depending on initialization. To disambiguate this, the scene setting should provide sufficient coverage of projector pixels and depth information along the projection axis. Lastly, we note that designing an easier acquisition process for real scenes is a highly desirable goal, as it both requires masking out objects in the foreground and our synthetic video like acquisition (where one pattern is used per view) did not work as well for real scenes. This might be due to the added motion blur associated with taking a video in low light conditions or to the interpolated views being too far off the ground truth extrinsics.

Lastly, despite the visually pleasing results, we note that the text-to-projection application works by generating 2D views, and they are not guaranteed to be 3D consistent in general. This means that artifacts may appear for highly overlapping views (viewpoints which the user inserted a prompts for) without careful attention to the binary masks. Despite this, we have shown success for overlapping scenarios (see supplementary).

\section{Conclusion}
We presented a differentiable framework for performing projection mapping that extracts scene geometry and materials, as well as models the light transport induced by a projector. This allows for calibrating a procam setup and optimizing it for (multiple) novel viewpoints, enabling scene editing for AR applications. In essence, our key contribution is the introduction of the projector primitive into a neural reflectance field, which allows for photo-realistic real scene editing. 

We believe that using a neural reflectance field is a highly suitable representation for direct light applications, and hope to see the usage of our technique to reduce the technical expertise required of projection mapping artists in the future. Two interesting enhancements are to incorporate more projectors into the optimization for supporting more coverage of the scene, and to achieve more photo realism by accounting for inter-reflections and other global illumination effects in unconstrained environments.

\acknowledgments{%
  This work was supported by JSPS KAKENHI Grant Numbers JP20H05958 and JST, PRESTO Grant Number JPMJPR19J2, Japan.
  \\\\
  \noindent This work was supported by Len Blavatnik and the Blavatnik family foundation, the Yandex Initiative in Machine Learning, and ISF (1337/22).
  \\\\
  \noindent All projected images were obtained from \url{https://www.pexels.com/} and are in the public domain.
  The Castle model is an edited version of BlendSwap model ID 29766, CC-0.
  The Pear model is an edited version of BlendSwap model ID 8439, CC-0.
  The Zoo scene is a composition of Bob, Blub, Spot, CC-0.
}

\bibliographystyle{abbrv-doi-hyperref}

\bibliography{egbib}


\appendix 

\end{document}